\RequirePackage{fix-cm}
\documentclass[smallextended]{svjour3} 
\smartqed 
\usepackage{graphicx}
\journalname{}

\usepackage[utf8]{inputenc} \usepackage{url} \usepackage{hyperref}
\usepackage[dvipsnames]{xcolor} \usepackage{tcolorbox}
\usepackage{amsmath} \usepackage{amssymb} \usepackage{latexsym}
\usepackage{stmaryrd,stix}

\usepackage{silence} 

\makeatletter
\patchcmd{\@addmarginpar}{\ifodd\c@page}{\ifodd\c@page\@tempcnta\m@ne}{}{}
\makeatother \reversemarginpar 

\usepackage{tikz} \usetikzlibrary{positioning, fit, calc}
\usetikzlibrary{backgrounds}

\usepackage[backend = biber, style = authoryear-comp, hyperref = true,
url = false, doi = true, giveninits=true, maxbibnames = 4,
maxcitenames = 2, uniquelist=false, uniquename=init
]{biblatex}

\DeclareSourcemap{ \maps[datatype=bibtex]{ \map{
      \step[fieldset=addendum, null]
    }} }

\usepackage{jabbrv}

\DeclareFieldInputHandler{journaltitle}{%
  \def\NewValue{\JournalTitle{#1}}}

\newcommand{\logic}{$\mathcal{PL}$}
\newcommand{\logikey}{\textsc{LogiKEy}}

\newcommand{\ar}{\shortrightarrow} \newcommand{\typearrow}{\ar}
\newcommand{\itype}{\iota} \newcommand{\M}{\mathcal{M}}
\newcommand{\Hk}{\mathcal{H}} \newcommand{\K}{\mathcal{K}}
\newcommand{\I}{\mathcal{I}}
\newcommand{\W}{\mathcal{W}} \newcommand{\V}{\mathcal{V}}
 
\newcommand{\typevar}{\text`t}

\newcommand{\Isabelle}{\textit{Isabelle/HOL}}

\newcommand{\Nitpick}{\textit{Nitpick}}

\newcommand{\enote}[1]{}  

\begin{document}

\title{Modelling Value-oriented Legal Reasoning in \logikey}


\author{Christoph Benzm\"uller\and David Fuenmayor\and Bertram Lomfeld}

\authorrunning{Benzm\"uller, Fuenmayor, Lomfeld}
\institute{C.~Benzm\"uller \at 
University of Bamberg, AI Systems Engineering \&
Freie Universit\"at Berlin, Dep.~of
  Mathematics and Computer Science \\
  \email{c.benzmueller@fu-berlin.de} \and D.~Fuenmayor (corresponding author) \at Universit\'e du Luxembourg, Dep.~of Computer Science \& Freie
  Universit\"at Berlin, Dep.~of Mathematics and Computer Science \\
  \email{david.fuenmayor@uni.lu} \and
  B.~Lomfeld \at Freie Universit\"at Berlin, Dep.~of Law \\
  \email{bertram.lomfeld@fu-berlin.de} }

\date{Received: date / Accepted: date}

\maketitle

\begin{abstract}
The logico-pluralist \logikey\ knowledge engineering methodology and framework is applied to the modelling of a theory of legal balancing in which legal knowledge (cases and laws) is encoded by utilising context-dependent value preferences. The theory obtained is then used to formalise, automatically evaluate, and reconstruct illustrative property law cases (involving appropriation of wild animals) within the {\Isabelle} proof assistant system, illustrating how \logikey\ can harness interactive and automated theorem proving technology to provide a testbed for the development and formal verification of legal domain-specific languages and theories. 
Modelling value-oriented legal reasoning in that framework, we establish novel bridges between latest research in knowledge representation and reasoning in non-classical logics, automated theorem proving, and applications in legal reasoning.
 
\end{abstract}

\pagebreak

\section{Introduction}

Law today has to reflect highly pluralistic environments
in which a plurality of values, world-views and logics coexist.
One function of modern, reflexive law is
to enable the social interaction within and between such worlds
\parencite{teubner83,lomfeld17_methode}.
Any sound model of legal
reasoning needs to be pluralistic, supporting different value systems, value preferences, and maybe even different logical notions, while at the same time reflecting the uniting force of law.

\sloppy Adopting such a perspective, in this paper we apply the logico-pluralistic \logikey\ knowledge engineering methodology and framework \parencite{J48} to the modelling of a theory of value-based legal balancing, a \textit{discoursive grammar} of justification
\parencite{lomfeld19_grammatik}, which we then employ 
to formally reconstruct and automatically assess, using the {\Isabelle} proof assistant system, some illustrative property law cases involving the appropriation of wild animals (termed ``wild animal cases''; cf.~\textcite{bench-capon_sartor03}, \textcite{berman_hafner93}, and \textcite[Ch.\,II.\,A.1]{MerillSmith17} for background).
Lomfeld's \textit{discoursive grammar} is encoded, for our purposes, as a logic-based domain-specific language (DSL) in which the legal knowledge embodied in statutes and case corpora becomes represented as \textit{context-dependent} preferences among (combinations of) values constituting a pluralistic value system or ontology. This knowledge can thus be complemented by further legal and world knowledge, e.g., from legal ontologies~\parencite{CasanovasPPEV14,DBLP:conf/icail/HoekstraBBB09}.

\tikzset{block/.style={draw, thick, text width=2cm , minimum
    height=1.3cm, align=center}, line/.style={-latex} }
 
\tikzset{ font={\fontsize{12pt}{12}\selectfont}}
  
\tikzset{testpic/.pic={ 
\node[block, fill=Green!20, text width=10cm,
    minimum width=14cm, minimum height=2cm] (m1) { \textbf{\huge L3 --- Applications} };
\node[block, below=.5cm of m1, fill=Orange!30, text width=12cm,
    minimum width=14cm, minimum height=2cm] (m2) { \textbf{\huge L2 --- Domain-Specific  Language(s)} };
\node[block, below=.5cm of m2,
    fill=Yellow!20, text width=10cm, minimum width=14cm, minimum
    height=2cm] (m3) { \textbf{\huge L1 --- Object Logic(s)} };
\node[block, below=.5cm of m3,
    fill=Blue!20, text width=11cm, minimum width=14cm, minimum
    height=2cm] (m4) { \textbf{\huge L0 --- Meta-Logic (HOL)} };
\node[block, above right=-.5cm and
    1.5cm of m1, fill=black!20, text width=8cm, minimum width=9cm,
    minimum height=2cm, rotate=-90] (r1) { \textbf{\huge \logikey\
        Methodology} };
 
  \begin{scope}[on background layer] 
    \node[draw, fill=Blue!40, inner xsep=5mm, inner ysep=5mm, fill
    opacity=0.4,fit=(m1)(m2)(m3)(m4)](m1tom4){};
  \end{scope}
 
 \begin{scope}[on background layer] 
   \node[draw, fill=black!20, inner xsep=20mm,inner ysep=5mm, fill
   opacity=0.3,fit=(m1)(m2)(m3)(m4)(m1tom4)](all){};
 \end{scope}
}}

\begin{figure}[!bp] \centering \resizebox{.70\columnwidth}{!}{
    \begin{tikzpicture}
      \pic {testpic};
    \end{tikzpicture}
  }
  \caption{\logikey\ development methodology} \label{fig:Methodology2}
\end{figure}

The \logikey\ framework supports plurality at different layers; cf. Fig.~\ref{fig:Methodology2}.
Classical higher-order logic (HOL) is fixed as a \textit{universal meta-logic} \parencite{J41} at the base layer (L0), on top of which a plurality of (combinations of) object logics can become encoded (layer L1). Employing these logical notions we can now articulate a variety of logic-based domain-specific languages (DSLs), theories and ontologies at the next layer (L2), thus enabling the modelling and automated assessment of different application scenarios (layer L3).
These linked layers, as featured in the \logikey\ approach, facilitate fruitful interdisciplinary collaboration between specialists in different AI-related domains and domain experts in the design and development of knowledge-based systems. 

\logikey, in this sense, fosters a \emph{division of labour} among different specialist roles. For example, `logic theorists' can concentrate on investigating the semantics and proof calculi for different object logics, while `logic engineers' (e.g., with a computer science background) can focus on the encoding of suitable combinations of those formalisms in the meta-logic HOL and on the development and/or integration of relevant automated reasoning technology. Knowledge engineers can then employ these object logics for knowledge representation (by developing ontologies, taxonomies, controlled languages, etc.), while domain experts (ethicists, lawyers, etc.) collaborate with requirements elicitation and analysis, as well as providing domain-specific counseling and feedback.
These tasks can be supported in an integrated fashion by harnessing (and extending) modern mathematical proof assistant systems (aka.~interactive theorem provers), which thus become a testbed for the development of logics and ethico-legal theories.

The work reported below is a
\logikey-supported collaborative research effort involving two computer scientists
(Benzm\"uller \& Fuenmayor) together with a lawyer and legal philosopher (Lomfeld),  
who have joined forces with the goal of studying the computer-encoding
and automation of a theory of value-based legal balancing: Lomfeld's
\textit{discoursive grammar} \parencite{lomfeld19_grammatik}. A formally-verifiable legal domain-specific language (DSL) has been developed for the encoding of this theory (at \logikey's layer L2). This DSL has been built on top of a suitably chosen object-logical language: a modal logic of preferences
(at layer L1), by drawing upon the representation and reasoning infrastructure integrated within the proof assistant {\Isabelle} (layer L0).
The resulting system is then employed for the assessment of legal cases in property law (at layer L3), which includes the formal modelling of background legal and world knowledge, as required to enable the verification of predicted legal case outcomes and the
automatic generation of value-oriented logical justifications (backings) for them. 

From a wider perspective, \logikey\ aims at supporting the practical development of computational tools for legal and normative reasoning based on formal methods.
One of the main drives for its development
has been the introduction of automated reasoning techniques for the design, verification (offline \& online), and control of intelligent autonomous systems, as a step towards \textit{explicit ethical agents}~\parencite{Moor2009,Scheutz17}. The argument here is that ethico-legal control mechanisms (such as ethical governors; cf.~\textcite{Arkin2009}) of intelligent autonomous  systems should be understood and designed as knowledge-based systems, where the required ethical and legal knowledge becomes \textit{explicitly} represented as a logical theory, i.e., as a set of formulas (axioms, definitions and theorems) encoded in a logic.
We have set a special focus on the (re-)use of modern proof assistants based on HOL ({\Isabelle}, \textit{HOL-Light}, \textit{HOL4}, etc.) and integrated automated reasoning tools (\textit{theorem provers} and \textit{model generators}) for the interactive development and verification of ethico-legal theories. To carry out the technical work reported in this paper, we have chosen to work with {\Isabelle}, but the essence of our contributions can easily be applied to other proof assistants and automated reasoning systems for HOL.

Technical results concerning in particular our {\Isabelle} encoding have been presented at the \textit{International Conference on Interactive Theorem Proving} (ITP 2021) \parencite{C91}, and earlier ideas have been discussed at the \textit{Workshop on Models of Legal Reasoning} (MLR 2020). In the present paper, we elaborate on these results and provide a more self-contained exposition, by giving further background information on Lomfeld's \textit{discoursive grammar}, on the meta-logic HOL, and on the modal logic of preferences by \textcite{BenthemGR09}. More fundamentally, this paper presents the full picture, as framed by the underlying \logikey\ framework, and highlights methodological insights, applications, and perspectives relevant to the \textit{AI \& Law} community. One of our main motivations is to help build bridges between recent research in knowledge representation and reasoning in non-classical logics, automated theorem proving, and applications in normative and legal reasoning.

\paragraph{Paper structure:} 
After summarising Lomfeld's theory of value-based legal balancing
in \S\ref{sec:ValueTheory}, we briefly depict the \logikey\ development
and knowledge engineering methodology in \S\ref{sec:methodology}, and
our meta-logic HOL in \S\ref{sec:hol}.
We then outline our object logic of choice 
-- a (quantified) modal logic of preferences --
in \S\ref{sec:Logic}, where we also present
its encoding in the meta-logic HOL 
and formally verify the preservation of meta-theoretical properties using the {\Isabelle} proof assistant.
Subsequently, we
model in \S\ref{sec:ValueOntology}
Lomfeld's legal theory and provide a custom-built DSL, 
which is again formally assessed using {\Isabelle}.
As an illustrative application of our framework,
we present in \S\ref{sec:Application} the formal reconstruction and assessment of 
well-known example legal cases in property law (``wild animal cases''),
together with some considerations regarding the encoding of required legal and world knowledge.
Related and further work is addressed in \S\ref{sec:relwork},
and \S\ref{sec:Conclusion}~concludes the article.

\sloppy The {\Isabelle} sources of our formalisation work are available at
\url{http://logikey.org} under
\textit{\href{https://github.com/cbenzmueller/LogiKEy/tree/master/Preference-Logics/EncodingLegalBalancing}{Preference-Logics/EncodingLegalBalancing}}. They are also explained in some detail in the Appendix \ref{TheAppendix}.\enote{(Note to the reviewers/editors: Eventually our commented source
    data collection in this Appendix \ref{TheAppendix} should preferably be presented in
    a back-to-back data publication in a data journal (e.g.~
    Springer's ``Scientific Data'' or Elsevier's ``Data in Brief''). For the moment, however, we keep
    it here and wait for further advice from the reviewers/editors.)}

\section{A Theory of Legal Values: \textit{Discoursive Grammar}
  of Justification}
\label{sec:ValueTheory}

The case study with which we illustrate the \logikey\ methodology in the present paper consists in the formal encoding and assessment on the computer of a theory of value-based legal balancing, as put forward by \cite{lomfeld19_grammatik}.
Lomfeld himself has played the role of the domain expert in our joint research, which from a methodological perspective, can be characterised as being both in part theoretical and in part empirical. Lomfeld's primary role has been to provide background legal domain knowledge and to assess the adequacy of our formalisation results, while informing us of relevant conceptual and legal distinctions that needed to be made. In a sense, this created a win-win situation in which both Lomfeld's theory and \logikey's methodology have been put to the test. This section presents Lomfeld's theory and discusses some of its merits in comparison to related approaches.

Logical reconstructions quite often separate
deductive rule application and inductive case-contextual interpretation as completely distinct ways of legal reasoning (cf. the overview
in \textcite{prakken_sartor15}). Understanding the whole process of legal reasoning as an exchange of opposing action-guiding arguments, i.e., practical argumentation \parencite{alexy78,feteris2017}, a strict separation between logically distinct ways of legal reasoning breaks down. Yet, a variety of modes of rule-based \parencite{hage97,prakken1997,modgil_prakken_2018},
case-based \parencite{ashley90,aleven1997,horty11} and value-based
\parencite{berman_hafner93,bench-capon_ea05,grabmair2016} reasoning coexist in legal theory and (court) practice.

In line with current computational models combining these different modes of reasoning 
\parencite[e.g.,][]{bench-capon_sartor03,DBLP:conf/icail/MaranhaoS19}, we argue that a
discourse theory of law can consistently integrate them in the form of a multi-level system of legal reasoning. Legal rules or case precedents can thus be translated into (or analysed as) the
balancing of plural and opposing (socio-legal) values on a deeper level of reasoning
\parencite{lomfeld15_vertrag}.

There exist indeed some models for quantifying legal balancing, i.e., for weighing competing reasons in a case
\parencite[e.g.,][]{alexy03,sartor10}. We share the opinion that these approaches need to get
``integrated with logic and argumentation to provide a comprehensive
account of value-oriented reasoning'' \parencite{sartor18}. Hence a suitable
model of legal balancing would need to reconstruct
rule subsumption and case distinction as argumentation processes
involving conflicting values.

Here, the functional differentiation of legal norms into \textit{rules} and
\textit{principles} reveals its potential
\parencite{dworkin-taking-1978,alexy00}. Whereas legal rules have a
binary all-or-nothing validity driving out conflicting rules, legal
principles allow for a scalable dimension of weight. Thus, principles
could outweigh each other without rebutting the normative validity of
colliding principles.  Legal principles can be understood as a set
of plural and conflicting values on a deep level of socio-legal
balancing, which is structured by legal rules on an explicit and more
concrete level of legal reasoning \parencite{lomfeld15_vertrag}. The
two-faceted argumentative relation is partly mirrored in the
functional differentiation between \textit{goal-norms} and \textit{action-norms}
\parencite{sartor10} but should not be mixed up with a general
understanding of principles as abstract rules
\parencite{raz1972,verheij1998integrated} or as specific
constitutional law elements \parencite{neves2021,barak2012}.

In any event, if preferences between defeasible rules are
reconstructed and justified in terms of preferences between
underlying values, some questions about values necessarily pop up.
In the words of \textcite{bench-capon_sartor03}:
``Are values scalar? [\dots] Can values be ordered at all? [\dots] How
can sets of values be compared? [\dots] Can values conflict so the promotion of their combination is worse than promoting either separately? Can several less important values together overcome a
more important value?''.

Hence an encompassing approach for legal reasoning as practical
argumentation needs not only a formal reconstruction of the relation
between legal values (or principles) and legal rules, but also a
substantial framework of values (a basic value system) that allows to
systematise value comparison and conflicts as a \textit{discoursive
  grammar} \parencite{lomfeld15_vertrag,lomfeld19_grammatik} of
argumentation. In this article we define a value system not as a
  ``preference order on sets of values'' \parencite{weide_ea2010} but as
  a pluralistic set of values which allow for different preference
  orders.  The computational conceptualisation (as a formal logical theory) of such a set of
representational primitives for a pluralist basic value system can
then be considered as a value ``ontology''
\parencite{gruber1993,gruber2009,smith2003}, which of course needs to
be complemented by further ontologies for relevant background legal and world
knowledge (see e.g.~\textcite{CasanovasPPEV14,DBLP:conf/icail/HoekstraBBB09}).

Combining the discourse-theoretical idea that legal reasoning is
practical argumentation with a two-faceted model of legal norms, legal
\textit{rules} could be logically reconstructed as conditional preference
relations between conflicting underlying \textit{value principles}
\parencite{alexy00,lomfeld15_vertrag}. The legal consequence of a rule
$R$ thus implies the preference of value principle $A$ over value
principle $B$, noted $A > B$ (e.g. health security outweighs freedom to
move).\footnote{In \S\ref{sec:ValueOntology} these values will be
  assigned to particular parties/actors, so that ruling in favour of different
  parties may promote different values.} This value preference applies
under the condition that the rule's prerequisites $E_1$ and $E_2$
hold. Thus, if the propositions $E_1$ and $E_2$ are true in a given
situation (e.g.~a virus pandemic occurs and voluntary shut down
fails), then the value preference $A > B$ obtains. This value
preference can be said to weight or \textit{balance} the two values $A$ and
$B$ against each other. We can thus translate this concrete legal rule
as a \textit{conditional preference relation} between colliding value
principles:
\begin{align*}
  R:  (E_1 \wedge E_2)   \Rightarrow   A \prec B
\end{align*}

More generally, $A$ and $B$ could also be structured as aggregates of value principles, whereas the condition of the rule can consist in a conjunction of arbitrary propositions.
Moreover, it may well happen that, given some conditions,  several rules become relevant in a concrete legal case. In such cases the rules determine a structure of legal balancing between conflicting plural value principles. Moreover,
making explicit the underlying \textit{balancing} of values against each other (as value
preferences) helps to justify a legal consequence
(e.g.~sanctioned lock-down) or ruling in favour of a party
(e.g.~defendant) in a legal case.

But which value principles are to be balanced? How to find a suitable
justification framework? Based on comparative discourse analyses in
different legal systems, one can reconstruct a general dialectical (antagonistic)
taxonomy of legal value principles used in (at least Western)
legislation, legislative materials, cases, textbooks and scholar
writings \parencite{lomfeld15_vertrag}. The idea is to provide a plural and yet
consistent system of basic legal values and principles, independent of concrete
cases or legal fields, to justify legal decisions.

\begin{figure}[!htb]
  \centering
  \includegraphics[width=.7\textwidth]{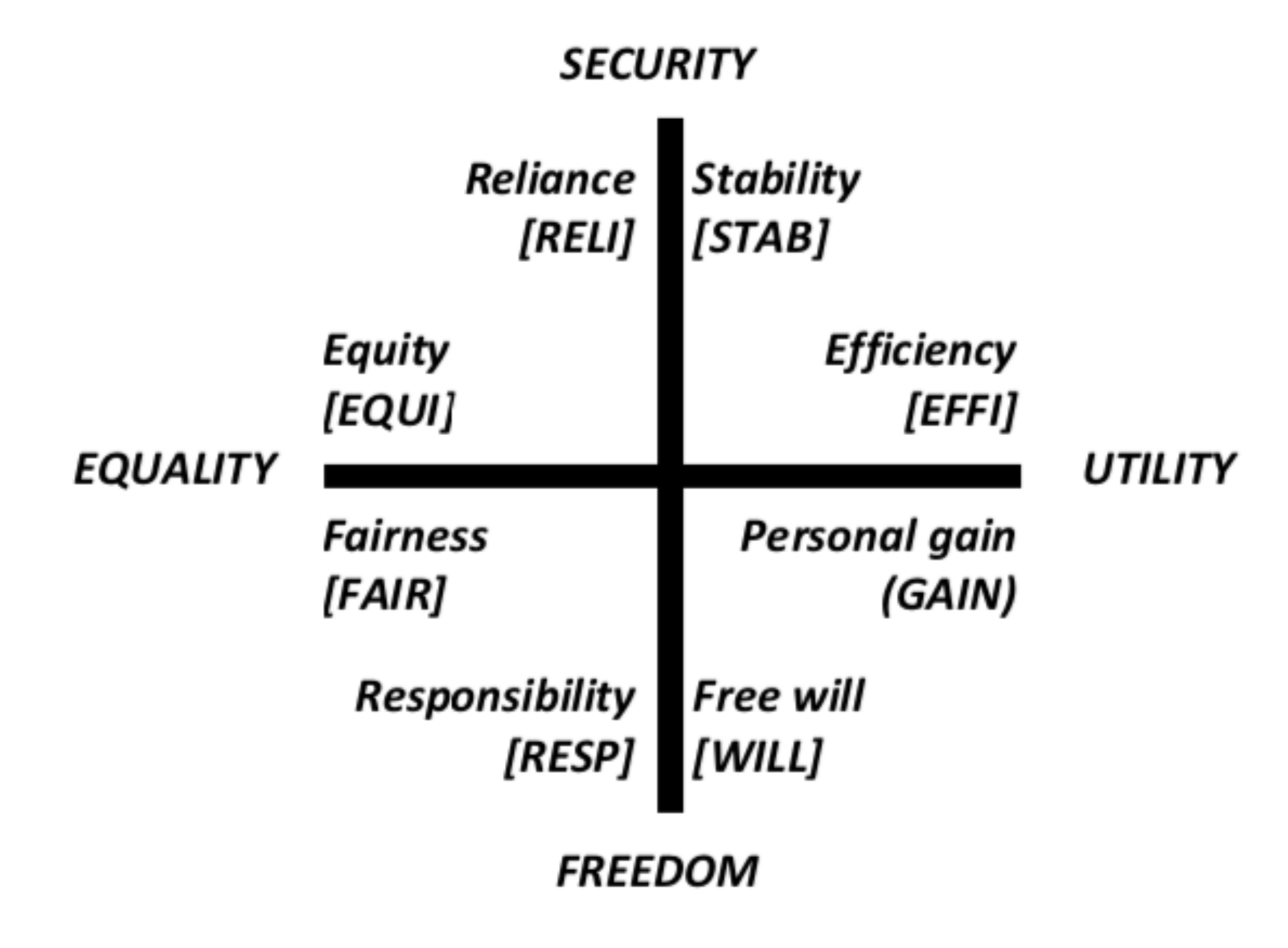}
  \caption{Basic legal value system ({ontology}) by \textcite{lomfeld19_grammatik}}
  \label{fig:ValueOntology}
\end{figure}

The proposed legal value system \parencite{lomfeld19_grammatik},
see Fig.~\ref{fig:ValueOntology}, is consistent with many existing
taxonomies of antagonistic psychological \parencite{rokeach1973,schwartz1992},
political \parencite{eysenck1954,mitchell2007} and economic values
\parencite{clark1991}.\footnote{All these taxonomies are pluralist
  frameworks that do encompass differences in global value patterns
  and cultural value evolution
  \parencite{hofstede2001,inglehart2018}. For an approach oriented at
  Maslow's hierarchy of needs \parencite{bench-capon2020}.}  
In all social orders one can observe a general antinomy between individual
and collective values. Ideal types of this fundamental dialectic are: the basic
value of FREEDOM for the individual, and the basic value of
SECURITY for the collective perspective.
Another classic social value
antinomy is between a functional-economic (utilitarian) and a more idealistic (egalitarian)
viewpoint, represented in the ethical debate by the 
essential dialectic concerning the basic values of UTILITY versus EQUALITY. These four normative poles stretch an axis of value coordinates for the general value system construction. We thus speak of a normative dialectics, since each of the antagonistic basic values and related principles can (and in most situations will) collide with each other.

Within this dialectical matrix eight more concrete legal \textit{value principles} are identified.  FREEDOM represents the normative basic value of
individual autonomy and comprises the legal (value) principles of --more functional-- individual choice or `free will' (WILL) and --more idealistic--
(self-)`responsibility' (RESP). The basic value of SECURITY addresses the
collective dimension of public order and comprises the legal
principles of --more functional-- collective `stability' (STAB) of a
social system and --more idealistic-- social trust or `reliance'
(RELI). The value of UTILITY means economic welfare on the personal
and collective level and comprises the legal principles of collective
overall welfare-maximisation, i.e., `efficiency' (EFFI) and individual
welfare-maximisation, i.e., economic benefit or `gain' (GAIN). Finally,
EQUALITY represents the normative ideal of equal treatment and equal
allocation of resources and comprises the legal principles of --more
individual-- equal opportunity or procedural `fairness' (FAIR) and --more
collective-- distributional equality or `equity' (EQUI).

This legal value system (or ontology) can consistently cover
existing value sets from AI \& Law accounts of value-oriented reasoning
\parencite[e.g.,][]{berman_hafner93,bench-capon12,gordon_walton12,sartor10},
mostly exemplified by modelling famous common law property cases, in particular, ``wild animal cases''.
A key feature of Lomfeld's \textit{discoursive grammar} of dialectical values consists in its purely qualitative modelling of legal balancing in terms of context-dependent logic-based preferences among values, without any need for determining quantitative weights.

\section{The \logikey\ Methodology} \label{sec:methodology}

\logikey, as a methodology \parencite{J48}, refers to the principles underlying the
organisation and the conduct of complex knowledge design and
engineering processes, with a particular focus on the legal and
ethical domain.
Knowledge engineering refers to all the
technical and scientific aspects involved in building, maintaining and
using knowledge-based systems employing logical formalisms as a representation language.
In particular, we speak of \textit{logic engineering} to highlight those tasks directly related to the syntactic and semantic definition, as well as to the meta-logical encoding and automation, of different combinations of object logics.
It is also \logikey's objective to fruitfully integrate
contributions from different research communities (such as interactive and automated
theorem proving, non-classical logics, knowledge representation, and domain specialists)
and to make them accessible at a suitable level of abstraction and technicality to practitioners
in diverse fields.

A fundamental characteristic of the \logikey\ methodology consists in the utilisation of classical higher-order logic (HOL, cf.~\textcite{J43}) as a general-purpose logical formalism in which to encode different (combinations of) object logics. 
This enabling technique is known as 
shallow\footnote{Shallow semantical embeddings are different from
  \emph{deep embeddings} of an object logic. In the latter case the
  syntax of the object logic is represented using an inductive data
  structure (e.g., following the definition of the language).  The
  semantics of a formula is then evaluated by recursively traversing
  the data structure, and additionally a proof theory for the logic
  maybe be encoded. Deep embeddings typically require technical
  inductive proofs, which hinder proof automation, that can be avoided
  when shallow semantical embeddings are used instead. For more
  information on shallow and deep embeddings we refer to the
  literature \parencite{DeepShallow,DeepShallow2}.} semantical
embeddings (SSEs).
HOL thus acts as the substrate in which a plurality of logical languages,
organised hierarchically at different abstraction layers, become ultimately encoded and reasoned with.
This in turn enables the provision of
powerful tool support: we can harness mathematical proof assistants (e.g. {\Isabelle}) as a testbed for the development of logics, and ethico-legal DSLs and theories.
More concretely, off-the-shelf theorem provers and (counter-)model generators
for HOL, as provided, e.g., in the interactive proof assistant
{\Isabelle} \parencite{blanchette2016hammering}, are assisting the \logikey\
knowledge \& logic engineers (as well as domain experts) to \textit{flexibly experiment} with underlying (object)
logics and their combinations, with general and domain knowledge, and
with concrete use cases---all at the same time. Thus, continuous
improvements of these off-the-shelf provers, without further ado,
leverage the reasoning performance in \logikey.

The \logikey\ methodology, cf.~Fig.~\ref{fig:Methodology2}, has been instantiated in this article to support
and guide the simultaneous development of the different modelling
layers as depicted in Fig. \ref{fig:Methodology}, and which will be the subject of discussion in the following sections.
According to the logico-pluralistic nature of \logikey, only the lowest
layer (L0), meta-logic HOL (cf.~\S\ref{sec:hol}), remains fixed, while all
other layers are subject to dynamic adjustments until a satisfying
overall solution in the overall modelling process is reached.
At the next layer (L1) we are faced with the choice of an object logic, in our case a
modal logic of preference (cf.~\S\ref{sec:Logic}). A legal DSL (cf.~\S\ref{sec:ValueOntology}), created after Lomfeld's \textit{discoursive
grammar} (cf.~\S\ref{sec:ValueTheory}), further extends this object logic at a higher level
of abstraction (layer L2).
At the upper layer (layer L3), we use this legal DSL 
to encode and automatically assess some example legal cases (``wild animal cases'') in
property law (cf.~\S\ref{sec:Application}) by relying upon previously encoded background legal and world knowledge.\footnote{In some cases it can be convenient to split one or more layers into sublayers. For instance, in our case study (cf.~\S\ref{sec:Application}), layer L3 has been further subdivided to allow for a more strict separation between general legal \& world knowledge (legal concepts and norms), cf.~\S\ref{sec:LegalAndWorldKnowledge}, from its \textit{application} to relevant facts in the process of deciding a case (factual/contextual knowledge), cf.~\S\ref{subsec:app:pierson}.}
The higher layers thus make use of the concepts introduced at the
lower layers.  Moreover, at each layer, the encoding efforts are guided
by selected tests and `sanity checks' in order to formally verify relevant
properties of the encodings at and up to that level.

\tikzset{testpic/.pic={ 
\node[block, fill=Green!20, text width=10cm,
    minimum width=14cm, minimum height=2cm] (m1) { \textbf{\huge
        L3 --- Assessment of Legal Cases \\[.1em] (Wild Animals' Appropriation) 
        } };
\node[block, below=.5cm of m1, 
    fill=Orange!30, text width=12cm, minimum width=14cm, minimum
    height=2cm] (m3) { \textbf{\huge L2 --- Value-oriented Legal Theory 
    } };
\node[block, below=.5cm of m3, 
    fill=Yellow!20, text width=12cm, minimum width=14cm, minimum
    height=2cm] (m5) { \textbf{\huge 
    L1 --- Modal Logic of Preferences} };
\node[block, below=.5cm of m5,
    fill=Blue!20, text width=11cm, minimum width=14cm, minimum
    height=2cm] (m6) { \textbf{\huge 
    L0 --- HOL} };
\node[block, above right=-0.5cm and 
    1.5cm of m1, fill=black!20, text width=9cm, minimum width=9cm, 
    minimum height=2cm, rotate=-90] (r1) { \textbf{\huge \logikey\
        Methodology} };
 
  \begin{scope}[on background layer] 
    \node[draw, fill=Blue!40, inner xsep=5mm, inner ysep=5mm, fill
    opacity=0.4,
    fit=(m1)(m3)(m5)(m6)](m1tom6){};
  \end{scope}
 
 \begin{scope}[on background layer] 
   \node[draw, fill=black!20, inner xsep=20mm,inner ysep=5mm, fill
   opacity=0.3,fit=(m1)(m2)(m3)(m4)(m5)(m6)(m1tom6)](all){};
 \end{scope}
}}

\begin{figure}[!tp] \centering \resizebox{.70\columnwidth}{!}{
    \begin{tikzpicture}
      \pic {testpic};
    \end{tikzpicture}
  }
  \caption{\logikey\ development methodology as instantiated in
    the given context} \label{fig:Methodology}
\end{figure}
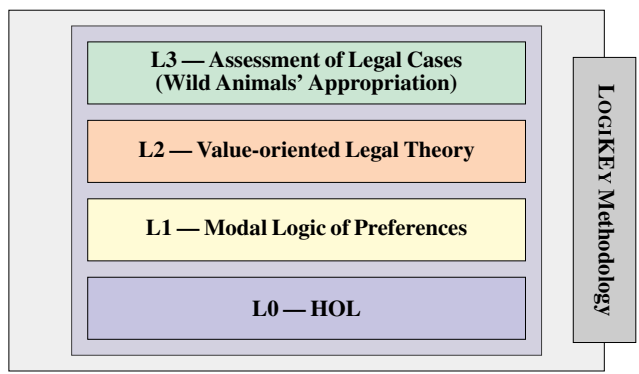

It is worth noting that the application of our approach to deciding concrete legal cases 
reflects ideas in the AI \& Law literature about understanding the solution of legal
cases as theory construction, i.e., ``building, evaluating and using
theories'' \parencite{bench-capon_sartor03}.\footnote{The authors judiciously quote
  \textcite{mccarty1995implementation}: ``The task for a lawyer or a
  judge in a `hard case' is to construct a theory of the disputed
  rules that produces the desired legal result, and then to persuade
  the relevant audience that this theory is preferable to any theories
  offered by an opponent.''}  This multi-layered, iterative knowledge engineering
process is supported in our \logikey\ framework by adapting
interactive and automated reasoning technology for HOL (as a meta-logic).

An important aspect thereby is that the \logikey\ methodology foresees
and enables the knowledge engineer to flexibly switch between the
modelling layers and to suitably adapt the encodings also at lower
layers if needed.  The engineering process thus has backtracking
points and several work cycles may be required; thereby the higher
layers may also pose modification requests to the lower layers.
Such demands, unlike in most other approaches, may also involve far-reaching modifications of the chosen logical foundations, e.g., in the particularly chosen modal preference logic.

The work we present in this article is in fact the result of an iterative, give-and-take
process encompassing several cycles of modelling, assessment and
testing activities, whereby a (modular) logical theory gradually
evolves until eventually reaching a state of highest coherence and
acceptability. In line with previous work on \textit{computational hermeneutics} \parencite{B19}, we may then speak of arriving at a state of \textit{reflective
  equilibrium} \parencite{sep-reflective-equilibrium}, as the end-point of an iterative process of mutual adjustment among (general) principles and (particular) judgements, where the latter are intended to become logically entailed by the former. It is also worth noting that the notion of \textit{reflective equilibrium} has been introduced by the philosopher John Rawls in moral and political philosophy as a method for the development of his \textit{theory of justice} \parencite{rawls1999TJ}, an analogous endeavour to ours in the present work. In fact, an earlier formulation of this approach is often credited to the philosopher Nelson Goodman, who proposed it as a method for the development of (inference rules for) deductive and inductive logical systems \parencite{goodman1955FFF}, again, very much in the spirit of \logikey.

\section{Meta-logic (L0) -- Classical Higher-Order Logic} \label{sec:hol}

To keep this article sufficiently self-contained we briefly introduce
a classical higher-order logic, termed HOL; more detailed information
on HOL and its automation can be found in the literature
\parencite{Andrews:gmdacitt72,Andrews72a,J43,J6,B5}.

The notion of HOL used in this article refers to a simply typed logic
of functions that has been put forward by \textcite{Church40}. 
Hence all terms of HOL get assigned a fixed and unique type, commonly written
as a subscript (i.e., the term $t_\alpha$ has $\alpha$ as its type).
HOL provides $\lambda$-notation, as an elegant and useful means to denote
unnamed functions, predicates and sets; $\lambda$-notation also
  supports compositionality, a feature we heavily exploit to obtain
  elegant, non-recursive encoding definitions for our logic embeddings
  in the remainder. Types in HOL eliminate paradoxes and
inconsistencies.

HOL comes with a set ${T}$ of \emph{simple types}, which is freely
generated from a set of \emph{basic types} $BT \supseteq \{o,\itype\}$
using the function type constructor $\typearrow$ (written as a right-associative infix operator). For instance, $o$, $\itype\,{\typearrow}\,o$ and $\itype\,{\typearrow}\,\itype\,{\typearrow}\,\itype$ are types. The type $o$ denotes
a two-element set of truth-values and $\itype$ denotes a non-empty set
of individuals.\footnote{{In this article, we will actually associate
    type $\itype$ later on (cf.~\S\ref{sec:Implementation}) with the domain
    of possible states/worlds.}}  Further base types may be added as
the need arises.

The \emph{terms} of HOL are inductively defined
starting from typed constant symbols ($C_{\alpha}$) and typed
variable symbols ($x_{\alpha}$) using \emph{$\lambda$-abstraction}
($(\lambda x_{\alpha}.\, s_{\beta})_{\alpha \typearrow \beta} $) and
\emph{function application}
($(s_{\alpha \typearrow \beta}\, t_\alpha)_{\beta}$), thereby obeying
type constraints as indicated. Type subscripts and parentheses are usually omitted to improve readability, if obvious from the context or irrelevant.
Observe that $\lambda$-abstractions introduce unnamed functions. For example, the
function that adds $2$ to a given argument $x$ can be encoded as
$(\lambda x.\, x + 2)$, and the function that adds two numbers can be encoded as $(\lambda x.\, (\lambda y.\, x + y))$.\footnote{Note that functions of more
  than one argument can be represented in HOL in terms of functions of
  one argument. In this case the values of these one-argument function
  applications are themselves functions, which are subsequently
  applied to the next argument. This technique, introduced by  \textcite{Schoenfinkel1924}, is commonly called \textit{currying}; cf.~\textcite{J43}.} HOL terms of type $o$ are also called
formulas.\footnote{HOL formulas (layer L0) should not be confused with the
    object-logical formulas (layer L1); 
    the latter will
    later be identified in \S\ref{sec:Implementation} with HOL
    predicates of type $\itype\,{\rightarrow}\,o$.}
    
Moreover, to obtain a proper logic, we add
$\neg_{o \typearrow o},\vee_{o \typearrow o \typearrow o}$ and
$\Pi_{(\alpha \typearrow o) \typearrow o}$ (for each type $\alpha$) as
predefined typed constant symbols to our language and call them
\emph{primitive logical connectives}.  \emph{Binder notation} for
quantifiers $\forall x_{\alpha}\ s_o$ is used as an abbreviation for
$\Pi_{(\alpha \typearrow o)\typearrow o}\lambda x_{\alpha}. s_{o}$.
  
The \emph{primitive logical connectives} are given a fixed
interpretation as usual, and from them other logical connectives can
be introduced as abbreviations. Material implication
$s_o \rightarrow t_o$ and existential quantification
$\exists x_\alpha s_o$, for example, may be introduced as shortcuts
for $\neg s_o\,\vee\,t_o$ and $\neg\forall x_\alpha \neg s_o$,
respectively.  Additionally, \emph{description or choice operators} or
\emph{primitive equality} $=_{\alpha\typearrow \alpha \typearrow o}$
(for each type $\alpha$), abbreviated as $=^\alpha$, may be added.
Equality can also be defined by exploiting Leibniz' principle,
expressing that two objects are equal if they share the same
properties.

It is well known that, as a consequence of G\"odel's Incompleteness
Theorems, HOL with standard semantics is necessarily incomplete. In
contrast, theorem proving in HOL is usually considered with respect to
so-called general semantics (or Henkin semantics) in which a
meaningful notion of completeness can be
achieved~\parencite{Henkin50, Andrews72a}.
Note that standard models are subsumed by Henkin general models such
that valid HOL-formulas with respect to general semantics are also valid in the standard sense.

For the purposes of the present article, we shall omit the formal presentation of HOL semantics and of its proof system(s). We fix instead some useful notation for use in the remainder.  We write
${\Hk}\vDash^{\text{HOL}} \varphi$ if
formula $\varphi$ of HOL is \emph{true} in a Henkin general model ${\Hk}$; $\vDash^{\text{HOL}} \varphi$ denotes that $\varphi$ is (Henkin) \emph{valid}, i.e., that
${\Hk}\vDash^{\text{HOL}} \varphi$ for all Henkin models ${\Hk}$.

\section{Object Logic (L1) -- A Modal Logic of Preferences} \label{sec:Logic}

Adopting the \logikey\ methodology of \S\ref{sec:methodology} to
support the given formalisation challenge, the first question to be
addressed is: how to (initially) select the object logic at layer L1? The
chosen logic not only must be expressive enough to allow the encoding
of knowledge about the law (and the world), as required for the application domain (cf.~our case study in \S\ref{sec:Application}), but must also provide the means to
represent the kind of conditional value preferences featured in Lomfeld's theory (as described in \S\ref{sec:ValueTheory}).
Importantly, it must also enable the adequate
modelling of the notions of value aggregation and conflict, as featured in our legal DSL (discussed in \S\ref{sec:ValueOntology}).

Our initial choice has been the family of modal logics of preference
presented by \textcite{BenthemGR09}, which we abbreviate by \logic\ in
the remainder.  \logic\ has been put forward as a modal logic
framework for the formalisation of preferences which also allows for
the modelling of \textit{ceteris paribus} clauses in the sense of ``all
other things being equal''. This reading goes back to the seminal work
of von Wright in the early 1960's \parencite{vonWright1963logic}.\footnote{
For the purposes of the application scenarios studied later on \S\ref{sec:Application}, we have focused on \logic's basic modal preference language, not yet employing \textit{ceteris paribus} clauses. Nevertheless, we have provided a complete encoding and assessment of full \logic\ in the associated {\Isabelle} sources.}

\logic\ appears well suited for effective automation using the SSEs
approach, which has been an important selection criterion. This
judgment is based on good prior experience with SSEs of related
(monadic) modal logic frameworks \parencite{J21,J23}, whose semantics employs accessibility
relations between possible worlds/states, just as \logic\ does.
We note, however, that this choice of (a family of) object logics (\logic)
is just one out of a variety of logical systems which can be encoded as fragments of
HOL employing the \textit{shallow semantical embedding} approach; cf.~\textcite{J41}.
This  approach also
allows us `upgrade' our object logic whenever necessary. In
fact, we add quantifiers and conditionals to \logic\ in \S\ref{subsec:PLextensions}.
Moreover, we may consider combining \logic\ with other logics, e.g., with normal modal logics by the mechanisms of \textit{fusion} and \textit{product} \parencite{sep-logic-combining}, or, more generally, by \textit{algebraic fibring} \parencite[Ch.~2--3]{LogicCombining}.
This illustrates a central objective of the
\logikey\ approach, namely that the precise choice of a formalisation
logic (i.e., the \textit{object logic} at L1) is to be seen as a parameter.

In the subsections below we start by informally outlining the family of modal logics of preferences \logic\ (hence postponing their formal definition to an appendix \S\ref{app:PL}). We then discuss its embedding as a fragment of HOL using the SSE approach.
As for \S\ref{sec:hol}, the technically and
formally less interested reader may actually skip the content of these
subsections and get back later.

\subsection{The modal logic of preferences \logic} \label{sec:pl}

We sketch the syntax and semantics of \logic\, adapting
the description from \textcite{BenthemGR09} (we refer to the appendix \S\ref{app:PL} for more details). 

The formulas of \logic\
are inductively defined as follows (where $\texttt{p}$ ranges over a
set $\texttt{Prop}$ of propositional constant symbols):
$$\varphi, \psi ::= \; \texttt{p} \mid \varphi \boldsymbol{\wedge} \psi \mid \boldsymbol{\neg} \varphi \mid \Diamond^\preceq\varphi \mid \Diamond^\prec\varphi \mid \textbf{E}\varphi$$

As usual in modal logic, \textcite{BenthemGR09} give \logic\ a Kripke-style semantics, which models propositions as sets of states or `worlds'. \logic\ semantics employs a reflexive and transitive accessibility relation $\preceq$ (resp., its strict counterpart $\prec$) to define the modal operators in the usual way. This relation is called a \textit{betterness ordering} (between states or `worlds'). 

For the sake of self-containedness, we summarize below the semantics of \logic.

A preference model $\M$ is a triple
$\M = \langle \W, \preceq, \delta \rangle$ where: (i) $\W$ is a set of
worlds/states; (ii) $\preceq$ is a \textit{betterness relation} (reflexive and transitive) on $\W$, where its strict subrelation $\prec$ is defined as: $w\prec v$ :=
$w\preceq v \land v\not\preceq w$ for all $v, w \in \W$ (totality of $\preceq$, i.e., $v\preceq w \lor w\preceq v$, is generally not assumed); (iii) $\delta$ is a standard modal valuation.
Below we show the truth conditions for \logic's modal
connectives (the rest being standard):
\begin{align*}
\M,w & \vDash\Diamond^\preceq\varphi \text{ iff } \exists v \in \W \text{ such that } w \preceq v \text{ and } \M,v \vDash \varphi \\
\M,w & \vDash\Diamond^\prec\varphi \text{ iff } \exists v \in \W \text{ such that } w \prec v \text{ and } \M,v \vDash \varphi \\
\M,w & \vDash\textbf{E}\varphi\ \ \ \text{ iff } \exists v \in \W \text{ such that } \M,v \vDash \varphi 
\end{align*}

A formula $\varphi$ is \textit{true at} world $w \in W$ in model
$\M$ if ${\M,w\vDash\varphi}$. $\varphi$ is
\textit{globally true in} $\M$, denoted ${\M\vDash\varphi}$, if $\varphi$ is
\textit{true at} every $w \in W$. Moreover, $\varphi$ is \textit{valid} (in a class
of models $\K$) if \textit{globally true in} every $\M$ ($\in \K$),
denoted ${\vDash_{\mathcal{PL}}\varphi}$ (${\vDash_{\K}\varphi}$).

Thus, $\Diamond^\preceq\varphi$ (resp.,
$\Diamond^\prec\varphi$) can informally be read as ``$\varphi$ is true in a
state that is considered to be at least as
good as (resp., strictly better than) the current state'' and $\textbf{E}\varphi$ can be read as
``there is a state where $\varphi$ is true''.

Further, standard connectives such as $\boldsymbol\vee$,
$\boldsymbol\rightarrow$ and $\boldsymbol\leftrightarrow$ can also
defined in the usual way.
The dual operators $\Box^\preceq\varphi$ (resp., $\Box^\prec\varphi$) and
$\textbf{A}\varphi$ can also be defined as
$\boldsymbol\neg\Diamond^\preceq\boldsymbol\neg\varphi$ (resp., $\boldsymbol\neg\Diamond^\prec\boldsymbol\neg\varphi$) and
$\boldsymbol\neg\textbf{E}\boldsymbol\neg\varphi$.

Readers acquainted with Kripke semantics for modal logic will notice that \logic\ features normal \textit{S4} and \textit{K4} diamonds operators $\Diamond^\preceq$ and $\Diamond^\prec$, together with a
global existential modality \textbf{E}. We can thus give the usual reading to $\Box$ and $\Diamond$ as \textit{necessity} and \textit{possibility}, respectively. 

Moreover, note that, since the \textit{strict} betterness relation $\prec$ is
not reflexive, it does not hold in general that
${\Box^\prec\varphi\boldsymbol\rightarrow\varphi}$ (modal axiom
$T$). Hence we can also give a `deontic reading' to $\Diamond^\prec\varphi$
and $\Box^\prec\varphi$; the former could then read as ``$\varphi$ is
admissible/permissible'' and the latter as ``$\varphi$ is
recommended/obligatory''. This deontic interpretation can be further
strengthened so that the latter entails the former by extending
\logic\ with the postulate
${\Box^\prec\varphi\boldsymbol\rightarrow\Diamond^\prec\varphi}$
(modal axiom $D$), or alternatively, by postulating the corresponding
(meta-logical) semantic condition, namely, that for each state there
exists a strictly better one (\textit{seriality} for $\prec$).

Observe that we use \textbf{boldface} fonts to distinguish standard logical connectives of \logic\ from their counterparts in HOL.

Similarly, eight different binary connectives
for modelling preference statements between propositions can
be defined in \logic:
\begin{align*}
{\preceq_{EE}}/{\prec_{EE}}\;,\; {\preceq_{EA}}/{\prec_{EA}}\;,\;
  {\preceq_{AE}}/{\prec_{AE}}\;,\; {\preceq_{AA}}/{\prec_{AA}}\,.
\end{align*}
These connectives arise from four different ways of `lifting' the
\textit{betterness ordering} $\preceq$ (resp., $\prec$) on states to a
\textit{preference ordering} on sets of states or propositions.
\begin{align*}
  (\varphi\ {\preceq}_{EE}/{\prec}_{EE}\ \psi)\ u & \text{\; iff\; }  
  \exists t\ \varphi\,s \wedge (\exists t\ \psi\,t \wedge s\ {\preceq}/{\prec}\ t) \\
  (\varphi\ {\preceq}_{EA}/{\prec}_{EA}\ \psi)\ u  & \text{\; iff\; }  
  \exists t\ \psi\,t \wedge (\forall s\ \varphi\,s \rightarrow s\ {\preceq}/{\prec}\ t) \\
  (\varphi\ {\preceq}_{AE}/{\prec}_{AE}\ \psi)\ u & \text{\; iff\; }  
  \forall s\ \varphi\,s \rightarrow (\exists t\ \psi\,t \wedge s\ {\preceq}/{\prec}\ t) \\
  (\varphi\ {\preceq}_{AA}/{\prec}_{AA}\ \psi)\ u & \text{\; iff\; }  
  \forall s\ \varphi\,s \rightarrow  (\forall t\ \psi\,t \rightarrow s\ {\preceq}/{\prec}\ t)
\end{align*}
Thus, different choices for a \textit{logic of preference} are possible if we  restrict ourselves to employing only a selected preference connective, where each choice provides the logic with particular characteristics, so that we can interpret preference statements between propositions (i.e., sets of states) in a variety of ways. As an illustration, according to the semantic interpretation provided by \textcite{BenthemGR09}, 
we can read $\varphi{\prec}_{AA}\psi$ as “every $\psi$-state being better than every
$\varphi$-state”, and read $\varphi{\prec}_{AE}\psi$ as “every $\varphi$-state having a better $\psi$-state” (and analogously for others).

In fact, the family of preference logics \logic\ can be seen as
encompassing, in substance, the proposals
by \textcite{vonWright1963logic} (variant ${\prec}_{AA}$) and
\textcite{halpern1997defining} (variants ${\preceq}_{AE}/ {\prec}_{AE}$).\footnote{Von Wright's proposal is discussed in some detail in \textcite{BenthemGR09}; cf.~also
  \textcite{Liu2008} for a discussion of further proposals.}
As we will see later in \S\ref{sec:ValueOntology}, there are only four choices (${\preceq}_{EA}/ {\prec}_{EA}$
and ${\preceq}_{AE}/ {\prec}_{AE}$) of modal preference relations that
satisfy the minimal conditions we impose for a logic of value
aggregation. Moreover, they are the only ones which satisfy
transitivity, a quite controversial property in the literature on
preferences.

Last but not least, \textcite{BenthemGR09} have provided `syntactic' counterparts for these binary preference connectives as derived operators in the
language of \logic\ (i.e., defined by employing the modal operators $\Diamond^\preceq\varphi$ (resp.,
$\Diamond^\prec\varphi$).
We note these `syntactic variants' in \textbf{boldface} font:
\begin{align*}
  (\varphi \boldsymbol{\preceq}_{EE} \psi) &:= \boldsymbol{E}(\varphi \boldsymbol{\wedge} \Diamond^\preceq\psi) & \text{and\;\;\; } (\varphi \boldsymbol{\prec}_{EE} \psi) &:=                      \boldsymbol{E}(\varphi \boldsymbol{\wedge} \Diamond^\prec\psi) \\
  (\varphi \boldsymbol{\preceq}_{EA} \psi) &:=  \boldsymbol{E}(\psi \boldsymbol{\wedge} \Box^\prec\boldsymbol{\neg}\varphi) & \text{and\;\;\; } (\varphi \boldsymbol{\prec}_{EA} \psi) &:=   \boldsymbol{E}(\psi \boldsymbol{\wedge} \Box^\preceq\boldsymbol{\neg}\varphi)\\
  (\varphi \boldsymbol{\preceq}_{AE} \psi) &:=  \boldsymbol{A}(\varphi \boldsymbol{\rightarrow} \Diamond^\preceq\psi) & \text{and\;\;\; } (\varphi \boldsymbol{\prec}_{AE} \psi) &:=                            \boldsymbol{A}(\varphi \boldsymbol{\rightarrow} \Diamond^\prec\psi) \\
  (\varphi \boldsymbol{\preceq}_{AA} \psi) &:=  \boldsymbol{A}(\psi \boldsymbol{\rightarrow} \Box^\prec\boldsymbol{\neg}\varphi) & \text{and\;\;\; } (\varphi \boldsymbol{\prec}_{AA} \psi) &:=              \boldsymbol{A}(\psi \boldsymbol{\rightarrow} \Box^\preceq\boldsymbol{\neg}\varphi_\sigma)
\end{align*}
The relationship between both, i.e., the semantically and
syntactically defined families of binary preference connectives is
discussed in \textcite{BenthemGR09}. In a nutshell, as regards the
\textit{EE}- and the \textit{AE}-variants, both definitions (syntactic
and semantic) are equivalent; concerning the \textit{EA}- and the
\textit{AA}-variants, the equivalence only holds for a total $\preceq$
relation. In fact, drawing upon our encoding of \logic\ as presented in the next subsection \S\ref{sec:Implementation}, we have employed {\Isabelle} for automatically verifying this sort of meta-theoretic correspondences; cf.~Lines 4--12
in Fig.~\ref{fig:MetaTheory1} in Appx.~\ref{subsec:A2}.

\subsection{Embedding \logic\ in HOL} \label{sec:Implementation}

For the implementation of \logic\ we utilise the \textit{shallow
  semantical embeddings} (SSE) technique, which encodes the language
constituents of an object logic, \logic\ in our case, as expressions
($\lambda$-terms) in HOL. A core idea is to model (relevant parts of)
the semantical structures of the object logic explicitly in HOL. This
essentially shows that the object logic can be unraveled as a fragment
of HOL and hence automated as such. For (multi-)modal normal logics,
like \logic, the relevant semantical structures are relational frames
constituted by sets of possible worlds/states and their accessibility
relations. \logic\ formulas can thus be encoded as predicates in HOL
taking possible worlds/states as arguments.\footnote{This corresponds
  to the well-known standard translation to first-order
  logic. Observe, however, that the additional expressivity of HOL
  allows us to also encode and flexibly combine non-normal modal
  logics (conditional, deontic, etc.; cf.~\cite{J31,J46,J45,B22}) and
  we can elegantly add quantifiers (cf.~\S\ref{subsec:PLextensions}).}
The detailed SSE of the basic operators of \logic\ in HOL is presented
and formally tested in Appx.~\ref{app:PL}. Further extensions to support
reasoning with \textit{ceteris paribus} clauses in \logic\ are studied
there as well.

As a result, we obtain a combined, interactive and automated, theorem
prover and model finder for \logic\ (and its extensions; cf.~\S\ref{subsec:PLextensions}) realised
within {\Isabelle}. This is a new contribution, since we are not aware
of any other existing implementation and automation of such a
logic. Moreover, as we will demonstrate below, the SSE technique
supports the automated assessment of meta-logical properties of the
embedded logic in {\Isabelle}, which in turn provides practical
evidence for the correctness of our encoding.

The embedding starts out with declaring the HOL base type $\itype$,
which is denoting the set of possible states (or worlds) in our
formalisation.  \logic\ propositions are modelled as predicates on
objects of type $\itype$ (i.e., as \textit{truth-sets} of
states/worlds) and, hence, they are given the type $\itype\ar o$,
which is abbreviated as $\sigma$ in the remainder.  The
\textit{betterness relation} $\preceq$ of \logic\ is introduced as an
uninterpreted constant symbol $\preceq_{\itype\ar \itype \ar o}$ in
HOL, and its strict variant $\prec$ is introduced as an abbreviation
$\prec_{\itype\ar \itype \ar o}$ standing for the HOL term
$\lambda v.\,\lambda w.\,(v\leq w \wedge \neg(w\leq
v))$.
In accordance with \textcite{BenthemGR09}, we postulate
that $\preceq$ is a preorder, i.e., reflexive and transitive.

In a next step, the $\sigma$-type lifted logical connectives of \logic\ are
introduced as abbreviations for $\lambda$-terms in the meta-logic
HOL. The conjunction operator $\boldsymbol{\wedge}$ of \logic, for
example, is introduced as an abbreviation
${\boldsymbol{\wedge}_{\sigma\ar\sigma\ar \sigma}}$, which stands for
the HOL term
${\lambda \varphi_\sigma. \lambda \psi_\sigma. \lambda w_\itype.\,
  (\varphi~w~\wedge~\psi~w)}$, so that
${\varphi_\sigma\boldsymbol{\wedge}\psi_\sigma}$ reduces to
${\lambda w_\itype.\, (\varphi~w~\wedge~\psi~w)}$, denoting the
set\footnote{In HOL (with Henkin semantics) sets are associated with
  their characteristic functions.} of all possible states $w$ in which
both $\varphi$ and $\psi$ hold. Analogously, for the negation we
introduce an abbreviation ${\boldsymbol{\neg}_{\sigma\ar\sigma}}$
which stands for
${\lambda \varphi_\sigma. \lambda w_\itype.\, \neg(\varphi~w)}$.

\sloppy The operators $\Diamond^\preceq$ and $\Diamond^\prec$ use
$\preceq$ and $\prec$ as guards in their definitions.  These operators
are introduced as shorthand $\Diamond^\preceq_{\sigma\ar\sigma}$ and
$\Diamond^\prec_{\sigma\ar\sigma}$ abbreviating the HOL terms
${\lambda \varphi_\sigma. \lambda w_\itype.\, \exists v_\itype (w\preceq v
  \wedge \varphi~v)}$ and
${\lambda \varphi_\sigma. \lambda w_\itype.\, \exists v_\itype (w\prec v
  \wedge \varphi~v)}$, respectively.
$\Diamond^\preceq_{\sigma\ar \sigma}\varphi_\sigma$ thus reduces to
${\lambda w_\itype.\, \exists v_\itype (w\preceq v \wedge \varphi~v)}$,
denoting the set of all worlds $w$ so that $\varphi$ holds in some
world $v$ that is at least as good as $w$; analogously for
$\Diamond^\prec_{\sigma\ar \sigma}$. Additionally, the \textit{global
  existential} modality $\textbf{E}_{\sigma\ar\sigma}$ is introduced
as shorthand for the HOL term
${\lambda \varphi_\sigma. \lambda w_\itype.\, \exists v_\itype
  (\varphi~v)}$. The duals
$\Box^\preceq_{\sigma\ar \sigma}\varphi_\sigma$,
$\Box^\prec_{\sigma\ar \sigma}\varphi_\sigma$ and
$\textbf{A}_{\sigma\ar\sigma}\varphi$ can easily be added so that they
are equivalent to
$\boldsymbol{\neg}\Diamond^\preceq_{\sigma\ar
  \sigma}\boldsymbol{\neg}\varphi_\sigma$,
$\boldsymbol{\neg}\Diamond^\prec_{\sigma\ar
  \sigma}\boldsymbol{\neg}\varphi_\sigma$ and
$\boldsymbol{\neg}\textbf{E}_{\sigma\ar\sigma}\boldsymbol{\neg}\varphi$
respectively.

Moreover, a special predicate $\lfloor \varphi_\sigma \rfloor$ (read
$\varphi_\sigma$ is \textit{valid}) for $\sigma$-type lifted \logic\
formulas in HOL is defined as an abbreviation for the HOL term
${\forall w_\itype (\varphi_\sigma\,w)}$.

The encoding of object logic \logic\ in meta-logic HOL is presented in
full detail in Appendix \ref{app:PL}.

Remember again that in the \logikey\ methodology the modeler is not
enforced to make an irreversible selection of an object logic (L1) before
proceeding with the formalisation work at higher \logikey\
layers. Instead the framework enables preliminary choices at all
layers which can easily be revised by the modeler later on if this is
indicated by e.g. practical experiments.

\subsection{Formally Verifying Encoding's Adequacy} \label{sec:faithful}

A pen-and-paper proof of the faithfulness (soundness \& completeness) of the SSE easily follows from previous results regarding the SSE of propositional multi-modal logics \parencite{J21} and their quantified extensions \parencite{J23}; cf.~also \textcite{J41} and the references therein. We sketch such an argument below, as it provides an insight into the underpinnings of SSE for the interested reader.

By drawing upon the approach in \textcite{J21}, it is possible to
define a mapping between semantic structures of the object logic
\logic\ (preference models $\M$) and the corresponding structures in HOL (general Henkin models ${\Hk}^\M$), 
in such a way that 
$${\vDash^{\text{HOL}(\Gamma)}\lfloor\varphi_\sigma \rfloor} \;\;\;\;\text{iff}\;\;\;\;
{\vDash_{\mathcal{PL}}\varphi} \;\;\;\;\text{iff} \;\;\;\;
\vdash_{\mathcal{PL}}\varphi,$$
where $\vdash_{\mathcal{PL}}$ denotes derivability in the (complete) calculus axiomatised by \textcite{BenthemGR09}.
Observe that HOL($\Gamma$) corresponds to HOL extended with the relevant types and constants plus a set $\Gamma$ of axioms encoding \logic\ semantic conditions, e.g., reflexivity and transitivity of ${\preceq_{\itype\ar \itype \ar o}}$.

Soundness of the SSE (i.e., ${\vDash^{\text{HOL}(\Gamma)}\lfloor\varphi_\sigma \rfloor}$
implies ${\vDash_{\mathcal{PL}}\varphi}$) is particularly important since it ensures that our modelling does not give any `false positives', i.e., proofs of \logic\ non-theorems.  
Completeness of the SSE (i.e., ${\vDash_{\mathcal{PL}}\varphi}$ implies ${\vDash^{\text{HOL}(\Gamma)}\lfloor\varphi_\sigma \rfloor}$) means that our modelling does not give any `false negatives', i.e., spurious counterexamples. Besides the pen-and-paper proof, completeness can also be mechanically verified by deriving the $\sigma$-type lifted \logic\ axioms and inference rules in HOL($\Gamma$); cf.~Fig.~\ref{fig:MetaTheory1} and Fig.~\ref{fig:MetaTheory2} in
Appx.~\ref{app:PL}.

To gain practical evidence for the faithfulness of our SSE of \logic\
in {\Isabelle}, and also to assess proof automation performance, we
have conducted numerous experiments in which we automatically
verify meta-theoretical results on \logic\ as presented by \textcite{BenthemGR09}. Note that these statements thus play a role analogous to that of a requirements specification document (cf.~Fig.~\ref{fig:MetaTheory1} and Fig.~\ref{fig:MetaTheory2} in
Appx.~\ref{subsec:A2}).

\subsection{Beyond \logic: Extending the Encoding with Quantifiers and Conditionals}
\label{subsec:PLextensions}
We can further extend our encoded logic \logic\ by adding quantifiers. This
is done by identifying $\boldsymbol{\forall} x_\alpha s_\sigma$
with the HOL term $\lambda w_\itype. \forall x_\alpha (s_\sigma w)$ and
$\boldsymbol{\exists} x_\alpha s_\sigma$ with
$\lambda w_\itype. \exists x_\alpha (s_\sigma w)$; cf.~\textit{binder notation} in \S\ref{sec:hol}.  This way quantified
expressions can be seamlessly employed in our modelling at upper
layers (as done exemplarily in \S\ref{sec:Application}). We refer the reader to
\textcite{J23} for a more detailed discussion (including faithfulness
proofs) of SSEs for \textit{quantified} (multi-)modal logics.

Moreover, observe that having a semantics based on
\textit{preferential structures} allows us to extend our logic with a (defeasible) conditional connective $\Rightarrow$. This can be
done in several closely related ways. As an illustration, drawing upon an approach by \textcite{boutilier1994toward}, we can further extend the SSE of \logic\ by defining the connective:
\begin{align*}
  \varphi_\sigma \Rightarrow \psi_\sigma := \boldsymbol{A}(\varphi_\sigma \rightarrow \Diamond^\preceq(\varphi_\sigma \wedge \Box^\preceq(\varphi_\sigma \rightarrow \psi_\sigma))).
\end{align*}
An intuitive reading of this conditional statement would be: ``every
$\varphi$-state has a reachable $\varphi$-state such that $\psi$ holds
there in also in every reachable $\varphi$-state'' (where we can
interpret ``reachable'' as ``at least as good''). This is equivalent,
for finite models, to demanding that all `best' $\varphi$-states are
$\psi$-states, cf.~\textcite{lewis1973counterfactuals}.
This can indeed be shown equivalent to the approach by \textcite{halpern1997defining}, who axiomatises a strict binary preference relation $\boldsymbol{\succ^s}$, interpreted as ``relative likelihood''.\footnote{In fact, \textcite{halpern1997defining} variant corresponds to employing the preference relation $\prec_{AE}$ discussed previously, augmented with an additional constraint to cope with infinite-sized countermodels to irreflexivity (building upon an approach by \textcite{lewis1973counterfactuals}). Thus, $\psi \boldsymbol{\succ^s} \varphi$ (read: $\psi$ is more likely than $\varphi$) iff every $\varphi$-state has a more likely $\psi$-state, say $v$, which \textit{dominates} $\varphi$ (i.e., no $\varphi$-state is more likely than $v$). \textcite{halpern1997defining}  goes on to define a conditional operator as follows: ${\varphi \Rightarrow \psi :=~\boldsymbol{A}\neg\varphi \vee ((\varphi \wedge \psi) \boldsymbol{\succ^s} (\varphi \wedge \neg\psi))}$.}
For further discussion regarding the properties and applications of this --and other similar--
preference-based conditionals we refer the interested reader to the
discussions in \textcite{Benthem2009} and \textcite[Ch.~3]{Liu2011}.

\section{Domain Specific Language (L2) -- Value-Oriented Legal Theory}\label{sec:ValueOntology}

In this section we incrementally define a domain-specific language (DSL) for reasoning with values in a legal context. We start by defining a ``logic of value preferences'' on top of the object logic \logic\ (layer L1). This logic is subsequently encoded in {\Isabelle}, and in the process it becomes suitably extended with custom means to encode the \textit{discoursive grammar} in \S\ref{sec:ValueTheory}. We thus obtain a HOL-based DSL formally modelling Lomfeld's theory. This formally-verifiable DSL is then put to the test using theorem provers and model generators.

Recall from the discussion of the \textit{discoursive grammar} in \S\ref{sec:ValueTheory} that value-oriented legal rules can become expressed as context-dependent preference statements between \textit{value principles} (e.g.~\textit{RELIance}, \textit{STABility}, \textit{WILL}, etc.).
Moreover, these value principles were informally associated to basic \textit{values}. (i.e., \textit{FREEDOM}, \textit{UTILITY}, \textit{SECURITY} and \textit{EQUALITY}),
in such a way as to arrange the first over (the quadrants of) a plane generated by two axes labelled by the latter. More specifically, each axis' pole is labelled by a basic value, with values lying at contrary poles playing somehow antagonistic roles (e.g.~\textit{FREEDOM} vs.~\textit{SECURITY}). We recall the corresponding diagram (Fig.~\ref{fig:ValueOntology}) below for the sake of illustration:

\begin{figure}[!htb]
  \centering
  \includegraphics[width=.7\textwidth]{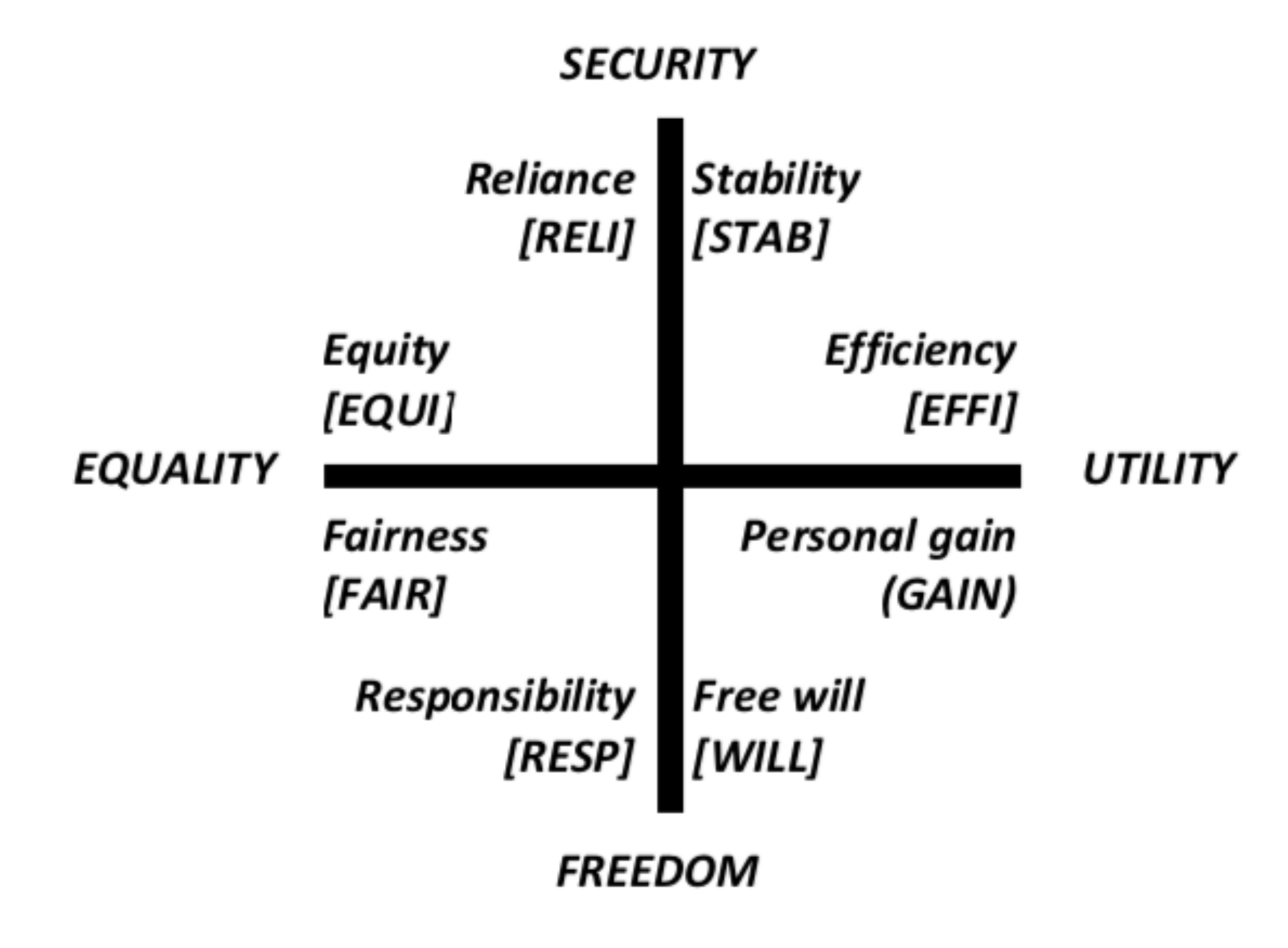}
  \caption{Basic legal value system (ontology) by \textcite{lomfeld19_grammatik}}
  \label{fig:ValueOntology-bis}
\end{figure}

Inspired by this theory, we model the notion of a (value) \textit{principle}
as consisting of a collection (in this case a set\footnote{\label{footnote-modelling}Observe that in doing so we are simplifying Lomfeld's value theory to the effect that, e.g., STAB\textit{ility} becomes identified with EFFI\textit{ciency}. This simplified model has proven sufficient for our modelling work in \S\ref{sec:Application}. A more granular encoding of principles is possible by adding a third axis to the value space in Fig.~\ref{fig:ValueOntology-bis}, thus allocating each principle to its own octant.}) of \textit{base values}. Thus, by considering principles as structured entities, we can more easily define adequate notions of aggregation and conflict among them; cf.~\S\ref{sec:ValueOntology}.

From a logical point of view it is additionally required to conceive value principles as truth-bearers, i.e., propositions.\footnote{We recall that, from a modal logic perspective, a proposition is modelled as the set of `worlds' (i.e., states or situations) in which it holds. Informally, we want to be able to express the fact that a given principle, say legal STABility, is being observed or respected in a particular situation, or, abusing modal logic jargon, that the principle is `satisfied' in that `world'. This can become further interpreted as providing a \textit{justification} for why that world or situation is desirable.}
We thus seem to face a dichotomy between, at the same time, modelling value principles as sets of basic values and modelling them as sets of worlds.
In order to adequately tackle this modelling challenge we make use of the mathematical notion of a \textit{Galois connection}.\footnote{An old mathematician's trick has been to employ --maybe unknowingly-- Galois connections (resp.~adjunctions) to relate two universes of mathematical objects with each other, in such a way that certain order structures become inverted (resp.~preserved).
In doing so, insights and results can be transferred from a well-known universe towards a less-known one, in order to gain information and help illuminate difficult problems; cf.~the discussion in \textcite{Erne2004}.}

For the sake of exposition, Galois connections are to be exemplified by the notion of \textit{derivation operators} in the theory of Formal Concept Analysis (FCA), from which we took inspiration; cf.~\textcite{ganter2012formal}. FCA is a mathematical theory of concepts and concept hierarchies as formal ontologies, which finds practical application in many computer science fields such as data mining, machine learning, knowledge engineering, semantic web, etc.\footnote{In particular, we want to highlight the potential of employing the powerful FCA methods, e.g.,~\textit{attribute exploration} \parencite{ganter2016conceptual}, to prospective `legal value mining' applications.}

\subsection{{Some Basic FCA Notions}}

A \textit{formal context} is a triple $K=\langle G,M,I\rangle$ where
$G$ is a set of \textit{objects}, $M$ is a set of \textit{attributes},
and $I$ is a relation between $G$ and $M$ (usually called
\textit{incidence relation}), i.e., $I \subseteq G \times M$. We read
$\langle g,m\rangle \in I$ as ``the object $g$ has the attribute
$m$''. Additionally we define two so-called \textit{derivation
  operators} $\uparrow$ and $\downarrow$ as follows:
\begin{align}
  A{\uparrow}~:=&~\{m \in M~|~\langle g,m \rangle \in I~\text{for all}~g\in A\}&\text{for~} A \subseteq G \label{form:derivation-ops1}\\
  B{\downarrow}~:=&~\{g \in G~~|~\langle g,m \rangle \in I~\text{for all}~m\in B\}&\text{for~} B \subseteq M \label{form:derivation-ops2}
\end{align}

$A{\uparrow}$ is the set of all attributes shared by all objects from
$A$, which we call the \textit{intent} of $A$.  Dually,
$B{\downarrow}$ is the set of all objects sharing all attributes from
$B$, which we call the \textit{extent} of $B$.  This pair of
derivation operators thus forms an antitone \textit{Galois connection}
between (the powersets of) $G$ and $M$, and we always have that
$B \subseteq A{\uparrow}$ iff $A \subseteq B{\downarrow}$.

\begin{figure}[!htb]
  \centering
  \includegraphics[width=.6\textwidth]{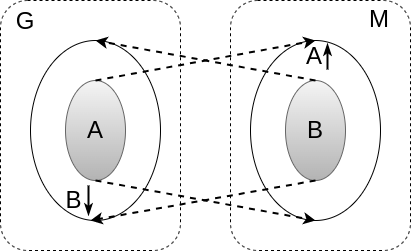}
  \caption{A suggestive representation of a Galois connection between a set of objects $G$ (e.g.~worlds) and set of their attributes $M$ (e.g.~values).}
  \label{fig:GaloisConnections}
\end{figure}

A \textit{formal concept} (in a context $K$) is defined as a pair
$\langle A,B\rangle$ such that $A \subseteq G$, $B \subseteq M$,
$A{\uparrow} = B$, and $B{\downarrow} = A$. We call $A$ and $B$ the
\textit{extent} and the \textit{intent} of the concept
$\langle A,B\rangle$, respectively.\footnote{The terms \textit{extent}
  and \textit{intent} are reminiscent of the philosophical notions of
  \textit{extension} and \textit{intension (comprehension)} reaching
  back to the 17th century \textit{Logique de Port-Royal}.} Indeed
$\langle A{\uparrow}{\downarrow},A{\uparrow}\rangle$ and
$\langle B{\downarrow},B{\downarrow}{\uparrow}\rangle$ are always
concepts.

The set of concepts in a formal context is partially ordered by set
inclusion of their extents, or, dually, by the (reversing) inclusion
of their intents. In fact, for a given formal context this ordering
forms a complete lattice: its \textit{concept lattice}. Conversely, it
can be shown that every complete lattice is isomorphic to the concept
lattice of some formal context.  We can thus define
lattice-theoretical meet and join operations on FCA concepts in order
to obtain an algebra of concepts:\footnote{This result can be
  seamlessly stated for infinite meets and joins (infima and suprema)
  in the usual way. It corresponds to the first part of the so-called
  \textit{basic theorem on concept lattices}
  \parencite{ganter2012formal}.}
\begin{align}
  \langle A_1,B_1\rangle \wedge \langle A_2,B_2\rangle &:= \langle (A_1 \cap A_2)~, (B_1 \cup B_2){\downarrow}{\uparrow}\rangle \\
  \langle A_1,B_1\rangle \vee \langle A_2,B_2\rangle &:= \langle (A_1 \cup A_2){\uparrow}{\downarrow}~, (B_1 \cap B_2)\rangle \label{form:fca-join}
\end{align}

\subsection{A Logic of Value Preferences}\label{subsec:LogicValuePreferences}

In order to enable the modelling of Lomfeld's legal theory as discussed in \S\ref{sec:ValueTheory}, we will enhance our object logic \logic\ with additional expressive means by drawing upon the FCA notions expounded above, and by assuming an \textit{arbitrary} domain set $\V$ of basic values.

A first step towards our legal DSL is to define a pair of operators ${\uparrow}$ and ${\downarrow}$ such that they form a Galois connection between the semantic domain $\W$ of worlds/states of \logic\ (as `objects' $G$) and the set of basic values $\V$ (as `attributes' $M$). By employing the operators ${\uparrow}$ and ${\downarrow}$ in an appropriate way, we can obtain additional well-formed \logic\ terms, thus converting our object logic \logic\ in a logic of value preferences. Details follow.

\subsubsection*{Principles, Values and Propositions}

We introduce a \textit{formal context} $K=\langle \W,\V,\I\rangle$ composed by the set of worlds $\W$, the set of basic values $\V$, and the (implicit) relation $\I \subseteq \W \times \V$, which we might interpret, intuitively, in a teleological sense: $\langle w,v \rangle \in \I$ means that value $v$ provides reasons for the situation (world/state) $w$ to obtain.

Now, recall that we aim at modelling value principles as sets
of basic values (i.e., elements of $2^\V$), while, at the same time, conceiving of them as
propositions (elements of $2^\W$).  Indeed, drawing upon the above FCA notions allows us to
overcome this dichotomy. Given the formal context $K=\langle \W,\V,\I\rangle$ we can define the pair of derivation operators $\uparrow$ and $\downarrow$
employing the corresponding definitions
(\ref{form:derivation-ops1}-\ref{form:derivation-ops2}) above.

We can now employ these derivation operators to switch between the
`(value) principles as sets of (basic) values' and the `principles as
propositions (sets of worlds)' perspectives. Hence, we can now
--recalling the informal discussion of the semantics of the object logic \logic\ in \S\ref{sec:Logic} -- give an intuitive reading for truth at
a world in a preference model to terms of the form $P{\downarrow}$;
namely, we can read $\M,w \vDash P{\downarrow}$ as ``principle $P$
provides a reason for (state of affairs) $w$ to obtain''. In the same
vein, we can read $\M \vDash A \rightarrow P{\downarrow}$ as
``principle $P$ provides a reason for proposition $A$ being the
case''.\footnote{Observe that this can be written semi-formally as: \textit{for all} $w$ \textit{in} $\M$ \textit{we have that} \textit{if} $\M,w \vDash A$ \textit{then}  $\M,w \vDash P{\downarrow}$, which can be interpreted as ``$P$ provides a reason for all those worlds that satisfy $A$''.}

\subsubsection*{Value Aggregation}

Recalling Lomfeld's theory, as discussed in \S\ref{sec:ValueTheory},
our logic of value preferences must provide means for expressing conditional preferences between value principles, according to the schema:
\begin{align*}
  E_1 \;\land\; \dots \;\land\; E_n  \;\; \Rightarrow \;\; (A_1 \oplus\dots\oplus A_n) \prec (B_1 \oplus\dots\oplus B_n)
\end{align*}

As regards the preference relation (connective $\prec$), we might think that, in principle, any choice among the eight preference
relation variants in \logic\ (cf.~\S\ref{sec:Logic}) will work.
Let us recall, however, that Lomfeld's theory also presupposed some (no further specified) mechanism for aggregating value principles (operator $\oplus$); thus, the joint selection of both a preference relation and a aggregation operator cannot be arbitrary: they need to interact in an appropriate way. We
explore first a suitable mechanism for value aggregation before we get back to this issue.

Suppose that, for example, we are interested in modelling a legal case in which, say, the principle of ``respect for property'' \textit{together with} the principle ``economic benefit for society'' \textit{outweighs}
the principle of ``legal certainty''.\footnote{Employing Lomfeld's value theory this corresponds to RELIance together with personal GAIN outweighing STABility.}
A binary connective $\oplus$ for modelling this notion of \textit{together with}, i.e., for aggregating legal principles (as reasons) must, expectedly, satisfy particular logical constraints in interaction with a (suitably selected) value preference relation $\prec$:
\begin{align*}
(A \prec B) \rightarrow (A \prec B\oplus C) &\text{~~but not~~} (A \prec B\oplus C) \rightarrow (A \prec B) & \text{right aggregation} \\
(A\oplus C \prec B) \rightarrow (A \prec B) &\text{~~but not ~~} (A \prec B) \rightarrow (A\oplus C \prec B) & \text{left aggregation} \\
(B \prec A) & \wedge (C \prec A) \rightarrow (B\oplus C \prec A) & \text{union property (opt.)}
\end{align*}

For our purposes, the aggregation connectives are most conveniently defined using set union (FCA join), which gives us commutativity.
As it happens, only the ${\prec}_{AE}/{\preceq_{AE}}$ and
${\prec}_{EA}/{\preceq_{EA}}$ variants from \S\ref{sec:Logic} 
satisfy the first two conditions.
They are also the only variants satisfying transitivity.
Moreover, if we choose to enforce the optional third aggregation principle (called ``union property''; cf.~\textcite{halpern1997defining}),
then we would be left with only one variant to consider, namely ${\prec}_{AE}/{\preceq_{AE}}$.\footnote{Lacking any strong opinion regarding the correctness of transitivity or the union property, we have still chosen this latter variant for our case study in \S\ref{sec:Application}, since it offers several benefits for our current
modelling purposes: it can be faithfully encoded in the language of
\logic\ \parencite{BenthemGR09} and its behaviour is well documented in the literature; cf.~\textcite{halpern1997defining}, \textcite[Ch.~4]{Liu2008}.
In fact, as mentioned in
\S\ref{subsec:PLextensions}, drawing upon the strict variant
$\prec_{AE}$ we can even define a defeasible conditional $\Rightarrow$
in \logic.}

In the end, after extensive computer-supported experiments in
\Isabelle\ we have identified the following candidate definitions for the
value aggregation and preference connectives which satisfy our modelling
desiderata:\footnote{Respective tests are presented in
  Figs.~\ref{fig:MetaTheoryApp1}--\ref{fig:MetaTheoryApp2} in
  Appx.~\ref{app:PL}.} 
\begin{itemize}
\item For the binary value aggregation connective $\oplus$ we have
  identified the following two candidates (both taking two value
  principles and returning a proposition):
  \begin{align*}
    A \oplus_{(1)} B :=&~(A \cap B){\downarrow} \\
    A \oplus_{(2)} B :=&~(A{\downarrow} \vee B{\downarrow})
  \end{align*}
  Observe that $\oplus_1$ is based upon the join operation on the
  corresponding FCA formal concepts (see
  Def.~\ref{form:fca-join}). $\oplus_2$ is a strengthening of the
  first, since $(A \oplus_2 B) \subseteq (A \oplus_1 B)$.

\item For a binary preference connective $\prec$ between propositions
  we have as candidates:
  \begin{align*}
    \varphi \prec_{(1)} \psi := \varphi \preceq_{AE} \psi \\
    \varphi \prec_{(2)} \psi := \varphi \prec_{AE} \psi \\
    \varphi \prec_{(3)} \psi := \varphi \preceq_{EA} \psi \\
    \varphi \prec_{(4)} \psi := \varphi \prec_{EA} \psi
  \end{align*}
\end{itemize}

In line with the \logikey\ methodology, we consider the concrete
choices of definitions for $\prec$, $\oplus$, and even $\Rightarrow$
(classical or defeasible) as parameters in our overall modelling
process. No particular determination is enforced in the \logikey\
approach, and we may alter any preliminary choices as soon as this
appears appropriate. In this spirit we experimented with the listed
different definition candidates for our connectives and explored their
behaviour. We will present our final selection in
\S\ref{subsec:EncodingValueOntology}.

\subsubsection*{Promoting Values}

Given that we aim at providing a logic of value preferences for use in legal reasoning, we
still need to consider the mechanism by which we can link legal
decisions, together with other legally relevant facts, 
to values. We conceive of such a mechanism as
a sentence schema, which reads intuitively as: ``Taking decision $D$ in
the presence of facts $F$ \textit{promotes} (value) principle $P$''.  The
formalisation of this schema can indeed be seen as a new predicate in the domain-specific language (DSL) that we have been gradually defining 
in this section. In the expression \textit{Promotes(F,D,P)} we have that $F$ is a conjunction of
facts relevant to the case (a proposition), $D$ is the legal decision,
and $P$ is the value principle thereby promoted.\footnote{We adopt the
  terminology of \textit{promoting} (or \textit{advancing}) a value from
  the literature
  \parencite{berman_hafner93,prakken2002exercise,bench-capon_sartor03}
  understanding it in a teleological sense: a decision promoting a
  value principle means taking that decision \textit{for the sake} of
  observing the principle; thus seeing the value principle \textit{as a reason}
  for taking that decision.}
\begin{align*}
  \textit{Promotes}(F,D,P) :=~~F \rightarrow \Box^{\prec}(D \leftrightarrow \Diamond^{\prec}P{\downarrow})
\end{align*}
It is important to remark that, in the spirit of the \logikey\ methodology, the definition above has arisen from
the many iterations of encoding, testing and `debugging' of the
modelling of the `wild animal cases' in \S\ref{sec:Application} (until reaching a \textit{reflective equilibrium}). We can still try to give this definition a somewhat intuitive interpretation, which might read along the lines of: 
``given the facts F, taking decision D is (necessarily) tantamount to (possibly) observing principle P'', with the caveat that the (bracketed) modal expressions would need to be read in a non-alethic mood (e.g. deontically as discussed in \S\ref{sec:pl}).

\subsubsection*{Value Conflict}

Another important idea inspired from Lomfeld's theory in
\S\ref{sec:ValueTheory} is the notion of value \textit{conflict}. As
discussed there (see Fig.~\ref{fig:ValueOntology}), values are
disposed around two axis of value coordinates, with values lying at
contrary poles playing antagonistic roles. For our modelling purposes
it makes thus sense to consider a predicate \textit{Conflict} on
worlds (i.e., a proposition) signalling situations where value
conflicts appear. Taking inspiration from the traditional logical principle of
\textit{ex contradictio sequitur quodlibet}, which we may intuitively paraphrase, for the present purposes, as \textit{ex conflictio sequitur quodlibet},\footnote{We shall not be held responsible for damages resulting from sloppy Latin paraphrasings!} we define \textit{Conflict} as the set of those worlds in which \textit{all} basic values become applicable:

\begin{align*}
  \textit{Conflict} := \bigwedge \{v\}{\downarrow} \text{~~~~for all $v$  in $\V$}
\end{align*}

Of course, and in the spirit of the \logikey\ methodology, the specification of such a predicate can be further improved upon by the
modeller as the need arises.

\subsection{Instantiation as a HOL-based Legal DSL}
\label{subsec:EncodingValueOntology} 
  
In this subsection we encode our logic of value preferences in HOL (recall discussion in \S\ref{sec:hol}), building incrementally on top of the corresponding HOL-encoding for our (extended) object logic \logic\ in \S\ref{sec:Implementation}. In the process, our encoding will be gradually extended with custom means to encode Lomfeld's legal theory (cf.~\S\ref{sec:ValueTheory}).
For the sake of illustrating a concrete, formally-verifiable modelling we also present in most cases the corresponding encoding in {\Isabelle} (see also Appx.~\ref{app:SecValueOntology}).

In a preliminary step, we introduce a new base HOL-type \textit{c} (for ``contender'') as an (extensible) two-valued type introducing the legal parties ``plaintiff'' (\texttt{p}) and ``defendant'' (\texttt{d}). For this we employ in {\Isabelle} the keyword \texttt{datatype}, which has the advantage of automatically generating (under the hood) the adequate axiomatic constraints (i.e., the elements \texttt{p} and \texttt{d} are distinct and exhaustive).  

We also introduce a function, suggestively termed $\texttt{other}_{c \ar c}$, with notation $(\cdot)^{-1}$. This function is used to return for a given party the
\textit{other} one; i.e., $\texttt{p}^{-1} = \texttt{d}$ and
$\texttt{d}^{-1} = \texttt{p}$. Moreover, we add a ($\sigma$-lifted) predicate $\texttt{For}_{c \ar \sigma}$ to model the ruling \textit{for} a given party and postulate that it always has to be ruled for either one party or the other:
$\texttt{For}\ x \boldsymbol{\leftrightarrow} \boldsymbol{\neg}
\texttt{For}\ x^{-1}$.

\vskip1em
\noindent\colorbox{gray!30}{\includegraphics[width=.97\textwidth]{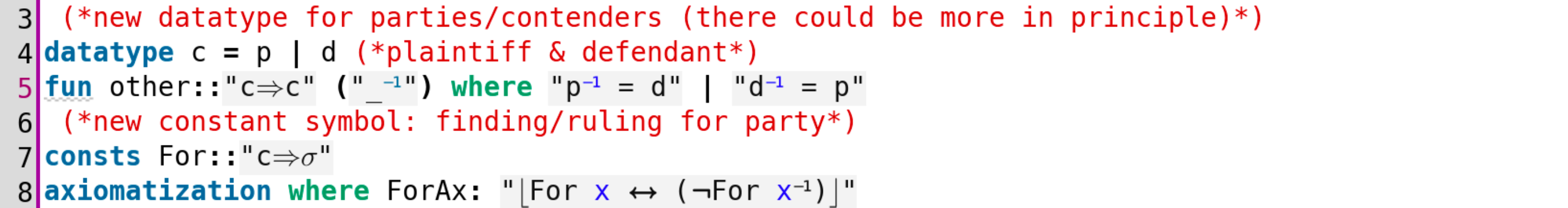}}
\vskip1em

As a next step, in order to enable the encoding of basic values, we introduce a four-valued datatype
$(\typevar)\,\texttt{VAL}$ (corresponding to our domain $\V$ of all values). Observe that
this datatype is parameterised with a type variable $\typevar$. In the
remainder we will always instantiate $\typevar$ with the type $c$ (see discussion below).
$$(\typevar)\,\texttt{VAL} :=~ \texttt{FREEDOM}\ \typevar \mid \texttt{UTILITY}\ \typevar \mid \texttt{SECURITY}\ \typevar \mid \texttt{EQUALITY}\ \typevar$$

We also introduce some convenient type-aliases:

${v~\text{:=}~(c)\,\texttt{VAL}\,{\ar}\,o}$ is introduced as the type for (characteristic functions of) sets of basic values. The reader will recall that this corresponds to the characterisation of value principles as given in the previous subsection (i.e., elements of $2^\V$). 

It is important to note, however, that to enable the modelling of legal cases (plaintiff v.~defendant) we need to further specify \textit{legal} value principles \textit{with respect to a legal party}, either plaintiff or defendant. For this we define ${cv~\text{:=}~c\,{\ar}\,v}$ intended as the type for specific legal (value) principles (wrt.~a legal party), so that they are functions taking objects of type \textit{c} (either \texttt{p} or \texttt{d}) to sets of basic values.

\noindent\colorbox{gray!30}{\includegraphics[width=.97\textwidth]{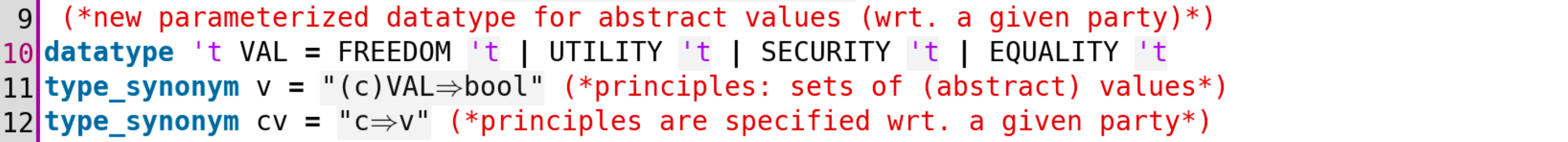}}
\vskip1em

We introduce useful set-constructor
operators for basic values ($\lBrace \dots\rBrace$) and a superscript notation for specification wrt.~a legal party. As an illustration, recalling the discussion in
\S\ref{sec:ValueTheory}, we have that, e.g., the legal principle of
STABility' wrt.~the plaintiff (notation $\texttt{STAB}^\texttt{\,p}$) can be encoded as a two-element set of basic values (wrt.~the plaintiff), i.e.,
$\lBrace \texttt{SECURITY}\ \texttt{p}$, $\texttt{UTILITY}\ \texttt{p}\rBrace$.

The corresponding {\Isabelle} encoding is:

\vskip1em
\noindent\colorbox{gray!30}{\includegraphics[width=.97\textwidth]{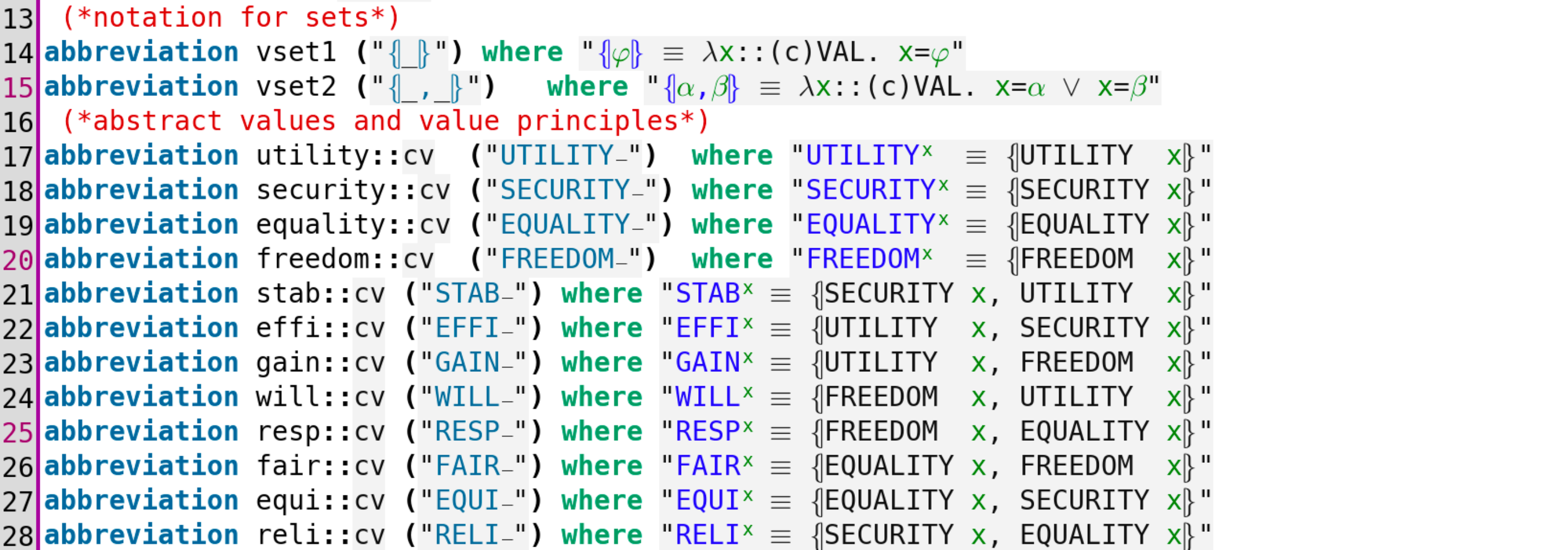}}
\vskip1em

After defining legal (value) principles as combinations (in this case:
sets\footnote{Recall our discussion in \S\ref{sec:ValueOntology} (cf.~footnote~\ref{footnote-modelling}).
In a future modelling of a (suitably enhanced) \textit{discoursive
    grammar} (\S\ref{sec:ValueTheory}) we might take into account the order of combination of basic values in forming value principles, to the effect
  that, e.g., STABility can be properly distinguished
  from EFFIciency.})  of basic values (wrt.~a
legal party), we need to relate them to propositions (sets of
worlds/states) in our logic \logic. For this we employ the
\textit{derivation operators} introduced in \S\ref{sec:ValueOntology},
whereby each value principle (set of basic values) becomes
associated with a proposition (set of worlds) by means of the operator
$\downarrow$ (conversely for ${\uparrow}$).  We encode this by
defining the corresponding \textit{incidence} relation, or, equivalently, a function $\I_{\iota\,{\ar}\,v}$ mapping
worlds/states (type $\iota$) to sets of basic values (type
$v = (c)\,\texttt{VAL}\,{\ar}\,o$). We define ${\downarrow}_{v \ar \sigma}$ so that, given some set of basic values $V_{v}$, $V{\downarrow}_{\sigma}$
denotes the set of all worlds that are $\I$-related to every
value in $V$ (analogously for ${\uparrow}_{\sigma \ar v}$).  The modelling in the
{\Isabelle} proof assistant is as follows:

\vskip1em
\noindent\colorbox{gray!30}{\includegraphics[width=.97\textwidth]{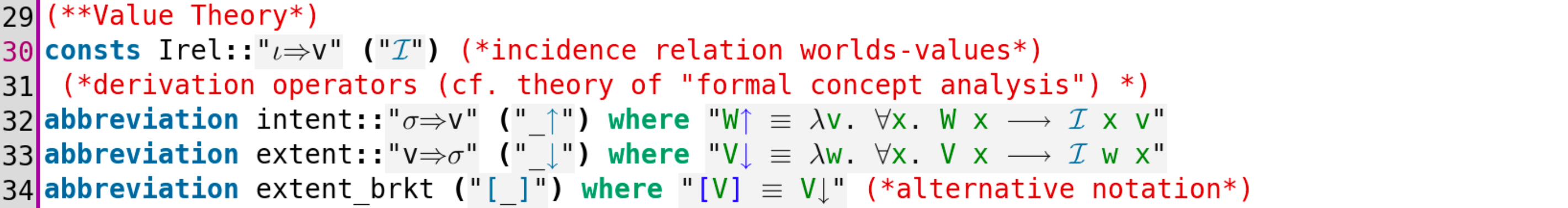}}
\vskip1em

Thus we can intuitively read the proposition (set of worlds) denoted by
${\text{STAB}^p{\downarrow}}$ as (those worlds in which) ``the legal principle of STABility is observed wrt.~the plaintiff''. For convenience, we introduce square brackets ($[\cdot]$) as an alternative notation to $\downarrow$-postfixing in our DSL, so we have $[V] = V{\downarrow}$.

Now, our concrete choice of an aggregation operator for values (out of the two options presented in \S\ref{subsec:LogicValuePreferences}) is $\oplus_{(2)}$, which thus becomes encodes in HOL as:
\begin{align*}
    A_v \oplus_{v \ar v \ar \sigma} B_v :=~(A{\downarrow})_\sigma \vee (B{\downarrow})_\sigma
  \end{align*}
  
Analogously, the chosen preference relation ($\prec$) is the variant $\prec_{AE}$ (i.e.~$\prec_{(2)}$ from the candidate modelling options discussed in \S\ref{sec:ValueOntology}), which, recalling \S\ref{sec:pl}, becomes equivalently encoded as any of the following:
  \begin{align*}
    \varphi_\sigma \prec_{\sigma \ar \sigma \ar \sigma} \psi_\sigma :=&~\forall s_i\ \varphi\,s \rightarrow (\exists t_i\ \psi\,t \wedge s\ {\prec}\ t)\\
    \varphi_\sigma \prec_{\sigma \ar \sigma \ar \sigma} \psi_\sigma :=&~\boldsymbol{A}_{\sigma \ar \sigma}(\varphi \boldsymbol{\rightarrow} \Diamond^\prec_{\sigma \ar \sigma}\psi)
  \end{align*}

In a similar fashion, we encode in HOL the value-logical predicate \textit{Promotes} as introduced in the previous subsection \S\ref{subsec:LogicValuePreferences}. The corresponding {\Isabelle} encoding is shown below:

\vskip1em
\noindent\colorbox{gray!30}{\includegraphics[width=.97\textwidth]{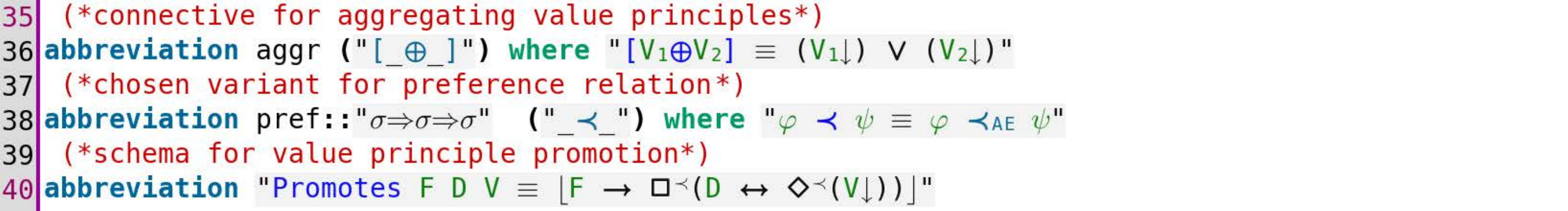}}
\vskip1em
We have similarly encoded the proposition \textit{Conflict} in HOL.
\vskip1em
\noindent\colorbox{gray!30}{\includegraphics[width=.97\textwidth]{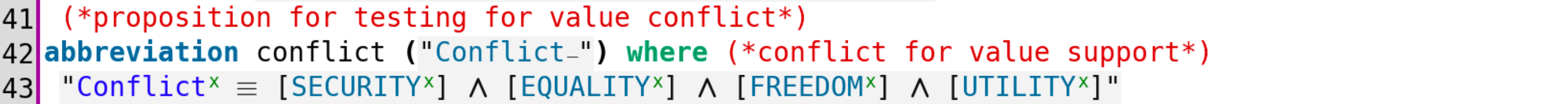}}
\vskip1em

\subsection{Formally Verifying DSL's Adequacy} 

\begin{figure}[!bp]
  \centering
  \colorbox{gray!30}{\includegraphics[width=.97\textwidth]{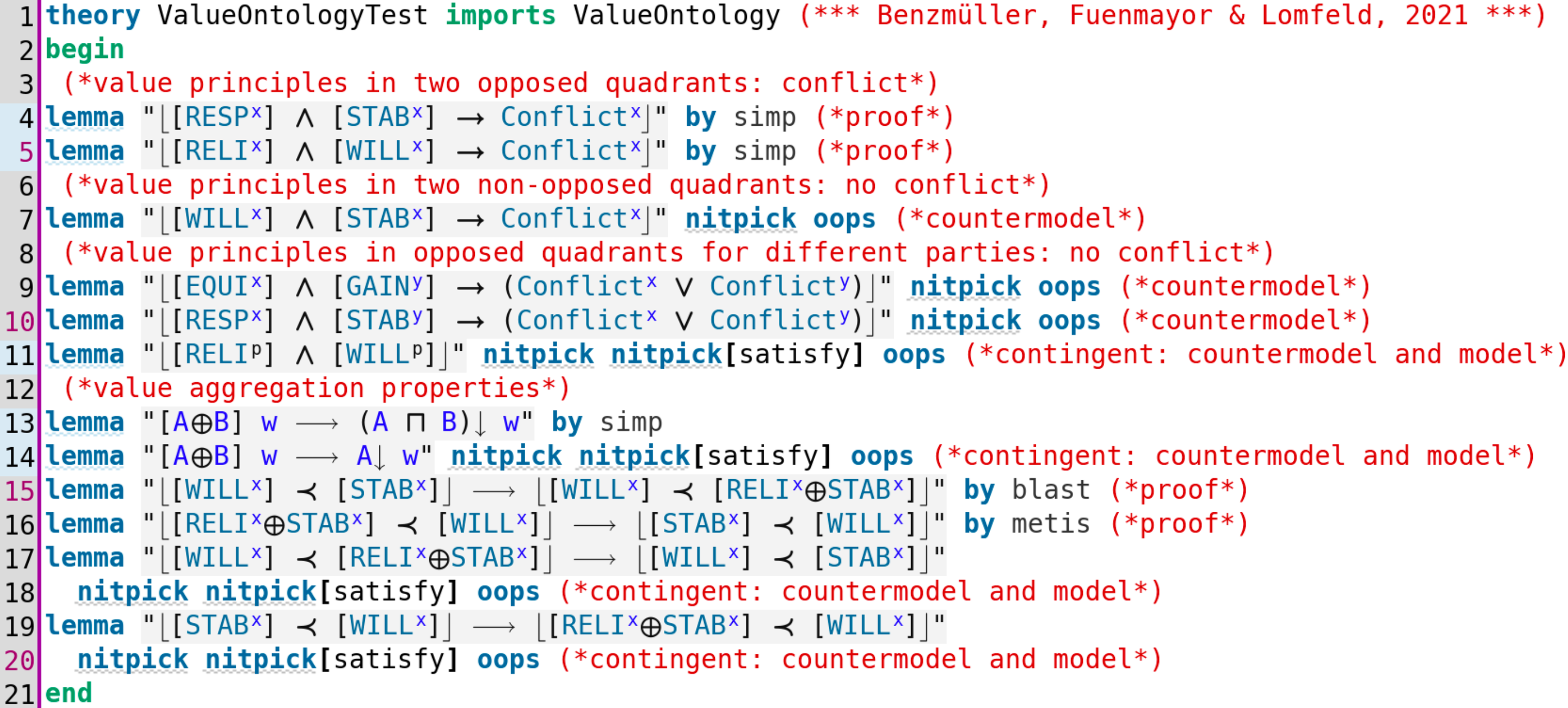}}
  \caption{Verifying the DSL}
  \label{fig:TestingValueOntology}
\end{figure}

In this subsection we put our HOL-based legal DSL to the test by employing the automated tools integrated into {\Isabelle}. In this process, the \textit{discoursive grammar}, as well as the continuous feedback by our legal domain expert (Lomfeld), served the role of a requirements specification for the formal verification of 
the adequacy of our modelling. 
We briefly discuss some of the
conducted tests as shown in Fig.~\ref{fig:TestingValueOntology};
further tests are presented in Fig.~\ref{fig:ValueOntologyTestLong} in
Appx.~\ref{app:SecValueOntology} and in \textcite{C91}.

\begin{figure}[!htb] \centering
  \colorbox{gray!30}{\includegraphics[width=.97\textwidth]{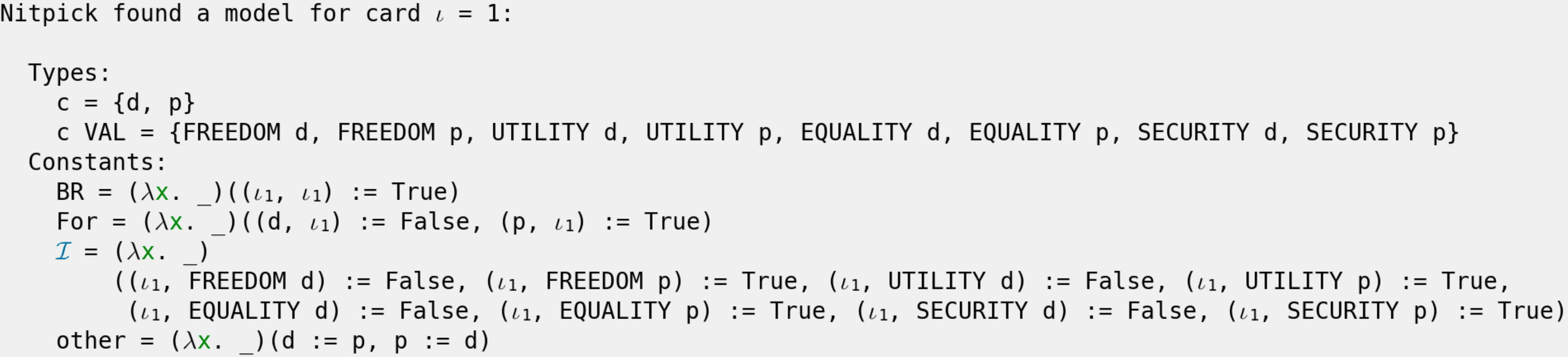}}
  \caption{Satisfying model for the statement in Line 11 of
    Fig.~\ref{fig:TestingValueOntology}.} \label{fig:Model1}
\end{figure}

In accordance with the dialectical interpretation of the
\textit{discoursive grammar} (recall Fig.~\ref{fig:ValueOntology} in
\S\ref{sec:ValueTheory}), our modelling foresees that observing values (wrt.~the same party) from two opposing value quadrants, say RESP \& STAB, or RELI \& WILL,
entails a value conflict; theorem provers quickly confirm this as
shown in Fig.~\ref{fig:TestingValueOntology} (Lines 4--5).  Moreover,
observing values from two non-opposed quadrants, such as WILL \& STAB
(Line 7) should not imply any conflict: the model finder
{\Nitpick}\footnote{{\Nitpick} \parencite{Nitpick} searches for,
  respectively enumerates, finite models or countermodels to a
  conjectured statement/lemma. By default {\Nitpick} searches for
  countermodels, and model finding is enforced by stating the
  parameter keyword `satisfy'. These models are given as concrete
  interpretations of relevant terms in the given context so that the
  conjectured statement is satisfied or falsified.}  computes and
reports a countermodel (not shown here) to the stated conjecture.  A
value conflict is also not implied if values from opposing quadrants
are observed wrt.~different parties (Lines 9--10).

Note that the notion of value conflict has deliberately not been
aligned with logical inconsistency, neither in the object logic \logic\ not in the meta-logic HOL.  This way we can
represent conflict situations in which, for instance, RELI and WILL
(being conflicting values, see~Line 5 in
Fig.~\ref{fig:TestingValueOntology}) are observed wrt.~the plaintiff
($p$), without leading to a logical inconsistency in \Isabelle\ (thus
avoiding `explosion').  In Line 11 of
Fig.~\ref{fig:TestingValueOntology}, for example, {\Nitpick} is called
simultaneously in both modes in order to confirm the contingency of
the statement; as expected both a model (cf.~Fig.~\ref{fig:Model1})
and countermodel (not displayed here) for the statement are
returned. This value conflict can also be spotted by
inspecting the satisfying models generated by {\Nitpick}. One of such
models is depicted in Fig.~\ref{fig:Model1}, where it is shown that
(in the given possible world $\iota_1$) all of the basic values
(EQUALITY, SECURITY, UTILITY, and FREEDOM) are simultaneously observed
wrt.~$p$, which implies a value conflict according to our definition.

Such model structures as computed by {\Nitpick} are ideally
\textit{communicated to} (and \textit{inspected with}) domain experts
(Lomfeld in our case) early on and checked for plausibility, which, in
case of issues, might trigger adaptions to the axioms and
definitions. Such a process may require several cycles until arriving at a state of \textit{reflective equilibrium} (recall the
discussion from \S\ref{sec:methodology}) and, as a useful side effect,
it conveniently fosters cross-disciplinary mutual understanding.

Further tests in Fig.~\ref{fig:TestingValueOntology} (Lines 13-20)
assess the behaviour of the aggregation operator $\oplus$ by itself, and also
in combination with value preferences.  For example, we test for a
correct behaviour when `strengthening' the right-hand side: if STAB is
preferred over WILL, then STAB combined with, say, RELI is also
preferred over WILL alone (Line 15). Similar tests are conducted for
`weakening' of the left-hand side.\footnote{Further related tests are
  reported in Fig.~\ref{fig:ValueOntologyTestLong} in
  Appx.~\ref{app:SecValueOntology}.}

  \section{Applications (L3) -- Assessment of Legal Cases} \label{sec:Application}
  
In this section we provide a concrete illustration of our reasoning framework by formally encoding and assessing two classic common law property cases concerning the appropriation of wild animals (``wild animal cases''): Pierson v.~Post, and Conti v.~ASPCA.\footnote{Cf.~\textcite{bench2002missing,prakken2002exercise,berman_hafner93}, and also \textcite{gordon2006pierson} for the significance of the Pierson v.~Post case as a benchmark.}

Before starting with the analysis a word is in order about the support of our work by the tools \textit{Sledgehammer} \parencite{Sledgehammer,blanchette2016hammering} and \textit{Nitpick} \parencite{Nitpick} in {\Isabelle}. The ATP systems integrated via \textit{Sledgehammer} in {\Isabelle} include higher-order ATP systems, first-order ATP systems, and SMT (satisfiability modulo theories) solvers, and many of these systems in turn use efficient SAT solver technology internally. Indeed, proof automation with \textit{Sledgehammer} and (counter)model finding with \textit{Nitpick} were invaluable in supporting our exploratory modeling approach at various levels. 
These tools were very responsive in automatically proving (\textit{Sledgehammer}), disproving (\textit{Nitpick}), or showing consistency by providing a model (\textit{Nitpick}). In the first case, references to the required axioms and lemmas were returned (which can be seen as a kind of abduction), and in the case of models and counter-models they often proved to be very readable and intuitive. In this section, we highlight some explicit use cases of \textit{Sledgehammer} and \textit{Nitpick}. They have been similarly applied at all levels as mentioned before.

We have split our analysis in layer L3 into two `sub-layers' in order to highlight the separation between general legal \& world knowledge (legal concepts and norms), from its `application' to relevant facts in the process of deciding a case (factual/contextual knowledge). We shall first address the modelling of some background legal and world knowledge in \S\ref{sec:LegalAndWorldKnowledge}, as minimally required in order to formulate each of our legal cases in the form of a logical {\Isabelle} theory (cf.~\S\ref{subsec:app:pierson}).

\subsection{General Legal \& World Knowledge}\label{sec:LegalAndWorldKnowledge}
The realistic modelling of concrete legal cases requires further legal
\& world knowledge (LWK) to be taken into account. LWK is typically
modelled in so called ``upper'' and ``domain'' ontologies. The
question about which particular notion belongs to which category is
difficult, and apparently there is no generally agreed answer in the
literature. Anyhow, we introduce only a small and monolithic examplary logical
{\Isabelle} theory,\footnote{Isabelle documents are suggestively called
  ``theories''. They correspond to top-level modules bundling together
  related definitions, theories, proofs, etc.} called ``GeneralKnowledge'', with a minimal amount of axioms and definitions as required to encode our legal cases.  This LWK example
includes a small excerpt of a much simplified ``animal appropriation
taxonomy'', where we associate ``animal appropriation'' (kinds of)
situations with the value preferences they imply (i.e., conditional preference relations as discussed in \S\ref{sec:ValueTheory} and \S\ref{sec:ValueOntology}).

In a more realistic setting this knowledge base would be further split and
structured similarly to other legal or general ontologies, e.g., in
the \textit{Semantic Web} \parencite{CasanovasPPEV14,DBLP:conf/icail/HoekstraBBB09}.  Note, however, that the expressiveness in
our approach, unlike in many other legal ontologies or taxonomies, is
by no means limited to definite underlying (but fixed) logical
language foundations. We could thus easily decide for a more realistic
modelling, e.g.~avoiding simplifying propositional abstractions. For
instance, the proposition ``appWildAnimal'', representing the
appropriation of one or more wild animals, can anytime be replaced by
a more complex formula (featuring, e.g., quantifiers, modalities, and
conditionals; see \S\ref{subsec:PLextensions}).

Next steps include interrelating notions introduced in our
\textit{{\Isabelle} theory} ``GeneralKnowledge'' with values and value
preferences, as introduced in the previous sections. It is here where
the preference relations and modal operators of \logic\ as well as the notions introduced in our legal DSL are most useful.
Remember that, at a later point and in line with the \logikey\
methodology, we may in fact exchange \logic\ by an alternative choice
of an object logic; or, on top of it, we may further modify our legal DSL, e.g., we might choose and assess alternative candidates for our connectives $\prec$ and $\oplus$; moreover, we may want to replace material implication $\boldsymbol{\rightarrow}$ by a conditional implication to better
support defeasible legal reasoning.\footnote{Remember that a
  defeasible conditional implication can be defined employing \logic\
  modal operators; cf.~\S\ref{subsec:PLextensions}. Alternatively we
  may also opt for an SSE of a conditional logic in HOL using other
  approaches as, e.g., in \textcite{C37}.}

We now briefly outline the {\Isabelle} encoding of our example LWK;
see Fig.~\ref{fig:GeneralKnowledge} in Appx.~\ref{subsec:LWK} for the
full details.

First, some non-logical constants that stand for kinds of legally
relevant situations (here: of appropriation) are introduced, and their
meaning is constrained by some postulates:

\vskip1em
\noindent\colorbox{gray!30}{\includegraphics[width=.97\textwidth]{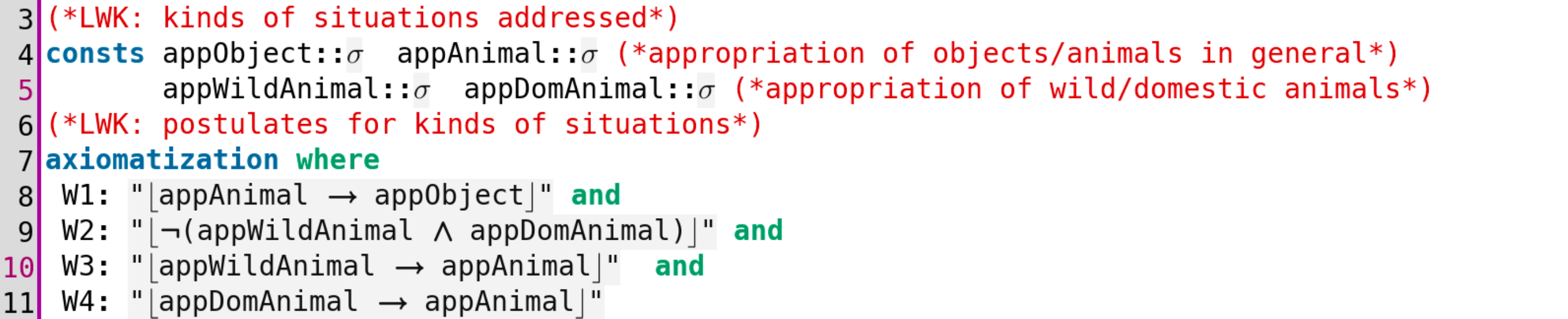}}
\vskip1em

Then the `default'\footnote{We use of the term `default' in the colloquial sense of `fallback', noting however, that there exist in fact several (non-monotonic) logical systems aimed at modelling such a kind of \textit{defeasible}, aka.~``default'', behaviour for rules/conditionals (i.e., meaning that they can be `overruled'). One of them has been suggestively called ``default logic''. We refer to \textcite{sep-reasoning-defeasible} for a discussion.
In fact, and in the spirit of \logikey, we could have also employed, for encoding these rules, a \logic-defined defeasible conditional as discussed in \S\ref{subsec:PLextensions}. For the illustrative purposes of the present paper, and in view of the good performance of our present modelling, we did not yet find this step necessary.} legal rules for several situations (here:
appropriation of animals) are formulated as conditional preference
relations:

\vskip1em
\noindent\colorbox{gray!30}{\includegraphics[width=.97\textwidth]{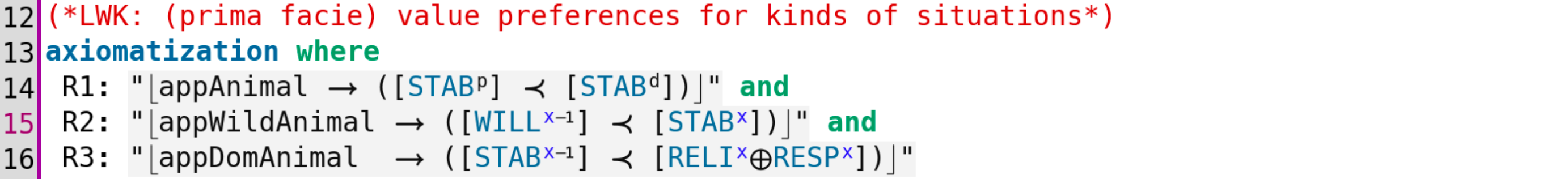}}
\vskip1em

For example, rule R2 could be read as: ``In a
wild-animals-appropriation kind of situation, observing
STABility wrt.~a party (say, the plaintiff) is
preferred over observing WILL wrt.~the other party
(defendant)''.  If there is no more specific legal rule from a
precedent or a codified statute then these `default' preference
relations determine the result. Of course, this default is not arbitrary but itself an implicit normative setting of the existing legal statutes or cases. Moreover, we can have rules
conditioned on more concrete legal \textit{factors}.\footnote{The
  introduction of legal \textit{factors} is an established
  practice in the implementation of case-based legal systems
  (cf.~\textcite{bench2017hypo} for an overview). They can be conceived
  --as we do-- as propositions abstracted from the facts of a case by
  the analyst/modeler in order to allow for assessing and comparing
  cases at a higher level of abstraction. Factors are typically either
  pro-plaintiff or pro-defendant, and their being true or false
  (resp.~present or absent) in a concrete case can serve to invoke
  relevant precedents or statutes.}
  As a didactic example, the legal
rule R4 states that the \textit{own}ership (say, the plaintiff's) of
the land on which the appropriation took place, together with the fact
that the opposing party (defendant) acted out of \textit{mal}ice
implies a value preference of \textit{reli}ance and
\textit{resp}onsibility over \textit{stab}ility. This rule has been
chosen to reflect the famous common law precedent of Keeble
v.~Hickeringill (1704, 103 ER 1127; cf.~also \textcite{berman_hafner93,bench2002missing}).

\vskip1em
\noindent\colorbox{gray!30}{\includegraphics[width=.97\textwidth]{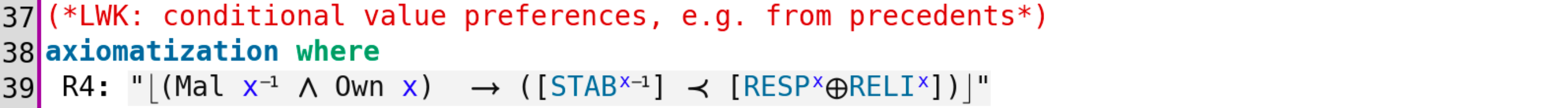}}
\vskip1em

As already discussed, for ease of illustration, terms like
``appWildAnimal'' are modelled here as simple propositional
constants. In practice, however, they may later be replaced, or
logically implied, by a more realistic modelling of the relevant
situational facts, utilising suitably complex (even quantified; cf.~\S\ref{subsec:PLextensions}) formulas depicting states of affairs to some desired level of granularity.

For the sake of modelling the appropriation of objects, we have
introduced an additional base type in our meta-logic HOL (recall
\S\ref{sec:hol}). The type $e$ (for `entities') can be employed for
terms denoting individuals (things, animals, etc.) when modelling
legally relevant situations.  Some simple vocabulary and taxonomic
relationships (here: for wild and domestic animals) are specified to
illustrate this.

\vskip1em
\noindent\colorbox{gray!30}{\includegraphics[width=.97\textwidth]{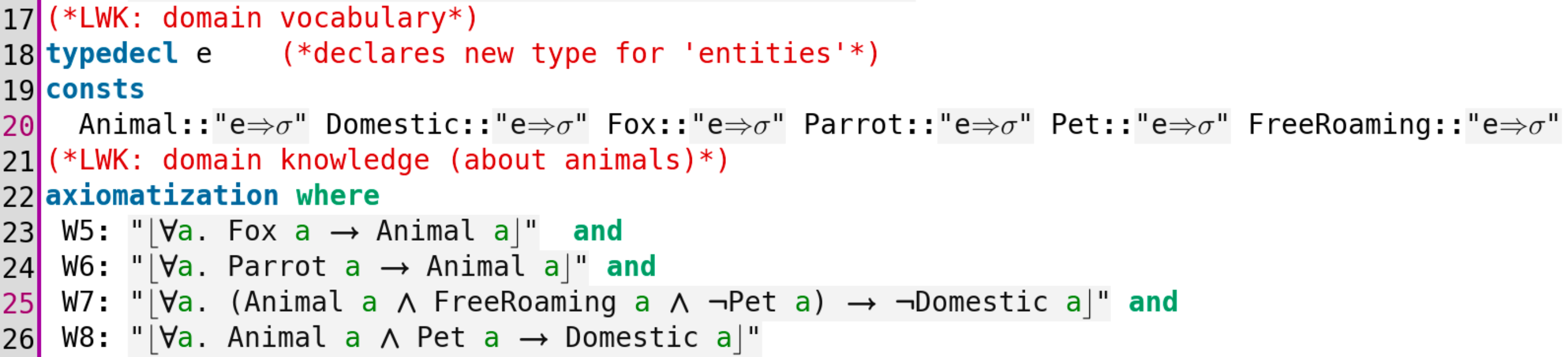}}
\vskip1em

As mentioned before, we have introduced some convenient legal
\textit{factors} into our example LWK to allow for the encoding of
legal knowledge originating from precedents or statutes at a more
abstract level. In our approach these factors are to be logically
implied (as deductive arguments) from the concrete facts of the case
(as exemplified in \S\ref{subsec:Pierson} below). Observe that our framework also
allows us to introduce definitions for those factors for which clear
legal specifications exist, such as property or possession.  At the
present stage, we will provide some simple postulates constraining
factors' interpretation.

\vskip1em
\noindent\colorbox{gray!30}{\includegraphics[width=.97\textwidth]{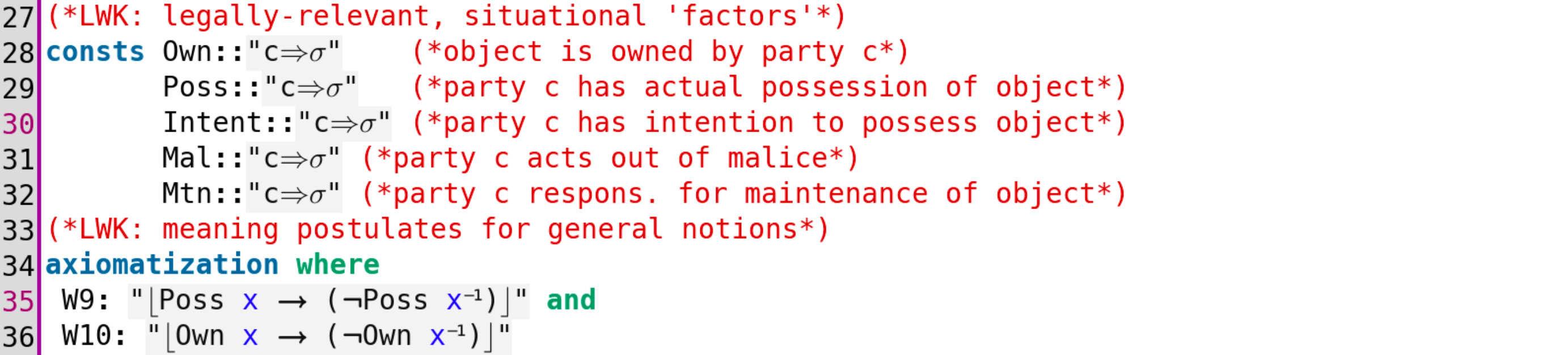}}
\vskip1em

Recalling \S\ref{sec:ValueOntology} we relate the introduced legal factors (and relevant situational facts)
to value principles and outcomes by means of the \textit{Promotes}
predicate:\footnote{We note that our normative assignment here is
  widely in accordance with classifications in the AI \& Law
  literature \parencite{berman_hafner93,bench-capon12}.}

\vskip1em
\noindent\colorbox{gray!30}{\includegraphics[width=.97\textwidth]{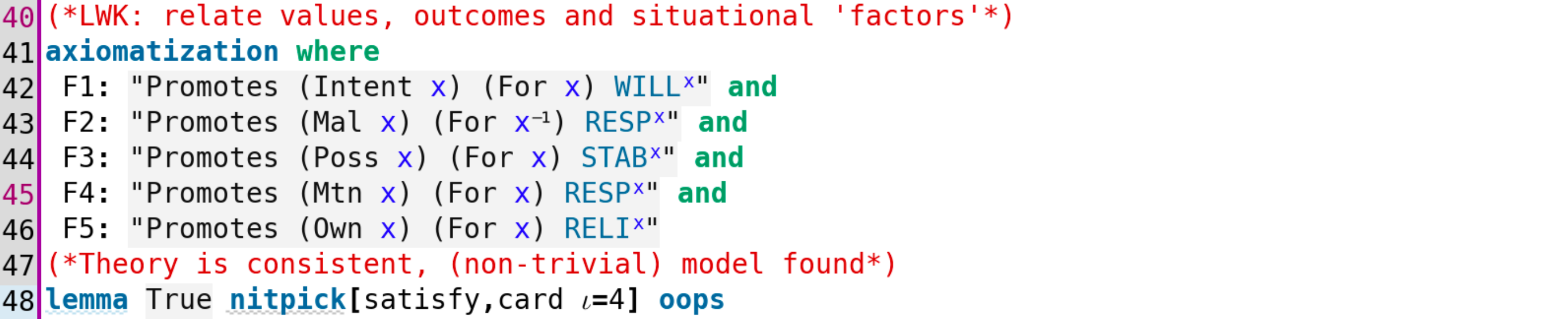}}
\vskip1em

Finally, the consistency of all axioms and rules provided is confirmed
by {\Nitpick}.

\subsection{Pierson v.~Post} \label{subsec:app:pierson}

This famous legal case \parencite{gordon2006pierson} can be succinctly described as follows:

\textit{Pierson killed and carried off a fox which Post already was hunting with
  hounds on public land. The Court found for Pierson} (1805, 3 Cai R 175).

For the sake of illustration we will consider in this subsection two modelling scenarios: in the first one a case is built to favour the defendant (Pierson); in the second one a case favouring the plaintiff (Post).

\subsubsection*{Ruling for Pierson} 

The formal modelling of an argument in favour of Pierson is outlined
next.\footnote{The entire formalisation of this argument is presented in
  Fig.~\ref{fig:Pierson} in Appx.~\ref{subsec:Pierson}.}

First we introduce some minimal vocabulary: a constant $\alpha$ of
type $e$ (denoting the appropriated animal), and the relations
\textit{pursue} and \textit{capture} between the animal and one of the
parties (of type $c$). A background (generic) theory as well as the
(contingent) case facts as suitably interpreted by Pierson's party are
then stipulated:

\vskip1em
\noindent\colorbox{gray!30}{\includegraphics[width=.97\textwidth]{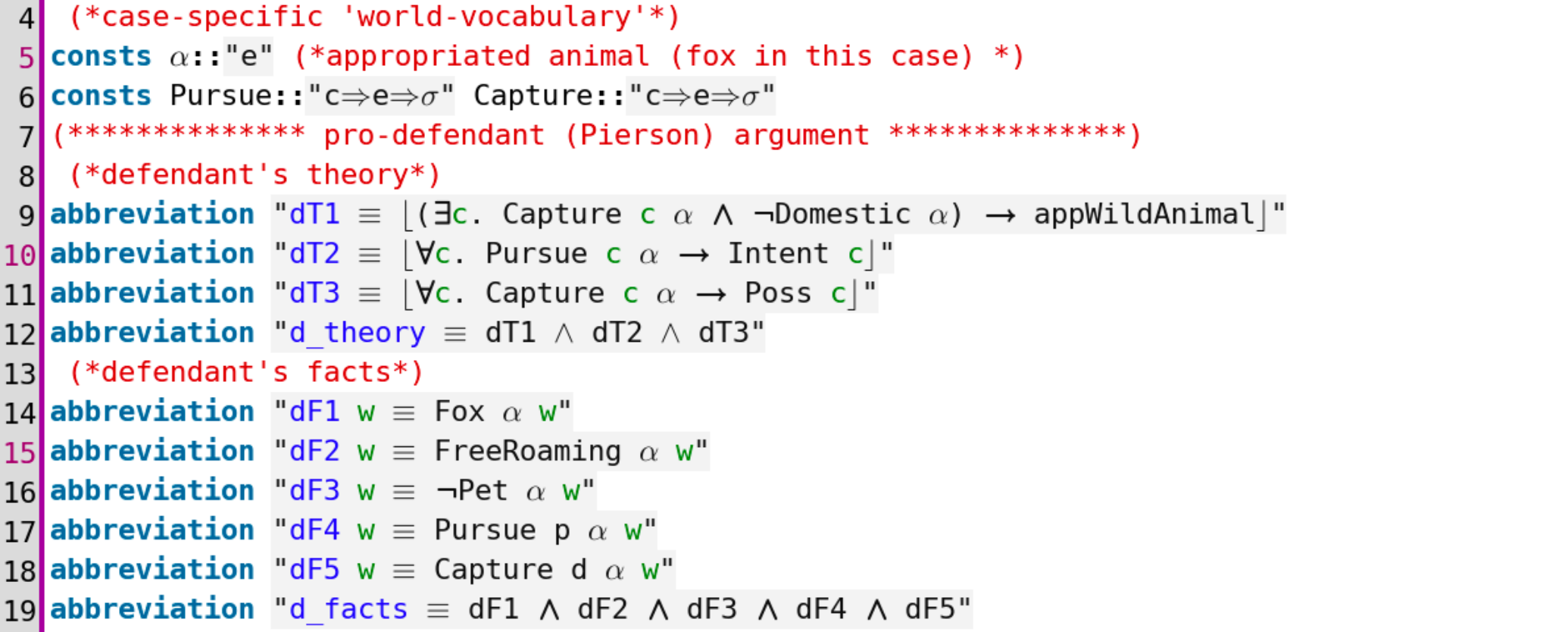}}
\vskip1em

The aforementioned decision of the court for Pierson was justified by
the majority opinion. The essential preference relation in the case is
implied in the idea that appropriation of (free-roaming) wild animals
requires actual corporal possession. The manifest corporal link to the
possessor creates legal certainty, which is represented by the value
STABility and outweighs the mere WILL to
possess by the plaintiff; cf. the arguments of classic lawyers
cited by the majority opinion: ``pursuit alone vests no property''
(Justinian), and ``corporal possession creates legal certainty''
(Pufendorf). According to Lomfeld's legal theory in
\S\ref{sec:ValueTheory} (cf.~Fig.~\ref{fig:ValueOntology}), this corresponds to a preference for the basic value SECURITY over FREEDOM. We can see that this legal rule
{R2}, as introduced in the previous section
(\S\ref{sec:LegalAndWorldKnowledge})\footnote{Also observe that the
  legal precedent rule R4 of Keeble v.~Hickeringill (see
    Fig.~\ref{fig:GeneralKnowledge}, Line 39) as appears in \S\ref{sec:LegalAndWorldKnowledge} does not
  apply to this case.  }
is indeed employed by {\Isabelle}'s automated tools to prove that,
given a suitable defendant's theory, the (contingent) facts imply a
decision in favour of Pierson in all `better' worlds (which we could even
give a `deontic' reading as some sort of \textit{recommendation}):

\vskip1em
\noindent\colorbox{gray!30}{\includegraphics[width=.97\textwidth]{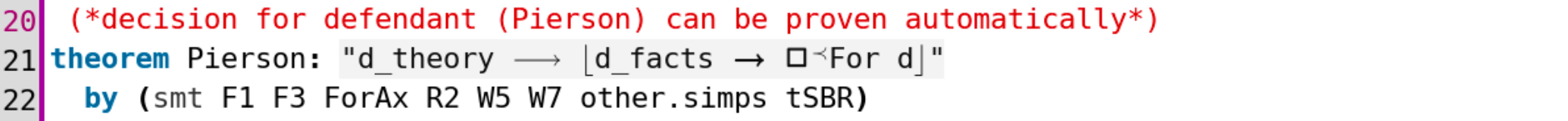}}
\vskip1em

The previous `one-liner' proof has indeed been automatically suggested by \textit{Sledgehammer} \parencite{Sledgehammer,blanchette2016hammering} which we credit, together with the model finder \textit{Nitpick} \parencite{Nitpick}, for doing the proof heavy-lifting in our work. 

A proof argument in favour of Pierson that uses the same dependencies
can also be constructed interactively using \textit{Isabelle}'s
human-readable proof language \textit{Isar} (\textit{Isabelle/Isar}; cf.~\textcite{wenzel2007isabelle}).  The individual steps of the proof are
this time formulated with respect to an explicit world/situation
parameter $w$. The argument goes roughly as follows:
\begin{enumerate}
\item \label{step1} From Pierson's facts and theory we infer that in
  the disputed situation $w$ a wild animal has been appropriated:
  $\text{appWildAnimal}\ w$
\item \label{step2} In this context, by applying the value preference
  rule R2, we get that observing STAB wrt.~Pierson (\texttt{d}) is
  preferred over observing WILL wrt.~Post (\texttt{p}):
  $\lfloor [\text{WILL}^p] \prec [\text{STAB}^d] \rfloor$
\item \label{step3} The possibility of observing WILL wrt.~Post thus entails the possibility of observing STAB wrt.~Pierson:
  $\lfloor \Diamond^\prec[\text{WILL}^p] \boldsymbol\rightarrow
  \Diamond^\prec[\text{STAB}^d] \rfloor$
\item \label{step4} Moreover, after instantiating the \textit{value
    promotion} schema F1 (\S\ref{sec:LegalAndWorldKnowledge}) for Post
  ($p$), and acknowledging that his pursuing the animal (Pursue $p$
  $\alpha$) entails his intention to possess (Intent $p$), we obtain
  (for the given situation $w$) a recommendation to `align' any
  ruling for Post with the possibility of observing WILL wrt.~Post:
  ${\Box^\prec(\text{For}\ p \boldsymbol\leftrightarrow
    \Diamond^\prec[\text{WILL}^p])\ w}$. Following the interpretation
  of the \textit{Promotes} predicate given in
  \S\ref{sec:ValueOntology}, we can read this `alignment' as involving
  both a logical entailment (left to right) and a justification (right to left); thus the
  possibility of observing WILL (wrt.~Post) both entails and
  justifies (as a reason) a legal decision for Post.

\item \label{step5} Analogously, in view of Pierson's ($d$) capture of
  the animal (Capture $d$ $\alpha$), thus having taken possession of
  it (Poss $d$), we infer from the instantiation of \textit{value
    promotion} schema F3 (for Pierson) a recommendation to align a
  ruling for Pierson with the possibility of observing the value
  principle STAB wrt.~Pierson):
  $\Box^\prec(\text{For}\ d \boldsymbol\leftrightarrow
  \Diamond^\prec[\text{STAB}^d])\ w$
\item \label{step6} From (\ref{step4}) and (\ref{step5}) in
  combination with the courts duty to find a ruling for one of both
  parties (axiom \textit{ForAx}) we infer, for the given situation $w$, that either
  the possibility of observing WILL wrt.~Post or the
  possibility of observing STAB wrt.~Pierson (or both) hold
  in every `better' world/situation (thus becoming a recommended
  condition):
  $\Box^\prec( \Diamond^\prec[\text{WILL}^p] \boldsymbol\vee
  \Diamond^\prec[\text{STAB}^d])\ w$
\item From this and (\ref{step3}) we thus get that the possibility
  of observing STAB wrt.~Pierson is recommended in the given
  situation $w$: $ \Box^\prec(\Diamond^\prec[\text{STAB}^d])\ w$
\item And this together with (\ref{step5}) finally implies the
  recommendation to rule in favour of Pierson in the given situation
  $w$: $\Box^\prec(\text{For}\ d\ v)$
\end{enumerate}

\vskip1em
\noindent\colorbox{gray!30}{\includegraphics[width=.97\textwidth]{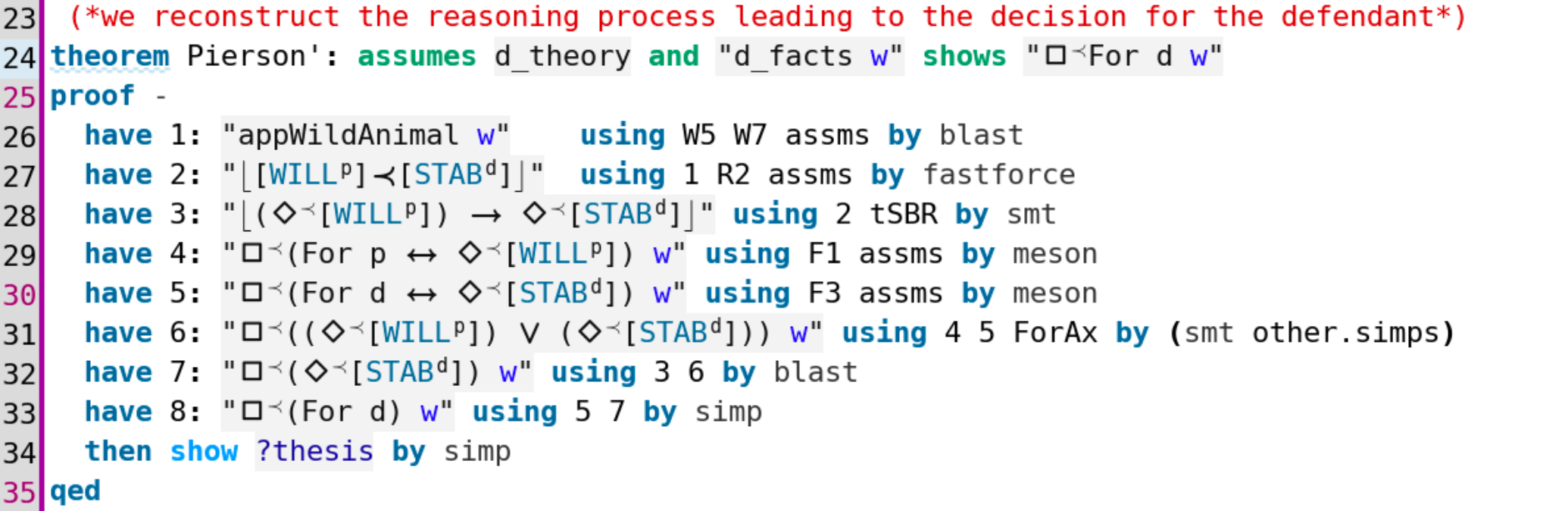}}
\vskip1em

The consistency of the assumed theory and facts (favouring Pierson)
together with the other postulates from the previously introduced
logical theories ``GeneralKnowledge'' and ``ValueOntology'' is verified by
generating a (non-trivial) model using {\Nitpick} (Line 38).  Further
tests confirm that the decision for Pierson (and analogously for Post)
is compatible with the premises and, moreover, that for neither party
value conflicts are implied.

\vskip1em
\noindent\colorbox{gray!30}{\includegraphics[width=.97\textwidth]{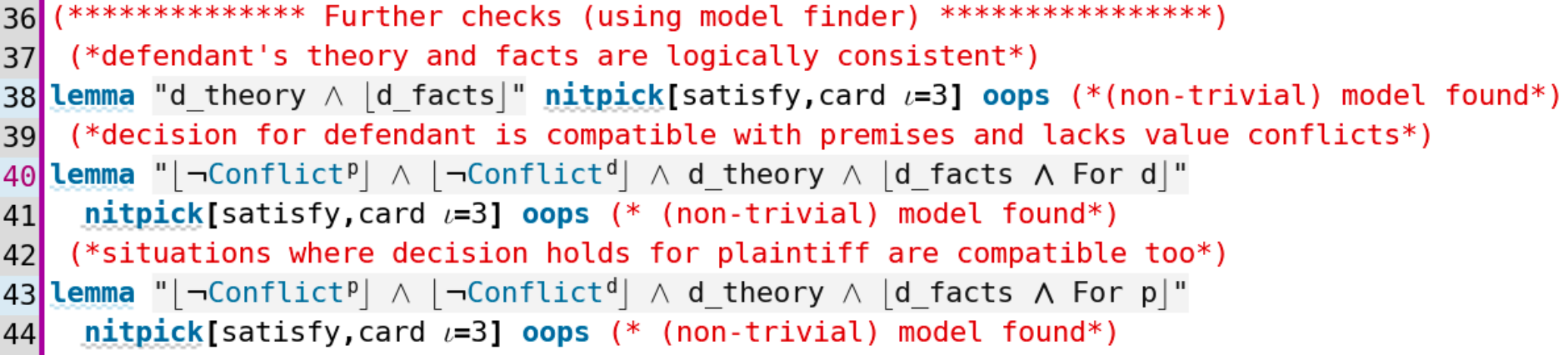}}
\vskip1em

We show next, how it is indeed possible to
construct a case (theory) suiting Post using our approach.

\subsubsection*{Ruling for Post}

We model a possible counterargument in favour of Post claiming an
interpretation (i.e., a \textit{distinction} in case law methodology) in that
the animal, being vigorously pursued (with large dogs and hounds) by a
professional hunter, is not ``free-roaming''.  In doing this, the
value preference
${\lfloor [\text{WILL}^p] \prec [\text{STAB}^d] \rfloor}$ (for
appropriation of wild animals), as in the previous Pierson's argument,
does not obtain.  Furthermore, Post's party postulates an alternative
(suitable) value preference for hunting situations.

\vskip1em
\noindent\colorbox{gray!30}{\includegraphics[width=.97\textwidth]{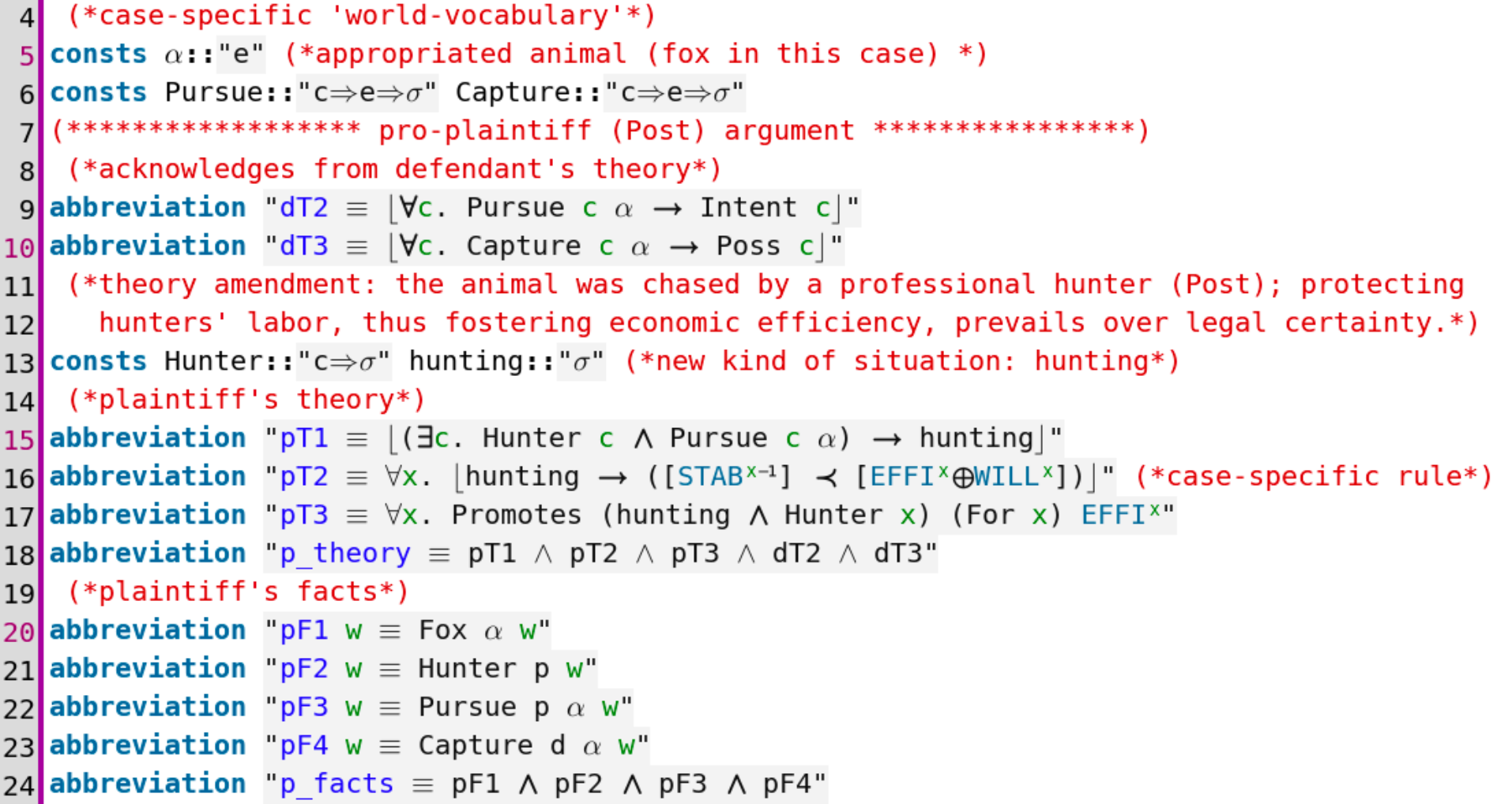}}
\vskip1em

Note that an alternative legal rule (i.e., a possible argument for
  overruling in case law methodology) is presented in Line 16 above,
entailing a value preference of the value principle combination
EFFIciency together with WILL over STABility:
${\lfloor [\text{STAB}^d] \prec [\text{EFFI}^p \oplus \text{WILL}^p]
  \rfloor}$.  Following the argument put forward by the dissenting
opinion in the original case (3 Cai R 175) we might justify this new
rule (inverting the initial value preference in the presence of EFFI)
by pointing to the alleged public benefit of hunters getting rid of
foxes, since the latter cause depredations in farms.

Accepting these modified assumptions the deductive validity of a
decision for Post can in fact be proved and confirmed automatically, again, thanks to \textit{Sledgehammer}:
\vskip1em
\noindent\colorbox{gray!30}{\includegraphics[width=.97\textwidth]{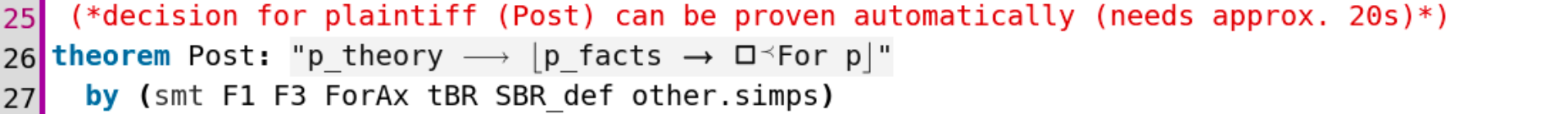}}
\vskip1em

Similar to above, a detailed, interactive proof for the argument in
favour of Post has been encoded and verified in \textit{Isabelle/Isar}. We have
also conducted further tests confirming the consistency of the
assumptions and the absence of value conflicts.\footnote{See the complete
  modelling in Fig.~\ref{fig:Post} in Appx.~\ref{subsec:Pierson}.}

\subsection{Conti v.~ASPCA}

An additional illustrative case study we have modelled in our
framework is Conti v. ASPCA (353 NYS 2d 288; cf.~\textcite{bench-capon_ea05}). In a nutshell:

\textit{Chester, a parrot owned by the ASPCA, escaped and was
  recaptured by Conti. The ASPCA found this out and reclaimed Chester
  from Conti. The court found for ASPCA.}

In this case, the court made clear that for domestic animals the
opposite preference relation as the standard in Pierson's case
applies.
More specifically, it was ruled that for a domestic animal it is in
fact sufficient that the owner did not neglect or stopped caring for
the animal, i.e., give up the responsibility for its maintenance
(RESP). This, together with ASPCA's reliance (RELI) in the parrot's
property, outweighs Conti's corporal possession (STAB) of the animal:
${\lfloor [\text{STAB}^d] \prec [\text{RELI}^p \oplus \text{RESP}^p]
  \rfloor}$.  Observe that a corresponding rule had previously been
integrated as R3 into our legal \& world knowledge
(\S\ref{sec:LegalAndWorldKnowledge}).

The plaintiff's theory and facts are encoded analogously to the
previous case:

\vskip1em
\noindent\colorbox{gray!30}{\includegraphics[width=.97\textwidth]{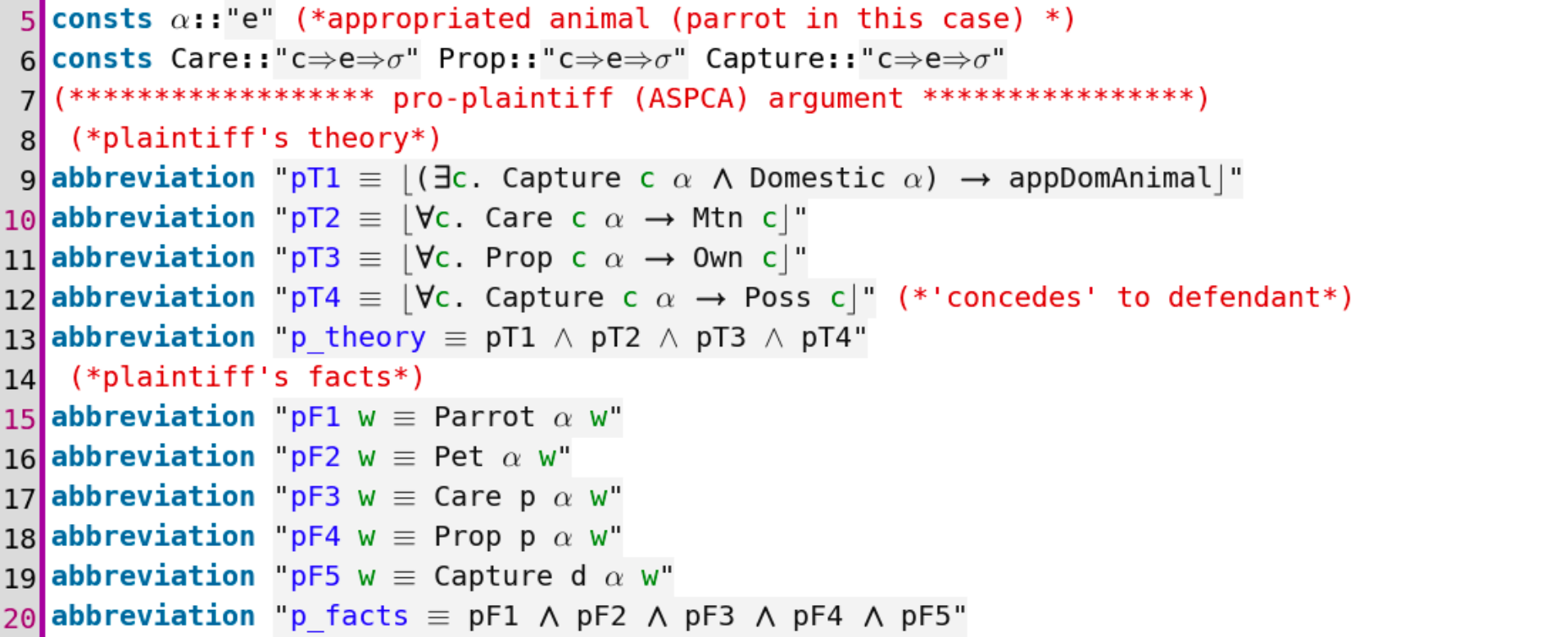}}
\vskip1em

Accepting these assumptions the deductive validity of a decision for
the plaintiff (ASPCA) can again be proved and confirmed automatically (thanks to \textit{Sledgehammer}):

\vskip1em
\noindent\colorbox{gray!30}{\includegraphics[width=.97\textwidth]{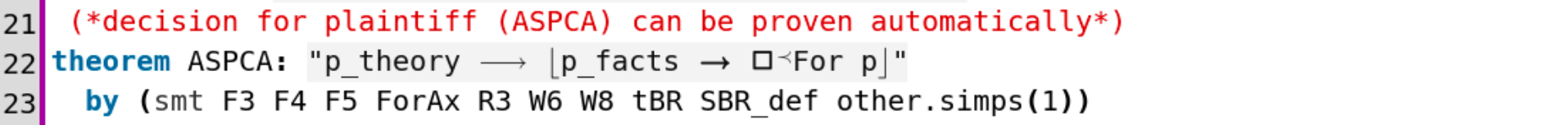}}
\vskip1em

In an analogous manner to Pierson's case, an interactive proof in
\textit{Isabelle/Isar} has been encoded and verified, and the consistency of
the assumptions and the absence of value conflicts has been
confirmed.\footnote{The full details of the encoding are presented in
  Fig.~\ref{fig:Conti} in Appx.~\ref{sec:Conti}.} 

\section{Related and Further Work} \label{sec:relwork}

Custom software systems for legal case-based reasoning have been
developed in the AI \& Law community, beginning with the influential
HYPO system in the 1980s \parencite{rissland1987case}; cf.~also the
survey paper by \textcite{bench2017hypo}. In later years, there has been a
gradual shift of interest from rule-based non-monotonic reasoning
(e.g., logic programming) towards argumentation-based approaches
(see~\textcite{prakken_sartor15} for an overview); however, we are not
aware of any other work that uses higher-order theorem proving and
proof assistants (the argumentation logic of \textcite{Krause1995} is an
early related effort that is worth mentioning). Another important
aspect of our work concerns value-oriented legal reasoning and
deliberation, where a considerable amount of work has been presented in AI \& Law
in response to the challenge posed by \textcite{berman_hafner93}. Our
approach, based mainly on Lomfeld's
(\citeyear{lomfeld19_grammatik,lomfeld15_vertrag}) theory, has also
been influenced by some of this work, in particular by
\textcite{prakken2002exercise,bench2002missing,
  bench-capon_sartor03}.

We are currently working towards further refining the modelling of Lomfeld's legal theory with the aim of providing more expressive (combinations of) object
logics at \logikey\ layer L1.  In this regard, it is somehow remarkable that the use of material implication to encode rules has proven
sufficient for the illustrative purposes of this paper. However, it is
important to note that a more realistic modelling of legal cases must
also provide mechanisms to deal with the inevitable emergence of
conflicts and contradictions in normative reasoning (overruling,
conflict resolution, etc.). In line with the {\logikey} approach, we
are working at introducing conditional connectives in our object logics with the aim of enabling \textit{defeasible} (or \textit{default}) reasoning. Such connectives can be introduced by reusing the
modal operators of \logic\ (recalling the discussion in \S\ref{subsec:PLextensions}) or, alternatively, through the shallow semantical embedding \parencite{J41} of a suitable conditional logic in HOL \parencite{J31}. 
Moreover, special kinds of paraconsistent (modal-like) \textit{Logics of Formal Inconsistency} \parencite{carnielli-coniglio-fuenmayor-2021} can also be integrated into our modelling to enable the
non-explosive representation of (and recovery from) contradictions by
purely object-logical means (cf.~\textcite{Topological_Semantics-AFP} for a related encoding in \Isabelle).
In a similar vein, we think that some of the recent work that employs expressive deontic logics for value-based legal balancing (e.g.~\textcite{DBLP:conf/icail/MaranhaoS19} and the references therein) can be fruitfully integrated in our approach.
It is the pluralistic nature of \logikey,
realised within a dynamic modelling framework (e.g.~{\Isabelle}), that enables and
supports such improvements without requiring expensive technical adjustments to
the underlying base reasoning technology.

As a broader application scenario, we are currently proposing that ethico-legal value-oriented theories and ontologies should constitute a core ingredient to enable the computation, assessment and communication of rational
justifications and explanations in the future ethico-legal governance of AI
\parencite{C87}. 
Thus, a sound and trustworthy implementation of any legally accountable
`moral machine' requires the development of formal theories and ontologies for the legal domain to guide and interconnect the encoding of concrete regulatory codes
and legal
cases. Understanding legal reasoning as dialectical practical
argumentation, the pluralist interpretation of concrete legal rules arguably
requires a complementary ethico-legal value-oriented theory such as, e.g., the \textit{discoursive grammar} of justification by
\textcite{lomfeld19_grammatik}, which we formally encoded in this paper.
In this sense, some first positive evidence has been provided regarding challenges that we have previously identified with respect to the ethical-legal governance of future AI systems \parencite{C81}. Indeed, it was this broader vision that primarily motivated our work on value-oriented legal reasoning in the first place.

\section{Conclusion} \label{sec:Conclusion}

We illustrate the application of the \logikey\ knowledge engineering methodology and framework to enable the interdisciplinary collaboration among different specialist roles. In the present case, they are a lawyer and legal philosopher (L.) and two computer scientists (B. and F.) who join forces with the aim of formally modelling a value-oriented legal theory \parencite[\textit{discoursive grammar} by][]{lomfeld19_grammatik} for the sake of providing means for computer-automated prediction and assessment of legal case decisions.

From a technical perspective, the core objective of this article has
been to demonstrate that the
\logikey\ methodology appears indeed suitable for the task of
value-oriented legal reasoning.
As instantiated in the present work, the \logikey\ methodology builds upon a HOL-encoding of a modal logic of preferences to model a
domain-specific theory of value-based legal balancing.
In combination with further legal and world knowledge this theory has been
successfully employed for
the formal encoding and computer-supported assessment, using the {\Isabelle} system,
of illustrative legal cases in property law (``wild animal cases'').

It is the flexibility of the
multi-layer modelling which is novel in our approach, in combination
with a very rich support for automated reasoning in expressive, quantified classical and
non-classical logics, thereby rejecting the idea that knowledge
representation means should be limited \emph{prima facie} to
decidable logic frameworks, due to complexity or performance
  considerations.  In the \logikey\ approach, the choice of a
  particular object logic is deliberately left to the knowledge
  engineer. The range of options varies from well-manageable decidable
  logics to sophisticated quantified non-classical logics and
  combinations thereof, depending on what is best suited to handle a
  particular knowledge representation (and reasoning) task at hand.

From a legal perspective, the reconstruction of legal balancing is,
already with classical argumentative tools, a non-trivial task, which is
methodologically not yet settled \parencite{sieckmann10}. Here, our
work proposed the structuring of legal balancing by means of a
dialectical ethico-legal value system (\textit{discoursive grammar}).
Legal rules and their various interpretations can thus be represented within a unified yet
pluralistic logic of value preferences. The integration of
this logic and the value system within the dynamic HOL-based
modelling environment allows us to experiment with different forms of
interpretation. This enables us, not only to find more accurate
reconstructions of legal argumentation, but also supports the modelling
of value-based legal balancing, taking into account notions of value preference,
aggregation, promotion and conflict; and also in a manner amenable to computer
automation. 
The modelling of Lomfeld's legal theory in \logikey\ enabled us to successfully predict (and to some extend justify) case outcomes by `just using logic', employing qualitative value preferences without the necessity to bring in numbers and weights into the model.

From a general perspective, supporting interactive and automated
value-oriented legal argumentation on the computer is a non-trivial
challenge which we address, for reasons as defended, e.g., by
Bench-Capon (\citeyear{bench-capon20}), with symbolic AI techniques
and formal methods. Motivated by recent pleas for \textit{explainable
  and trustworthy AI}, our primary goal is to work towards the
development of ethico-legal governors for future generations of
intelligent systems, or more generally, towards some form of legally
and ethically \textit{reasonable machines} \parencite{C87} capable
of exchanging rational justifications for the actions they take. While
building up a capacity to engage in value-oriented legal argumentation
is just one of a multitude of challenges this vision is faced with, it
clearly constitutes an important stepping stone towards this ambitious long-term goal.
 
\begin{acknowledgements}
  We thank the unknown reviewers of our prior paper at the MLR 2020
  workshop for their valuable comments and suggestions that have led
  to significant improvements of this article.
\end{acknowledgements}

\printbibliography

@Article{lomfeld19_grammatik,
  author = 	 {Bertram Lomfeld},
  title = 	 {Grammatik der Rechtfertigung: Eine kritische Rekonstruktion der Rechts(fort)bildung},
  journal = 	 {Kritische Justiz},
  year = 	 2019,
  volume = 	 52,
  number = 	 4}

@inproceedings{lomfeld17_methode,
  author = 	 {Bertram Lomfeld},
  title = 	 {Vor den Fällen: Methoden soziologischer Jurisprudenz},
  booktitle = {Die Fälle der Gesellschaft: Eine neue Praxis soziologischer Jurisprudenz},
  publisher = {Mohr Siebeck},
  address =	 {Tübingen},
  year      = 2017,
  editor = 	 {Lomfeld},
  pages = 	 {1-16}}

@Book{lomfeld15_vertrag,
  author = 	 {Lomfeld, Bertram},
  title = 	 {Die Gründe des Vertrages: Eine Diskurstheorie der Vertragsrechte},
  publisher = 	 {Mohr Siebeck},
  address =	 {Tübingen},
  year = 	 2015}

@incollection{B22,
  Author =	 {Benzm{\"u}ller, Christoph and Farjami, Ali and
                  Parent, Xavier},		  
  editor = 	 {Rahman, Shahid and Armgardt, Matthias and Nordtveit Kvernenes, Hans Christian},
  booktitle = 	 {New Developments in Legal Reasoning and Logic:
From Ancient Law to Modern Legal Systems},
  title = 	 {Dyadic Deontic Logic in HOL: Faithful Embedding and Meta-Theoretical Experiments},
  publisher = 	 {Springer Nature Switzerland AG},
  year = 	 2022,
  doi = {978-3-030-70084-3_14},
  pages = {353-377},
  Addendum =	 {\url{https://doi.org/10.1007/978-3-030-70084-3_14}}
  }

@InProceedings{C87,
  Author =	 {Christoph Benzm{\"u}ller and Bertram Lomfeld},
  Title =        {Reasonable Machines: A Research Manifesto},
  Addendum =	 {Preprint: \url{https://dx.doi.org/10.13140/RG.2.2.28918.63045}},
  Url =		 {https://doi.org/10.1007/978-3-030-58285-2_20},
  Doi =		 {10.1007/978-3-030-58285-2_20},
  Booktitle = {{KI} 2020: Advances in Artificial Intelligence -- 43rd German Conference on Artificial Intelligence, Bamberg, Germany, September 21–25, 2020, Proceedings},
  editor    = {Ute Schmid and Franziska Klügl and Diedrich Wolter},
  series    = {Lecture Notes in Artificial Intelligence},
  volume    = {12352},
  pages     = {251--258},
  publisher = {Springer, Cham},
  year      = 2020,
  isbn      = {978-3-030-30178-1},
}

@inproceedings{C81,
  Author =	 {David Fuenmayor and Christoph Benzm{\"u}ller},
  Booktitle =    {{ECAI} 2020 -- 24th European Conference on Artificial Intelligence, June 8-12, Santiago de Compostela, Spain},
  Editor =	 {De Giacomo, G. and Catala, A. and  Dilkina, B. and  Milano, M. and  Barro, S. and  Bugarín, A. and  Lang, J.},
  Series    =    {Frontiers in Artificial Intelligence and Applications},
  volume    = {325},
  Publisher =    {{IOS} Press},
  Year      =    {2020},
  Url = {https://dx.doi.org/10.3233/FAIA200445},
  Title =	 {Normative Reasoning with Expressive Logic Combinations},
  Addendum =	 {Preprint: \url{https://www.researchgate.net/publication/339413206}},
  pages =	 {2903-2904},
  doi =       {10.3233/FAIA200445},
}

@incollection{B19,
  Author =	 {Fuenmayor, David and Benzm{\"u}ller, Christoph},
  Title =	 {A Computational-Hermeneutic Approach for Conceptual Explicitation},
  Booktitle =	 {Model-Based Reasoning in Science and Technology. Inferential Models for Logic, Language, Cognition and Computation},
  Editor =	 {A. Nepomuceno and L. Magnani and F. Salguero and C. Bares and M. Fontaine},
  Publisher =	 {Springer},
  Series =       {SAPERE},
  Volume = {49},
  Pages = {441-469},
  Year =	 2019,
  Doi =  {10.1007/978-3-030-32722-4_25},
  OPTaddendum = 	 {Preprint \url{http://doi.org/10.13140/RG.2.2.30869.78564}},
}

@article{J48,
Author =	 {Benzm{\"u}ller, Christoph and Parent, Xavier and van
                  der Torre, Leendert},
Journal =	 {Artificial Intelligence},
Publisher = {Elsevier},
Title =	 {Designing Normative Theories for Ethical and Legal Reasoning:
    LogiKEy Framework, Methodology, and Tool Support},
Year =	 2020,
Volume = {287},
Pages = {103348},
Doi = {10.1016/j.artint.2020.103348},
Url = {https://doi.org/10.1016/j.artint.2020.103348},
Addendum =	 {Preprint: \url{https://www.researchgate.net/publication/342146653} or \url{https://arxiv.org/abs/1903.10187}},
}

@article{J46,
Author =	 {Christoph Benzm{\"u}ller and Ali Farjami and Paul Meder and Xavier Parent},
Journal =	 {Journal of Applied Logics -- IfCoLoG Journal of Logics and their Applications (Special Issue: Reasoning for Legal AI)},
Editor =	 {Robaldo, Livio and van der Torre, Leon},
Title =	 {{I/O} Logic in {HOL}},
Pages =        {715--732},
Volume = 6,
Number = 5,
Addendum =         {Preprint: \url{https://www.researchgate.net/publication/332786587}},
Url =       {https://www.collegepublications.co.uk/ifcolog/?00034},
Year =	 2019,
}

@article{J45,
Author =	 {Christoph Benzm{\"u}ller and Ali Farjami and Xavier Parent},
Journal =	 {Journal of Applied Logics -- IfCoLoG Journal of Logics and their Applications (Special Issue: Reasoning for Legal AI)},
Editor =	 {Robaldo, Livio and van der Torre, Leon},
Title =	 {{{\AA}qvist's} Dyadic Deontic Logic {E} in {HOL}},
Volume = 6,

Number = 5,
Pages =        {733--755},
Addendum =         {Preprint: \url{https://www.researchgate.net/publication/332786724}},
Url =       {https://www.collegepublications.co.uk/ifcolog/?00034},
Year =	 2019,
}

@InCollection{J43,
  author =	 {Benzm\"uller, Christoph and Andrews, Peter},
	title        =	{Church's Type Theory},
	booktitle    =	{The Stanford Encyclopedia of Philosophy},
	editor       =	{Edward N. Zalta},
	howpublished =	{\url{https://plato.stanford.edu/entries/type-theory-church/}},
	year         =	2019,
	edition      =	{Summer 2019},
	pages        =  {1--62 (in pdf version)},
	publisher    =	{Metaphysics Research Lab, Stanford University},
	url          =  {https://plato.stanford.edu/entries/type-theory-church/},
}

@article{J41,
  author =	 {Christoph Benzm{\"u}ller},
  title =	 {Universal (Meta-)Logical Reasoning: Recent
                  Successes},
  journal =	 {Science of Computer Programming},
  year =	 2019,
  volume =	 172,
  pages =	 {48-62},
  Addendum =	 {Preprint:
                  \url{http://doi.org/10.13140/RG.2.2.11039.61609/2}},
  doi =		 {10.1016/j.scico.2018.10.008},
}

@article{J31,
  Author =	 {Christoph Benzm{\"u}ller},
  Doi =		 {10.1007/s10992-016-9403-0},
  Journal =	 {Journal of Philosophical Logic},
  Volume =	 46,
  Number =	 3,
  Pages =	 {333-353},
  Title =	 {Cut-Elimination for Quantified Conditional Logic},
  Addendum =	 {Preprint:
                  \url{https://www.researchgate.net/publication/293488069}},
  Comment =	 {<a href="http://rdcu.be/npVQ">Springer Nature
                  Link</a>},
  Year =	 2017,
}

@incollection{B5,
  Author =	 {Christoph Benzm{\"u}ller and Dale Miller},
  Booktitle =	 {Handbook of the History of Logic, Volume 9 ---
                  Computational Logic},
  Doi =		 {10.1016/B978-0-444-51624-4.50005-8},
  Editor =	 {Gabbay, Dov M. and Siekmann, J\"org H. and Woods,
                  John},
  Isbn =	 {978-0-444-51624-4},
  Pages =	 {215-254},
  Publisher =	 {North Holland, Elsevier},
  Title =	 {Automation of Higher-Order Logic},
  Addendum =	 {Preprint:
                  \url{http://christoph-benzmueller.de/papers/B5.pdf}},
  Year =	 2014,
}

@inproceedings{C37,
  Address =	 {Beijing, China},
  Author =	 {Christoph Benzm{\"u}ller},
  Booktitle =	 {23rd International Joint Conference on Artificial
                  Intelligence (IJCAI-13)},
  Comment =	 {<a href="http://christoph-benzmueller.de/papers/2013-IJCAI-Poster.pdf">poster</a>},
  Editor =	 {Francesca Rossi},
  Publisher =	 {AAAI Press},
  Isbn =	 {978-1-57735-633-2},
  Pages =	 {746-753},
  Title =	 {Automating Quantified Conditional Logics in {HOL}},
  Addendum =	 {Preprint:
                  \url{http://christoph-benzmueller.de/papers/C37.pdf}},
  Year =	 2013,
}

@article{J21,
  Author =	 {Christoph Benzm{\"u}ller and Paulson, Lawrence C.},
  Doi =		 {10.1093/jigpal/jzp080},
  Journal =	 {The Logic Journal of the IGPL},
  Number =	 6,
  Pages =	 {881-892},
  Title =	 {Multimodal and Intuitionistic Logics in Simple Type
                  Theory},
  Addendum =	 {Preprint:
                  \url{http://christoph-benzmueller.de/papers/J21.pdf}},
  Volume =	 18,
  Year =	 2010,
}

@article{J23,
  Author =	 {Christoph Benzm{\"u}ller and Paulson, Lawrence C.},
  Doi =		 {10.1007/s11787-012-0052-y},
  Journal =	 {Logica Universalis (Special Issue on Multimodal
                  Logics)},
  Number =	 1,
  Pages =	 {7-20},
  Title =	 {Quantified Multimodal Logics in Simple Type Theory},
  Addendum =	 {Preprint:
                  \url{https://www.researchgate.net/publication/221677897}},
  Volume =	 7,
  Year =	 2013,
}

@article{J6,
  Author =	 {Christoph Benzm{\"u}ller and Chad Brown and Michael
                  Kohlhase},
  Doi =		 {10.2178/jsl/1102022211},
  Journal =	 {Journal of Symbolic Logic},
  Number =	 4,
  Pages =	 {1027-1088},
  Title =	 {Higher-Order Semantics and Extensionality},
  Addendum =	 {Preprint:
                  \url{https://www.researchgate.net/publication/38338872}},
  Volume =	 69,
  Year =	 2004,
}

@techreport{R2,
  Author =	 {Christoph Benzm{\"u}ller},
  Institution =	 {Saarland University, SEKI Publications (ISSN
                  1437-4447)},
  Number =	 {A/06/93},
  Pages =	 {1-59},
  Title =	 {{HDMS--A und OBSCURE in KORSO--- Die Funktionale
                  Essenz von HDMS--A aus Sicht der algorithmischen
                  Spezifikationsmethode --- TEIL 3: Spezifikation der
                  atomaren Funktionen}},
  Type =	 {Technischer Bericht},
  Addendum =	 {Preprint:
                  \url{http://christoph-benzmueller.de/papers/R2.pdf}},
  Year =	 1993,
}

@article{carnielli-coniglio-fuenmayor-2021,
    title={Logics of Formal Inconsistency Enriched with Replacement: An Algebraic and Modal Account},
    DOI={10.1017/S1755020321000277},
    journal={The Review of Symbolic Logic},
    publisher={Cambridge University Press},
    author={Carnielli, Walter and Coniglio, Marcelo E. and Fuenmayor, David},
    year={2021},
    pages={1–36},
    note={online first}
}

@techreport{Arkin2009,
OPTtitle = {An Ethical Governor for Constraining Lethal Action in an Autonomous System},
author = {Arkin, Ronald C. and Ulam, Patrick and Duncan, Brittany A.},
school = {Georgia Institute of Technology},
number = {GVU-09-02},
year = {2009},
}

@book{rawls1999TJ,
  title={A Theory of Justice},
  author={Rawls, John},
  year={1971},
  note={Revised edition 1999},
  publisher={Harvard university press}
}

@book{goodman1955FFF,
  title={Fact, fiction, and forecast},
  author={Goodman, Nelson},
  year={1955},
  publisher={Harvard University Press}
}

@InCollection{sep-reflective-equilibrium,
	author       =	{Daniels, Norman},
	title        =	{{Reflective Equilibrium}},
	booktitle    =	{The {Stanford} Encyclopedia of Philosophy},
	editor       =	{Edward N. Zalta},
	howpublished =	{\url{https://plato.stanford.edu/archives/sum2020/entries/reflective-equilibrium/}},
	year         =	{2020},
	edition      =	{{S}ummer 2020},
	publisher    =	{Metaphysics Research Lab, Stanford University}
}

@article{blanchette2016hammering,
  title={Hammering towards {QED}},
  author={Blanchette, Jasmin C. and Kaliszyk, Cezary and Paulson, Lawrence C. and Urban, Josef},
  journal={Journal of Formalized Reasoning},
  volume={9},
  number={1},
  pages={101--148},
  year={2016}
}

@InCollection{sep-logic-combining,
	author       =	{Carnielli, Walter and Coniglio, Marcelo Esteban},
	title        =	{{Combining Logics}},
	booktitle    =	{The {Stanford} Encyclopedia of Philosophy},
	editor       =	{Edward N. Zalta},
	howpublished =	{\url{https://plato.stanford.edu/archives/fall2020/entries/logic-combining/}},
	year         =	{2020},
	edition      =	{{F}all 2020},
	publisher    =	{Metaphysics Research Lab, Stanford University}
}

@article{gordon2006pierson,
  title={Pierson vs. {Post} revisited},
  author={Gordon, Thomas F. and Walton, Douglas},
  journal={Frontiers in Artificial Intelligence and Applications},
  volume={144},
  pages={208},
  year={2006},
  publisher={IOS Press}
}

@InCollection{sep-reasoning-defeasible,
	author       =	{Koons, Robert},
	title        =	{{Defeasible Reasoning}},
	booktitle    =	{The {Stanford} Encyclopedia of Philosophy},
	editor       =	{Edward N. Zalta},
	howpublished =	{\url{https://plato.stanford.edu/archives/win2017/entries/reasoning-defeasible/}},
	year         =	{2017},
	edition      =	{Winter 2017},
	publisher    =	{Metaphysics Research Lab, Stanford University}
}

@Inbook{Erne2004,
author="Ern{\'e}, Marcel",
editor="Denecke, K.
and Ern{\'e}, M.
and Wismath, S. L.",
title="Adjunctions and Galois Connections: Origins, History and Development",
bookTitle="Galois Connections and Applications",
year="2004",
publisher="Springer Netherlands",
address="Dordrecht",
pages="1--138",
OPTisbn="978-1-4020-1898-5",
doi="10.1007/978-1-4020-1898-5_1",
url="https://doi.org/10.1007/978-1-4020-1898-5_1"
}

@Inbook{Benthem2009,
author="{van Benthem}, Johan",
editor="Gr{\"u}ne-Yanoff, Till
and Hansson, Sven Ove",
title="For Better or for Worse: Dynamic Logics of Preference",
bookTitle="Preference Change: Approaches from Philosophy, Economics and Psychology",
year="2009",
publisher="Springer Netherlands",
address="Dordrecht",
pages="57--84",
OPTisbn="978-90-481-2593-7",
doi="10.1007/978-90-481-2593-7_3",
url="https://doi.org/10.1007/978-90-481-2593-7_3"
}

@inproceedings{DBLP:conf/icail/HoekstraBBB09,
  author    = {Rinke Hoekstra and
               Joost Breuker and
               Marcello Di Bello and
               Alexander Boer},
  editor    = {Joost Breuker and et.al.},
  title     = {{LKIF} Core: Principled Ontology Development for the Legal Domain},
  booktitle = {Law, Ontologies and the Semantic Web - Channelling the Legal Information
               Flood},
  series    = {Frontiers in Artificial Intelligence and Applications},
  volume    = {188},
  pages     = {21--52},
  publisher = {{IOS} Press},
  year      = {2009},
  url       = {https://doi.org/10.3233/978-1-58603-942-4-21},
  doi       = {10.3233/978-1-58603-942-4-21},
  bibsource = {dblp computer science bibliography, https://dblp.org}
}

@article{CasanovasPPEV14,
  author    = {Pompeu Casanovas and
               Monica Palmirani and
               Silvio Peroni and
               Tom M. van Engers and
               Fabio Vitali},
  title     = {Semantic Web for the Legal Domain: The next step},
  journal   = {Semantic Web},
  volume    = {7},
  number    = {3},
  pages     = {213--227},
  year      = {2016},
  url       = {https://doi.org/10.3233/SW-160224},
  doi       = {10.3233/SW-160224},
  bibsource = {dblp computer science bibliography, https://dblp.org}
}

@inproceedings{C91,
  Author =	 {Benzm\"uller, Christoph and Fuenmayor, David},
  Editor =      {Liron Cohen and Cezary Kaliszyk},
  doi =          {10.4230/LIPIcs.ITP.2021.7},
  Title =	 {Value-oriented Legal Argumentation in {Isabelle/HOL}},
  Booktitle =    {{International Conference on Interactive Theorem Proving (ITP), Proceedings}},
  series =       {LIPIcs},
  volume =    {193},
  number =    {23},
  pages =     {23:1-23:18},
  publisher =    {Schloss Dagstuhl - Leibniz-Zentrum f{\"{u}}r Informatik},
  Addendum =     {Preprint: \url{https://dx.doi.org/10.13140/RG.2.2.21514.80320}},
  Year =	 2021,
}

@book{MerillSmith17,
 author = {Thomas W. Merrill and Henry E. Smith},
 title = {Property: Principles and Policies},
 publisher = {Foundation Press},
 year = {2017},
 url = {https://scholarship.law.columbia.edu/books/278},
}

@InProceedings{bench-capon20,
  author = 	 {Bench-Capon, Trevor},
  editor = 	 {H\"otzendorfer, W. and Tschol, C. and Kummer, F.},
  booktitle = 	 {In International Trends in Legal Informatics: A Festschrift for Erich Schweighofer},
  title = 	 {The Need for Good Old-Fashioned {AI} and {L}aw.},
  publisher = 	 {Weblaw AG},
  year = 	 2020}

@inproceedings{rissland1987case,
  title={A case-based system for trade secrets law},
  author={Rissland, Edwina L. and Ashley, Kevin D.},
  booktitle={Proceedings of the 1st international conference on Artificial Intelligence and Law},
  pages={60--66},
  year={1987}
}

@article{Topological_Semantics-AFP,
  author  = {David Fuenmayor},
  title   = {Topological semantics for paraconsistent and paracomplete logics},
  journal = {Archive of Formal Proofs},
  OPTmonth   = dec,
  year    = 2020,
  note    = {\url{https://isa-afp.org/entries/Topological_Semantics.html},
            Formal proof development},
  ISSN    = {2150-914x},
}

@Article{Krause1995,
  author = 	 {Krause, P. and  Ambler, S. and  Elvang--Goransson and  M. and Fox, J.},
  title = 	 {A Logic Of Argumentation for Reasoning under Uncertainty},
  journal = 	 {Computational Intelligence},
  year = 	 {1995},
  doi =          {https://doi.org/10.1111/j.1467-8640.1995.tb00025.x},
  OPTkey = 	 {},
  OPTvolume = 	 {11},
  OPTnumber = 	 {},
  OPTpages = 	 {113-131},
  OPTmonth = 	 {},
  OPTnote = 	 {},
  OPTannote = 	 {}
}

@inproceedings{weide_ea2010,
  author =	 {van der Weide, T.L. and Dignum, F. and J.-J.Ch. Meyer, J._J.Ch. and Prakken, H. and Vreeswijk, G.A.W.},
  title =	 {Practical Reasoning Using Values},
  booktitle =	 {Argumentation in Multi-Agent Systems (ArgMAS)},
  editor = {Peter McBurney and Iyad Rahwan and  Simon Parsons and Nicolas Maudet},
  publisher =  {Berlin: Springer},
  pages    = {79-93},
  year =	 {2010},
}

@book{grabmair2016,
  title =	 {Modeling purposive legal argumentation and case outcome prediction using argument schemes in the value judgment formalism},
  author =	 {Grabmair, Matthias},
  year =	 2016,
  publisher =	 {Dissertation University of Pittsburgh},
}

@article{bench-capon2020,
  author    = {Bench-Capon, TJM},
  title     = {Ethical approaches and autonomous systems},
  journal   = {Artificial Intelligence},
  volume    = {281},
  year      = {2020},
  url       = {https://doi.org/10.1016/j.artint.2020.103239}
}

@book{hofstede2001,
  title =	 {Culture's Consequences},
  author =	 {Geert Hofstede},
  year =	 2001,
  publisher =	 {Thousands Oaks: Sage},
}

@book{inglehart2018,
  title =	 {Cultural Evolution},
  author =	 {Inglehart, Ronald},
  year =	 2018,
  publisher =	 {Cambridge University Press},
}

@book{clark1991,
  title =	 {Political Economy: A Comparative Approach},
  author =	 {Clark, Barry},
  year =	 1991,
  publisher =	 {New York: Praeger},
}

@book{mitchell2007,
  title =	 {Eight Ways to Run the Country},
  author =	 {Mitchell, Brian},
  year =	 2007,
  publisher =	 {Westport: Praeger},
}

@book{eysenck1954,
  title =	 {The Psychology of Politics},
  author =	 {Eysenck, Hans},
  year =	 1954,
  publisher =	 {London: Routledge},
}

@book{rokeach1973,
  title =	 {The Nature of Human Values},
  author =	 {Rokeach, Milton},
  year =	 1973,
  publisher =	 {New York: Free Press Macmillan},
}

@article{schwartz1992,
  author    = {Shalom Schwartz},
  title     = {Universals in the Content and Structure of Values},
  journal   = {Advances in Experimental Social Psychology},
  volume    = {25},
  number    = {},
  pages     = {1-65},
  year      = {1992},
}

@inproceedings{smith2003,
  author =	 {Barry Smith},
  title =	 {Ontology},
  booktitle =	 {Blackwell Guide to the Philosophy of Computing and Information},
  editor = {Luciano Floridi},
  publisher =  {Oxford: Blackwell},
  year =	 {2003},
}

@article{gruber1993,
  author    = {Tom Gruber},
  title     = {A Translation Approach to Portable Ontology Specifications},
  journal   = {Knowledge Acquisition},
  volume    = {5},
  number    = {},
  pages     = {199-220},
  year      = {1993},
}

@inproceedings{gruber2009,
  author =	 {Tom Gruber},
  title =	 {Ontology},
  booktitle =	 {Encyclopedia of Database Systems},
  editor = {Ling Liu and M. Tamer Özsu},
  year =	 2009,
  publisher =  {Springer},
}

@book{barak2012,
  title =	 {Proportionality},
  author =	 {Barak, Aharon},
  year =	 2012,
  publisher =	 {Cambridge University Press},
}

@book{neves2021,
  title =	 {Constitutionalism and the Paradox of Principles and Rules},
  author =	 {Marcelo Neves},
  year =	 2021,
  publisher =	 {Oxford University Press},
}

@article{raz1972,
  author    = {Raz, Joseph},
  title     = {Legal Principles and the Limits of Law},
  journal   = {Yale Law Journal},
  volume    = {81},
  number    = {},
  pages     = {823-854},
  year      = {1972},
}

@book{aleven1997,
  title =	 {Teaching case-based reasoning through a model and examples},
  author =	 {Aleven, Vinccent},
  year =	 1997,
  publisher =	 {PhD Dissertation University of Pittsburgh},
  url = {https://citeseerx.ist.psu.edu/viewdoc/download?doi=10.1.1.47.3347&rep=rep1&type=pdf}
}

@book{prakken1997,
  title =	 {Logical Tools for Modelling Legal Argument},
  author =	 {Prakken, Henry},
  year =	 1997,
  publisher =	 {Dordrecht: Springer}
}

@inproceedings{modgil_prakken_2018,
  author =	 {Sanjay Modgil and Henry Prakken},
  title =	 {Abstract Rule-Based Argumentation},
  booktitle =	 {Handbook of Formal Argumentation},
  editor = {Baroni, Pietro and Gabbay, Dov and Giacomin, Massimiliano and {van der Torre}, Leendert},
  pages =	 {287-364},
  year =	 2018,
  publisher =  {College Publications},
}

@book{feteris2017,
  title =	 {Fundamentals of Legal Argumentation},
  author =	 {Feteris, Eveline},
  year =	 2017,
  publisher =	 {Dordrecht: Springer}
}

@article{Schoenfinkel1924,
  author    = {Schönfinkel, Moses},
  title     = {Über die Bausteine der mathematischen Logik},
  journal   = {Mathematische Annalen},
  volume    = {92},
  number    = {},
  pages     = {305-316},
  year      = {1924},
}

@inproceedings{DBLP:conf/icail/MaranhaoS19,
  author =	 {Juliano Maranh{\~{a}}o and Giovanni Sartor},
  title =	 {Value assessment and revision in legal
                  interpretation},
  booktitle =	 {Proceedings of the Seventeenth International
                  Conference on Artificial Intelligence and Law,
                  {ICAIL} 2019, Montreal, QC, Canada, June 17-21,
                  2019},
  pages =	 {219--223},
  year =	 2019,
  url =		 {https://doi.org/10.1145/3322640.3326709},
  doi =		 {10.1145/3322640.3326709},
}

@article{Church40,
  author    = {Alonzo Church},
  title     = {A Formulation of the Simple Theory of Types},
  journal   = {J. Symb. Log.},
  volume    = {5},
  number    = {2},
  pages     = {56--68},
  year      = {1940},
  url       = {https://doi.org/10.2307/2266170},
  doi       = {10.2307/2266170},
  bibsource = {dblp computer science bibliography, https://dblp.org}
}

@book{ganter2012formal,
  title =	 {Formal concept analysis: mathematical foundations},
  author =	 {Ganter, Bernhard and Wille, Rudolf},
  year =	 2012,
  publisher =	 {Springer Berlin}
}

@book{alexy78,
  editor =	 {Robert Alexy},
  title =	 {Theorie der juristischen Argumentation},
  publisher =	 {Frankfurt/M: Suhrkamp},
  year =	 1978
}

@article{alexy00,
  author =	 {Robert Alexy},
  title =	 {On the Structure of Legal Principles},
  journal =	 {Ratio Juris},
  volume =	 13,
  pages =	 {294-304},
  year =	 2000
}

@article{alexy03,
  author =	 {Robert Alexy},
  title =	 {On Balancing and Subsumption: A Structural
                  Comparison},
  journal =	 {Ratio Juris},
  volume =	 16,
  pages =	 {433-449},
  year =	 2003,
}

@book{ashley90,
  title =	 {Modelling Legal Argument: Reasoning with Cases and
                  Hypotheticals},
  author =	 {Kevin D. Ashley},
  year =	 1990,
  publisher =	 {Cambridge/MA: MIT Press},
}

@article{bench-capon_sartor03,
  author =	 {Bench-Capon, Trevor and Sartor, Giovanni },
  title =	 {A model of legal reasoning with cases incorporating
                  theories and value},
  journal =	 {Artificial Intelligence},
  volume =	 150,
  pages =	 {97-143},
  year =	 2003,
}

@article{bench-capon_ea05,
  author =	 {Trevor Bench-Capon and Katie Atkinson and Alison
                  Chorley},
  title =	 {Persuasion and value in legal argument},
  journal =	 {Journal of Logic and Computation},
  volume =	 15,
  pages =	 {1075-1097},
  year =	 2005,
}

@article{bench-capon12,
  author =	 {Trevor Bench-Capon},
  title =	 {Representing Popov v Hayashi with dimensions and
                  factors},
  journal =	 {Artificial Intelligence and Law},
  volume =	 20,
  pages =	 {15-35},
  year =	 2012,
}

@inproceedings{berman_hafner93,
  author =	 {Donald Berman and Carole Hafner},
  title =	 {Representing teleological structure in case-based
                  legal reasoning: the missing link},
  booktitle =	 {Proceedings 4th ICAIL},
  publisher =	 {New York: ACM Press},
  year =	 1993,
  pages =	 {50-59}
}

@article{gordon_walton12,
  author =	 {Thomas Gordon and Douglas Walton},
  title =	 {A Carneades reconstruction of Popov v Hayashi},
  journal =	 {Artificial Intelligence and Law},
  volume =	 20,
  pages =	 {37-56},
  year =	 2012,
}

@book{hage97,
  author =	 {Jaap Hage},
  title =	 {Reasoning With Rules},
  publisher =	 {Dordrecht: Kluwer},
  year =	 1997,
}

@article{horty11,
  author =	 {John Horty},
  title =	 {Rules and reasons in the theory of precedent},
  journal =	 {Legal Theory},
  volume =	 17,
  pages =	 {1-33},
  year =	 2011,
}

@article{prakken_sartor15,
  author =	 {Henry Prakken and Giovanni Sartor},
  title =	 {Law and logic: A review from an argumentation
                  perspective},
  journal =	 {Artificial Intelligence},
  volume =	 227,
  pages =	 {214-225},
  year =	 2015,
}

@article{teubner83,
  author =	 {Gunther Teubner},
  title =	 {Substantive and Reflexive Elements in Modern Law},
  journal =	 {Law \& Society Rev.},
  volume =	 17,
  pages =	 {239-285},
  year =	 1983,
}

@article{sartor10,
  author =	 {Giovanni Sartor},
  title =	 {Doing justice to rights and values: teleological
                  reasoning and proportionality},
  journal =	 {Artificial Intelligence and Law},
  volume =	 18,
  pages =	 {175-215},
  year =	 2010,
}

@inproceedings{sartor18,
  author =	 {Giovanni Sartor},
  title =	 {A Quantitative Approach to Proportionality},
  booktitle =	 {Handbook of Legal Reasoning and Argumentation},
  publisher =	 {Springer},
  address =	 {Dordrecht},
  year =	 2018,
  editor =	 {Bongiovanni et al},
  pages =	 {613-636}
}

@book{sieckmann10,
  editor =	 {Jan-Reinard Sieckmann},
  title =	 {Legal Reasoning: The Methods of Balancing},
  series =	 {ARSP Beiheft},
  volume =	 124,
  publisher =	 {Franz Steiner},
  address =	 {Stuttgart},
  year =	 2010
}

@article{BenthemGR09,
  author =	 {{van Benthem}, Johan and Girard, Patrick and Roy,
                  Olivier},
  title =	 {Everything Else Being Equal: {A} Modal Logic for
                  \emph{Ceteris Paribus} Preferences},
  journal =	 {J. Philos. Log.},
  volume =	 38,
  number =	 1,
  pages =	 {83--125},
  year =	 2009,
  url =		 {https://doi.org/10.1007/s10992-008-9085-3},
  doi =		 {10.1007/s10992-008-9085-3},
  bibsource =	 {dblp computer science bibliography,
                  https://dblp.org}
}

@PhdThesis{Liu2008,
  title={Changing for the better: Preference dynamics and agent diversity},
  author={Liu, Fenrong},
  year={2008},
  school={Inst. for Logic, Language and Computation,
                  Universiteit van Amsterdam}
}

@book{Liu2011,
	location = {Dordrecht},
	title = {Reasoning about Preference Dynamics},
	OPTisbn = {978-94-007-1343-7 978-94-007-1344-4},
	OPTurl = {http://link.springer.com/10.1007/978-94-007-1344-4},
	publisher = {Springer Netherlands},
	author = {Liu, Fenrong},
	urldate = {2021-01-13},
	date = {2011},
	langid = {english},
	doi = {10.1007/978-94-007-1344-4},
}

@article{Moor2009,
  title={Four kinds of ethical robots},
  author={Moor, James},
  journal={Philosophy Now},
  volume={72},
  pages={12--14},
  year={2009}
}

@InProceedings{DeepShallow2,
  author =	 "Svenningsson, Josef and Axelsson, Emil",
  editor =	 "Loidl, Hans-Wolfgang and Pe{\~{n}}a, Ricardo",
  title =	 "Combining Deep and Shallow Embedding for {EDSL}",
  booktitle =	 "Trends in Functional Programming",
  year =	 2013,
  publisher =	 "Springer Berlin Heidelberg",
  address =	 "Berlin, Heidelberg",
  pages =	 "21--36",
  isbn =	 "978-3-642-40447-4"
}

@inproceedings{DeepShallow,
  author =	 {Jeremy Gibbons and Nicolas Wu},
  title =	 {Folding domain-specific languages: deep and shallow
                  embeddings (functional Pearl)},
  booktitle =	 {Proceedings of the 19th {ACM} {SIGPLAN}
                  international conference on Functional programming,
                  Gothenburg, Sweden, September 1-3, 2014},
  pages =	 {339--347},
  doi =		 {10.1145/2628136.2628138},
  editor =	 {Johan Jeuring and Manuel M. T. Chakravarty},
  publisher =	 {{ACM}},
  year =	 2014,
}

@book{LogicCombining,
  title =	 {Analysis and Synthesis of Logics},
  author =	 {Carnielli, Walter and Coniglio, Marcelo and Gabbay,
                  Dov M. and Gouveia Paula and Sernadas, Cristina},
  year =	 2008,
  series =	 {Applied Logics Series},
  number =	 35,
  publisher =	 {Springer}
}

@Article{Andrews:gmdacitt72,
  author =	 "Peter B. Andrews",
  journal =	 "Journal of Symbolic Logic",
  number =	 2,
  pages =	 "385--394",
  title =	 "General Models, Descriptions, and Choice in Type
                  Theory",
  volume =	 37,
  year =	 1972,
}

@article{Andrews72a,
  author    = {Peter B. Andrews},
  title     = {General Models and Extensionality},
  journal   = {J. Symb. Log.},
  volume    = {37},
  number    = {2},
  pages     = {395--397},
  year      = {1972},
  url       = {https://doi.org/10.2307/2272982},
  doi       = {10.2307/2272982},
  bibsource = {dblp computer science bibliography, https://dblp.org}
}

@book{Isabelle,
  author =	 {Nipkow, T. and Paulson, L.C. and Wenzel, M.},
  publisher =	 {Springer},
  series =	 {LNCS},
  title =	 {{{Isabelle/HOL}: A Proof Assistant for Higher-Order
                  Logic}},
  volume =	 2283,
  year =	 2002,
}

@article{Sledgehammer,
  author =	 {Jasmin Christian Blanchette and Sascha B{\"{o}}hme
                  and Paulson, Lawrence C.},
  journal =	 {Journal of Automated Reasoning},
  number =	 1,
  pages =	 {109--128},
  title =	 {Extending {Sledgehammer} with {SMT} Solvers},
  volume =	 51,
  year =	 2013,
}

@book{vonWright1963logic,
  title={The logic of preference},
  author={{von Wright}, Georg Henrik},
  publisher =	 {Edinburgh University Press},
  year={1963}
}

@article{Henkin50,
  author =	 {Henkin, Leon},
  journal =	 {Journal of Symbolic Logic},
  number =	 2,
  pages =	 {81--91},
  title =	 {Completeness in the Theory of Types},
  volume =	 15,
  year =	 1950,
}

@inproceedings{Nitpick,
  author =	 {Jasmin Christian Blanchette and Tobias Nipkow},
  title =	 {Nitpick: {A} Counterexample Generator for
                  Higher-Order Logic Based on a Relational Model
                  Finder},
  booktitle =	 {ITP 2010},
  pages =	 {131--146},
  year =	 2010,
  editor =	 {Matt Kaufmann and Lawrence C. Paulson},
  series =	 {LNCS},
  volume =	 6172,
  publisher =	 {Springer},
}

@inproceedings{E,
  author =	 {Stephan Schulz},
  editor =	 {Kenneth L. McMillan and Aart Middeldorp and Andrei
                  Voronkov},
  title =	 {System Description: {E} 1.8},
  booktitle =	 {Logic for Programming, Artificial Intelligence, and
                  Reasoning - 19th International Conference, LPAR-19,
                  Stellenbosch, South Africa, December 14-19,
                  2013. Proceedings},
  pages =	 {735--743},
  year =	 2013,
  series =	 {LNCS},
  volume =	 8312,
  publisher =	 {Springer},
  url =		 {http://dx.doi.org/10.1007/978-3-642-45221-5},
  doi =		 {10.1007/978-3-642-45221-5},
  isbn =	 {978-3-642-45220-8},
  bibOPTurl =	 {http://dblp.uni-trier.de/rec/bib/conf/lpar/2013},
  bibsource =	 {dblp computer science bibliography, http://dblp.org}
}

@article{Scheutz17,
  author =	 {Matthias Scheutz},
  title =	 {The Case for Explicit Ethical Agents},
  journal =	 {{AI} Magazine},
  volume =	 38,
  number =	 4,
  pages =	 {57--64},
  year =	 2017,
  url =
                  {https://www.aaai.org/ojs/index.php/aimagazine/article/view/2746},
  bibsource =	 {dblp computer science bibliography,
                  https://dblp.org}
}

@article{halpern1997defining,
  title={Defining relative likelihood in partially-ordered preferential structures},
  author={Halpern, Joseph Y.},
  journal={Journal of Artificial Intelligence Research},
  volume={7},
  pages={1--24},
  year={1997}
}

@inproceedings{boutilier1994toward,
  title={Toward a logic for qualitative decision theory},
  author={Boutilier, Craig},
  booktitle={Principles of knowledge representation and reasoning},
  pages={75--86},
  year={1994},
  doi={10.1016/B978-1-4832-1452-8.50104-4},
  organization={Elsevier}
}

@book{ganter2016conceptual,
  title={Conceptual exploration},
  author={Ganter, Bernhard and Obiedkov, Sergei and Rudolph, Sebastian and Stumme, Gerd},
  year={2016},
  publisher={Springer}
}

@article{verheij1998integrated,
  title={An integrated view on rules and principles},
  author={Verheij, Bart and Hage, Jaap C and Van Den Herik, H Jaap},
  journal={Artificial Intelligence and Law},
  volume={6},
  number={1},
  pages={3--26},
  year={1998},
  publisher={Springer}
}

@article{prakken2002exercise,
  title={An exercise in formalising teleological case-based reasoning},
  author={Prakken, Henry},
  journal={Artificial Intelligence and Law},
  volume={10},
  number={1-3},
  pages={113--133},
  year={2002},
  publisher={Springer}
}

@article{bench2002missing,
  title={The missing link revisited: The role of teleology in representing legal argument},
  author={Bench-Capon, Trevor},
  journal={Artificial Intelligence and Law},
  volume={10},
  number={1-3},
  pages={79--94},
  year={2002},
  publisher={Springer}
}

@inproceedings{mccarty1995implementation,
  title={An implementation of Eisner v. Macomber},
  author={McCarty, L Thorne},
  booktitle={Proceedings of the 5th International Conference on Artificial Intelligence and Law},
  pages={276--286},
  year={1995}
}

@book{dworkin-taking-1978,
	location = {Cambridge, Mass},
	title = {Taking rights seriously},
	isbn = {978-0-674-86711-6},
	pagetotal = {371},
	publisher = {Harvard Univ. Press},
	author = {Dworkin, Ronald},
	date = {1978},
	note = {{OCLC}: 4313351},
}

@book{lewis1973counterfactuals,
  title={Counterfactuals},
  author={Lewis, David},
  year={1973},
  publisher={Harvard University Press}
}

@article{bench2017hypo,
  title={Hypo's legacy: introduction to the virtual special issue},
  author={Bench-Capon, Trevor},
  journal={Artificial Intelligence and Law},
  volume={25},
  number={2},
  pages={205--250},
  year={2017},
  publisher={Springer}
}

@article{wenzel2007isabelle,
  title={Isabelle/Isar—a generic framework for human-readable proof documents},
  author={Wenzel, Makarius},
  journal={From Insight to Proof-Festschrift in Honour of Andrzej Trybulec},
  volume={10},
  number={23},
  pages={277--298},
  year={2007},
  publisher={Citeseer}
}

\pagebreak
\begin{appendix}
  \section{Appendix - {\Isabelle} Encoding} \label{TheAppendix}
  \subsection{SSE of \logic\ in HOL} \label{app:PL}
    
\begin{figure}[!bp]
  \centering
  \colorbox{gray!30}{\includegraphics[width=.98\textwidth]{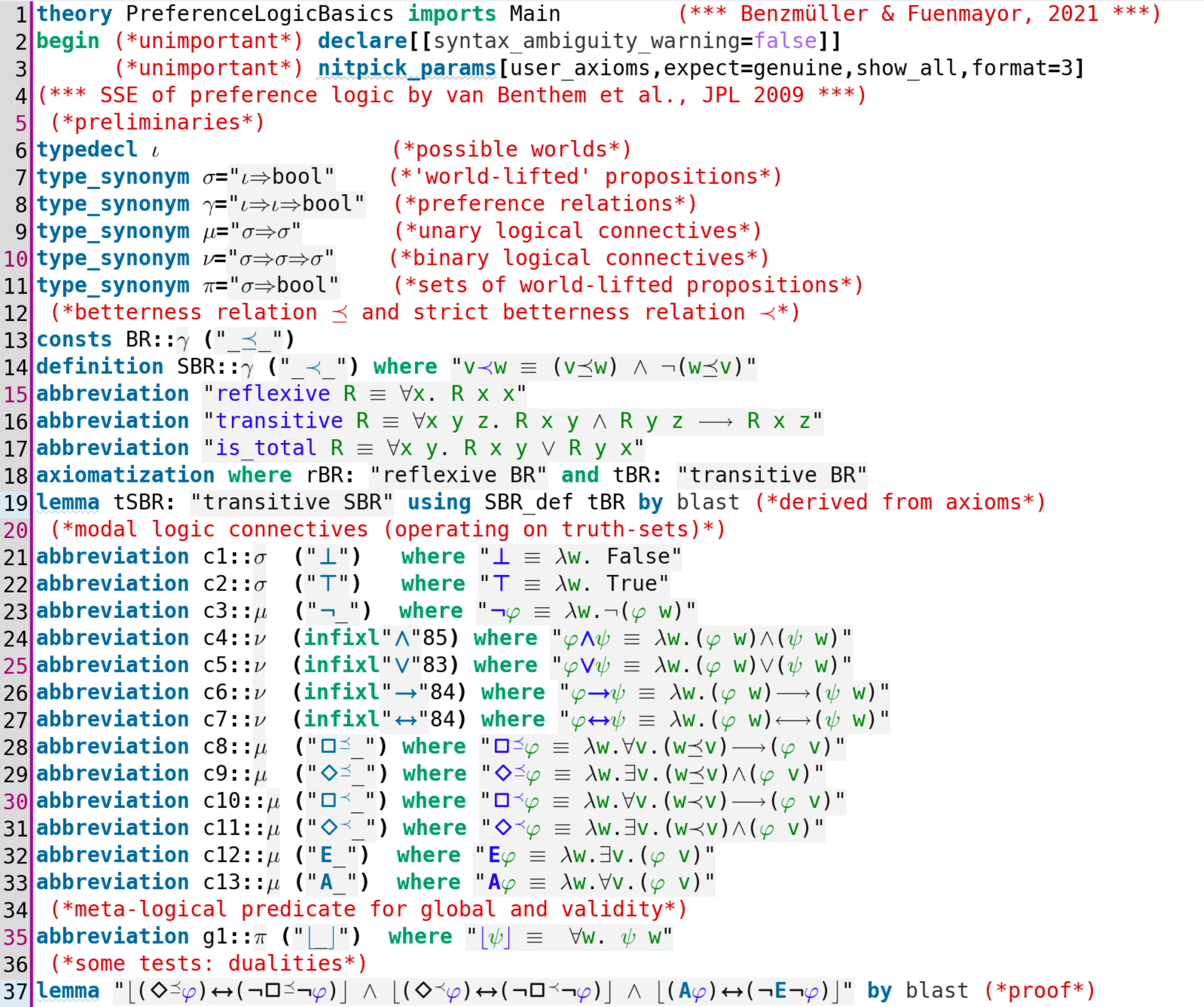}}
  \caption{SSE of \logic\ \parencite{BenthemGR09} in HOL (continued in
    Fig.~\ref{fig:Embedding2})}
  \label{fig:Embedding1}
\end{figure}

\begin{figure}[!htp]
  \centering
  \colorbox{gray!30}{\includegraphics[width=.98\textwidth]{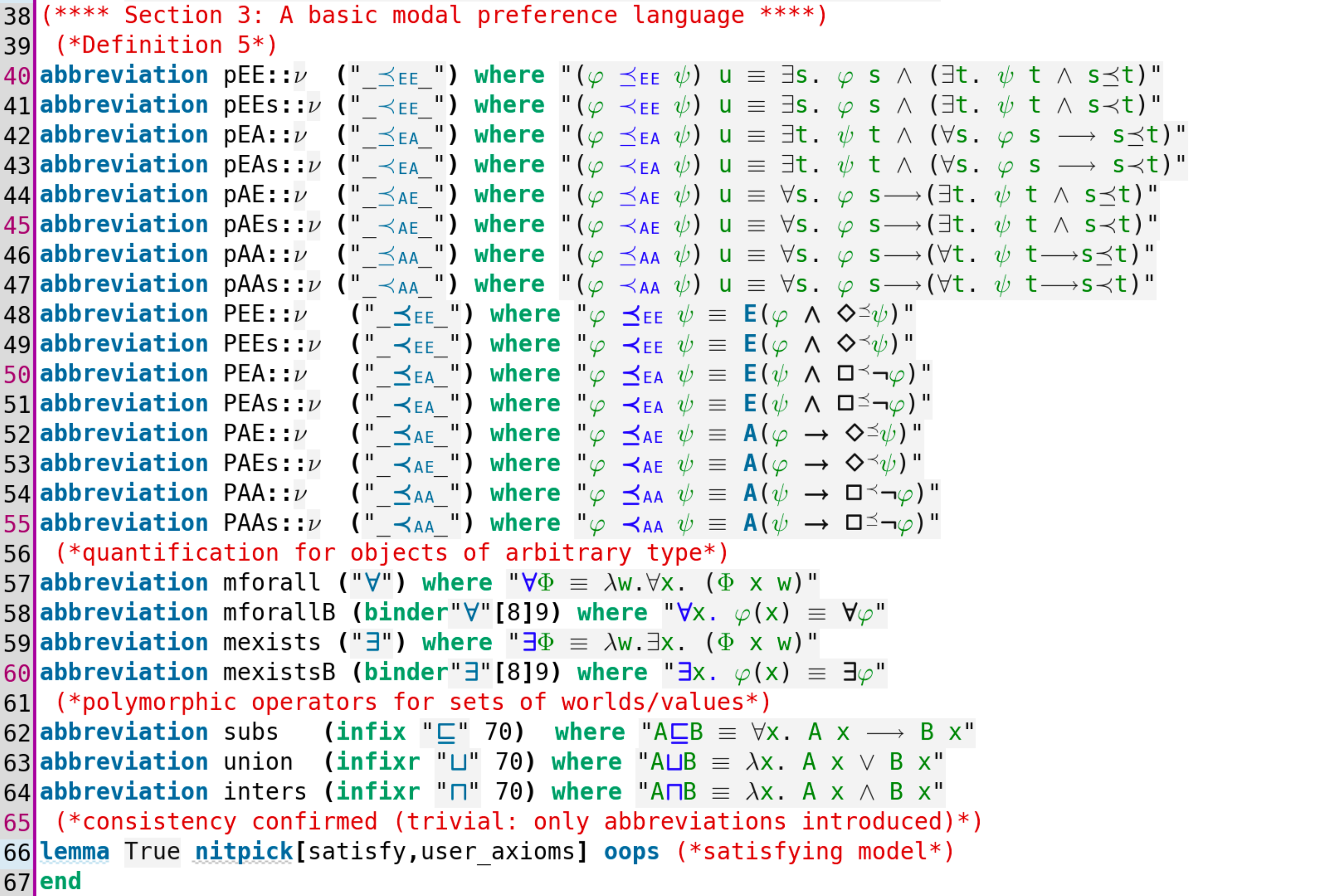}}
  \caption{SSE of \logic\ \parencite{BenthemGR09} in HOL (continued
    from Fig.~\ref{fig:Embedding1})}
  \label{fig:Embedding2}
\end{figure}

  We comment on the implementation of the SSE of \logic\ in
  {\Isabelle} as displayed in
  Figs.~\ref{fig:Embedding1}-\ref{fig:Embedding2}; see van
  \textcite{BenthemGR09} for further details on \logic. The defined
  theory is named "PreferenceLogicBasics" and it relies on base logic
  HOL, imported here as theory "Main".
  
First, a new base type $\itype$ is declared (Line 6), denoting the set
of possible worlds or states. Subsequently (Lines 7--11), useful type
abbreviations are introduced, including the type $\sigma$ for \logic\
propositions, which are modelled as predicates on objects of type
$\itype$ (i.e., as \textit{truth-sets} of worlds/states).  A
\textit{betterness relation} $\preceq$, and its strict variant
$\prec$, are introduced (Lines 13--14), with $\preceq$-accessible
worlds interpreted as those that are \textit{at least as good} as the
present one. Definitions for relation properties are provided, and it
is postulated that $\preceq$ is a preorder, i.e., reflexive and
transitive (Lines 15--18).

Subsequently, the $\sigma$-type lifted logical connectives of \logic\
are introduced as abbreviations of $\lambda$-terms in the meta-logic
HOL (Lines 21--33). The operators $\Box^\preceq$ and $\Box^\prec$ use
$\preceq$ and $\prec$ as guards in their definitions (Lines 28 and
30); analogous for $\Diamond^\preceq$ and $\Diamond^\prec$. An
\textit{universal} modality and its dual are also introduced (Lines
32--33). Moreover, a notion of (global) truth for \logic\ formulas
$\psi$ is defined (Line 35): proposition $\psi$ is globally true, we
also say `valid', if and only if it is true in all worlds.

As a first test some expected dualities of the modal operators are
automatically proved (Line 36).

\begin{figure}[!tp]
  \centering
  \colorbox{gray!30}{\includegraphics[width=.97\textwidth]{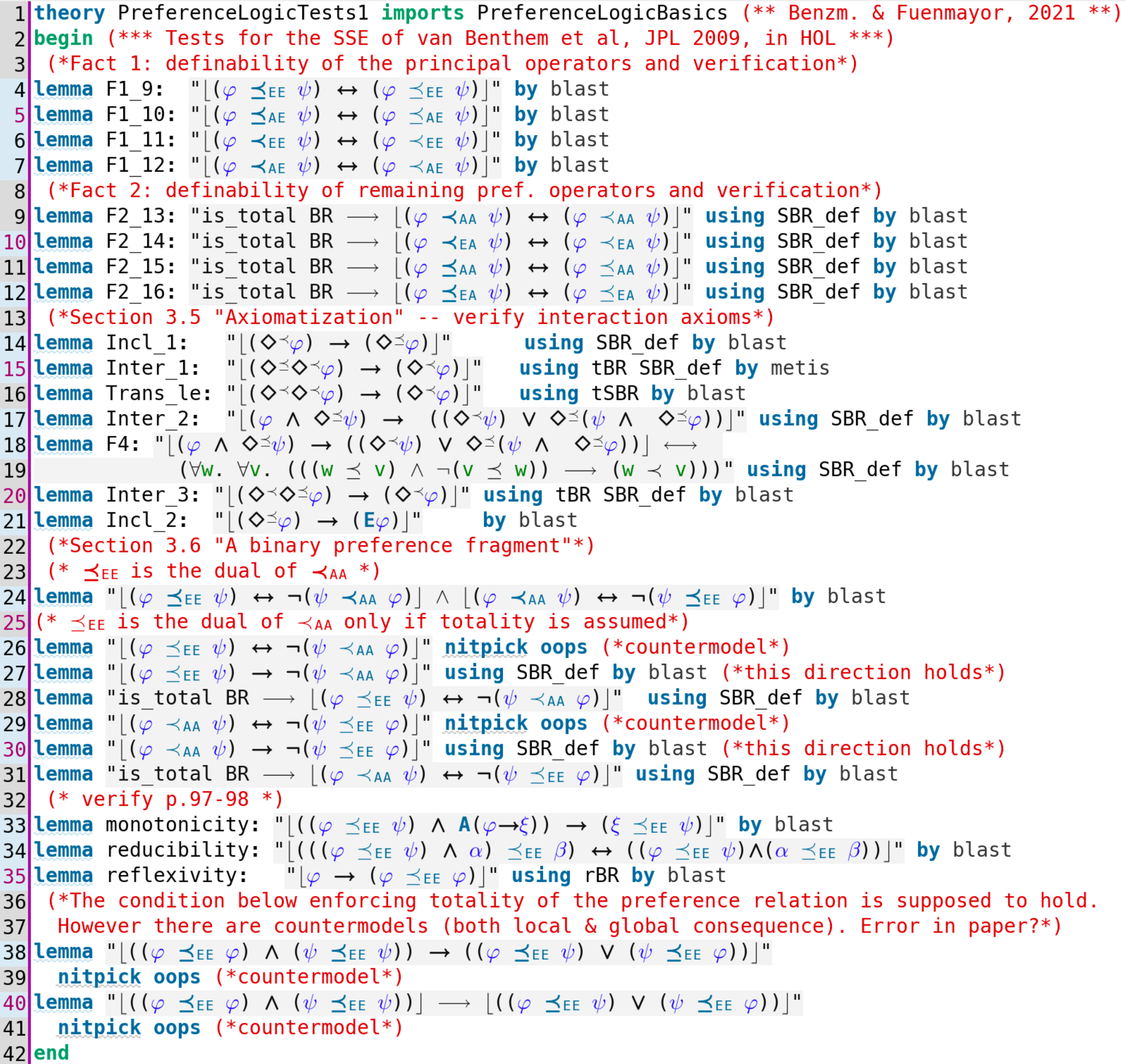}}
  \caption{Experiments: Testing the meta-theory of \logic}
  \label{fig:MetaTheory1}
\end{figure}

Subsequently, the \textit{betterness} ordering $\preceq$
(resp.~$\prec$) is lifted to a preference relation between \logic\
propositions (sets of worlds). Eight possible semantic definitions for
such preferences are encoded in HOL (Lines 40--47 in
Fig.~\ref{fig:Embedding2}).  The semantic definitions are complemented
by eight syntactic definitions of the same binary preferences stated
within the object language \logic\ (Lines 48--55). (ATP systems prove
the meta-theoretic correspondences between these semantic and
syntactic definitions; cf.~Lines 4--12 in Fig.~\ref{fig:MetaTheory1}.)

\logic\ is extended by adding quantifiers (Lines 57--60);
cf.~\parencite{J23} for explanations on the SSE of quantified modal
logics.  Moreover, useful polymorphic operators for subset, union and
intersection are defined (Lines 62--64).

The model finder {\Nitpick} \parencite{Nitpick} confirms the
consistency of the introduced theory (Line 66) by generating and
presenting a model for it (not shown here).

\begin{figure}[!htp]
  \centering
  \colorbox{gray!30}{\includegraphics[width=.97\textwidth]{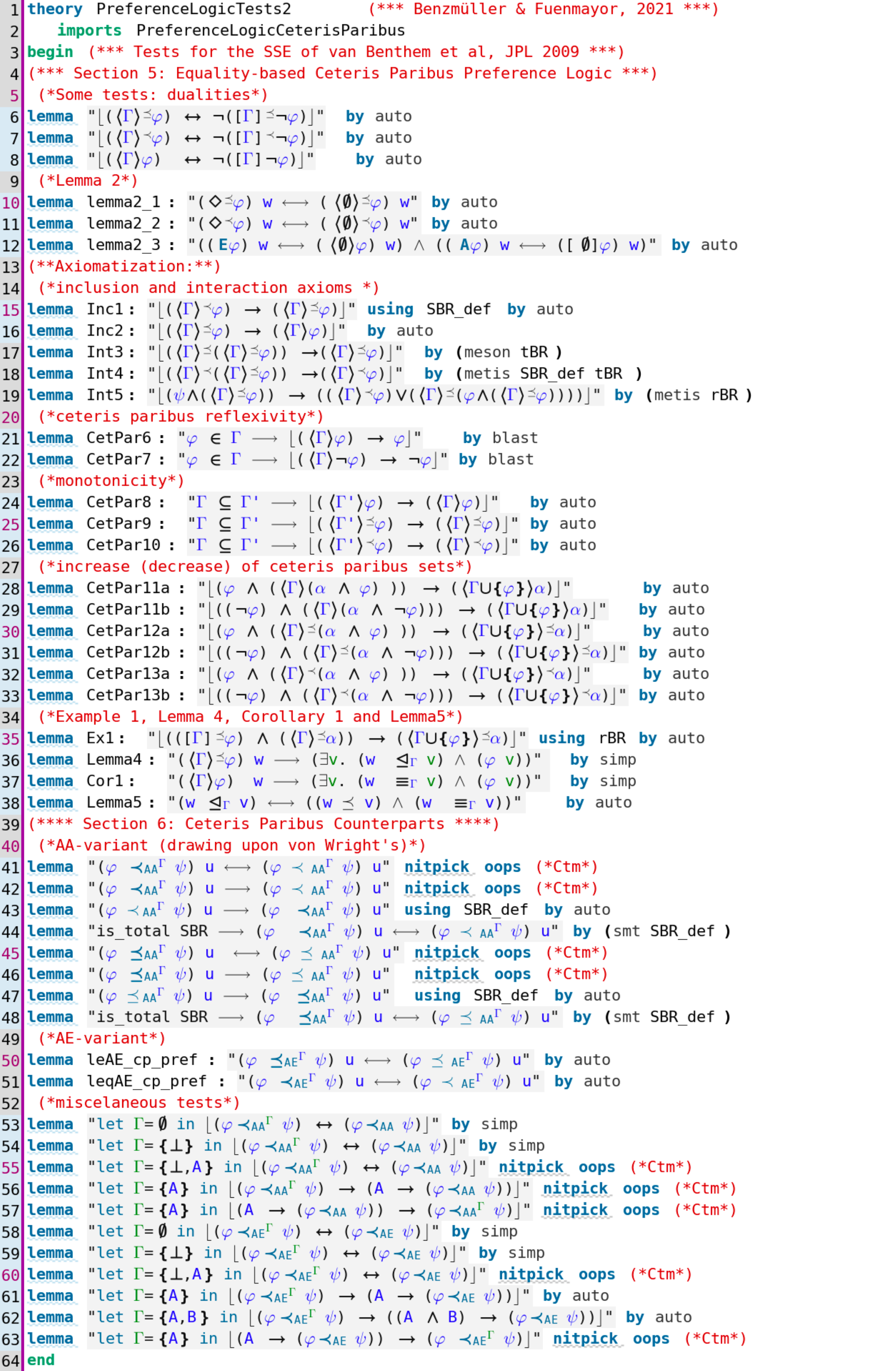}}
  \caption{Experiments (continued): Testing the meta-theory of \logic}
  \label{fig:MetaTheory2}
\end{figure}

To gain practical evidence for the faithfulness of our SSE of \logic\
in {\Isabelle}, and also to assess proof automation performance, we
have conducted numerous experiments in which we automatically
reconstruct meta-theoretical results on \logic; see
Figs.~\ref{fig:MetaTheory1}-\ref{fig:MetaTheory2}.

\begin{figure}[!tp]
  \centering
  \colorbox{gray!30}{\includegraphics[width=.97\textwidth]{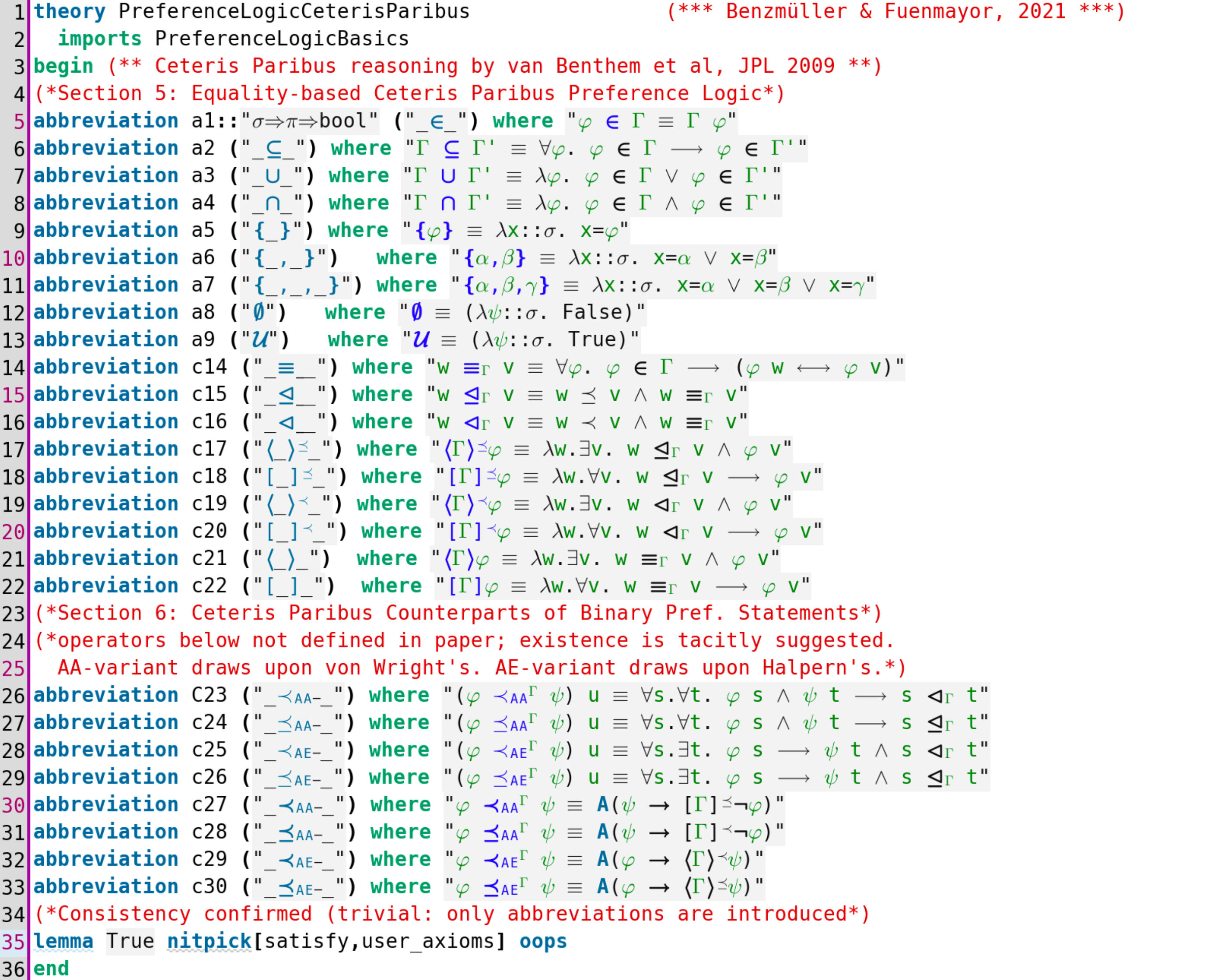}}
  \caption{SSE of \logic\ \parencite{BenthemGR09} in HOL (continued
    from Figs.~\ref{fig:Embedding1}-\ref{fig:Embedding2})}
  \label{fig:CeterisParibus}
\end{figure}

Extending our SSE of \logic\ in HOL some further preference relations
for \logic\ are defined in Fig.~\ref{fig:CeterisParibus}. These
additional relations support \textit{ceteris paribus} reasoning in
\logic.
We give some explanations:
\begin{description}
\item[Lines 5--13] Useful set theoretic notions are introduced as
  abbreviations for corresponding $\lambda$-terms in HOL.
\item[Lines 14--22] \logic\ is further extended with (equality-based)
  \textit{ceteris paribus} preference relations and modalities; here
  $\Gamma$ represents a set of formulas that are assumed constant
  between two possible worlds to compare. Hence our variant can be
  understood as ``these (given) things being equal''-preferences. This
  variant can be used for modelling von Wright's notion of
  \textit{ceteris paribus} (``all other things being equal'')
  preferences, eliciting an appropriate $\Gamma$ by extra-logical
  means.
\item[Lines 26--33:] Except for $\prec^\Gamma_{AA}$, the remaining
  operators we define here were not explicitly defined by
  \textcite{BenthemGR09}; however, their existence is tacitly
  suggested.
\end{description}

\begin{figure}[!tp]
  \centering
  \colorbox{gray!30}{\includegraphics[width=.97\textwidth]{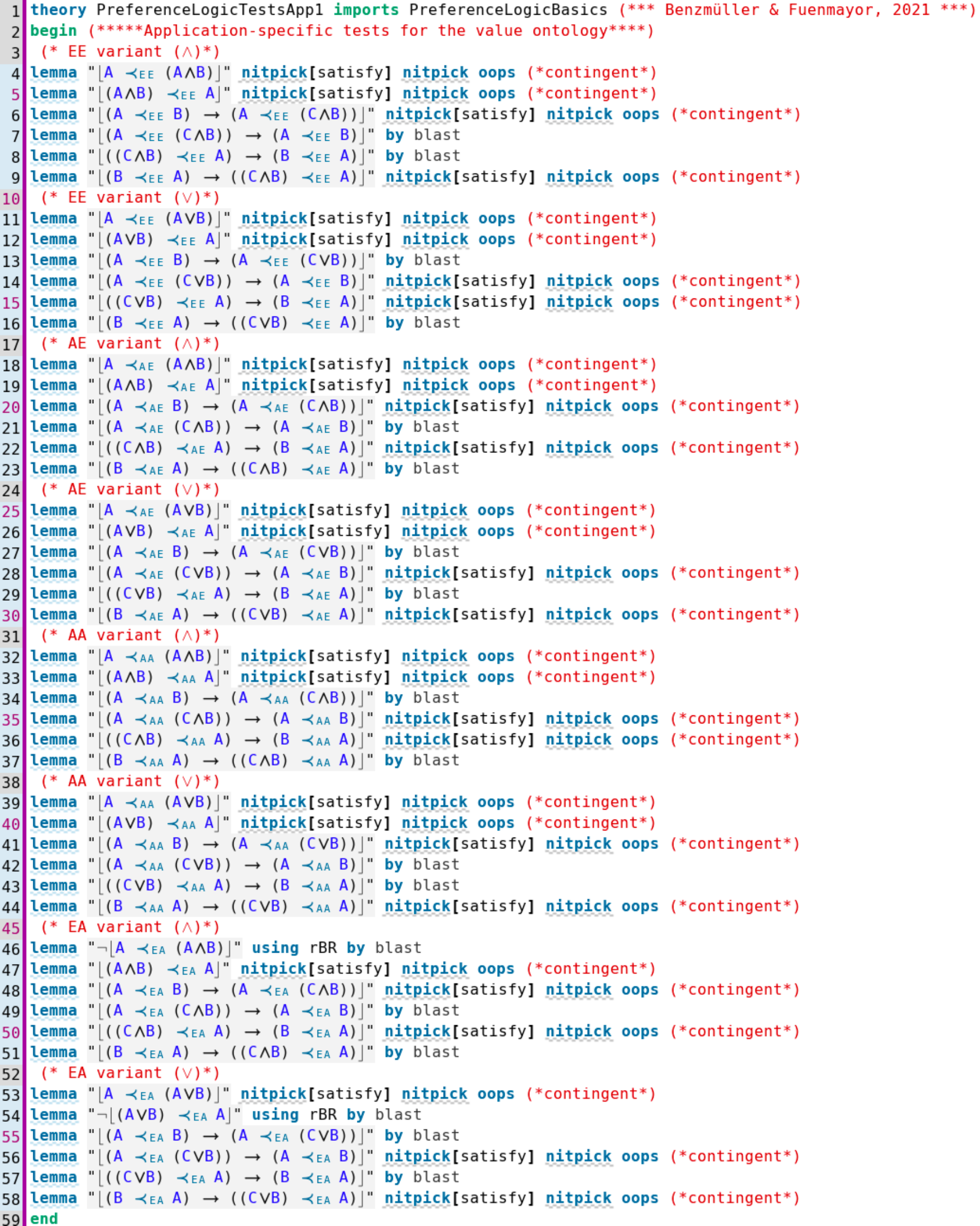}}
  \caption{Experiments (continued): Checking properties of strict
    preference relations}
  \label{fig:MetaTheoryApp1}
\end{figure}

\begin{figure}[!tp]
  \centering
  \colorbox{gray!30}{\includegraphics[width=.97\textwidth]{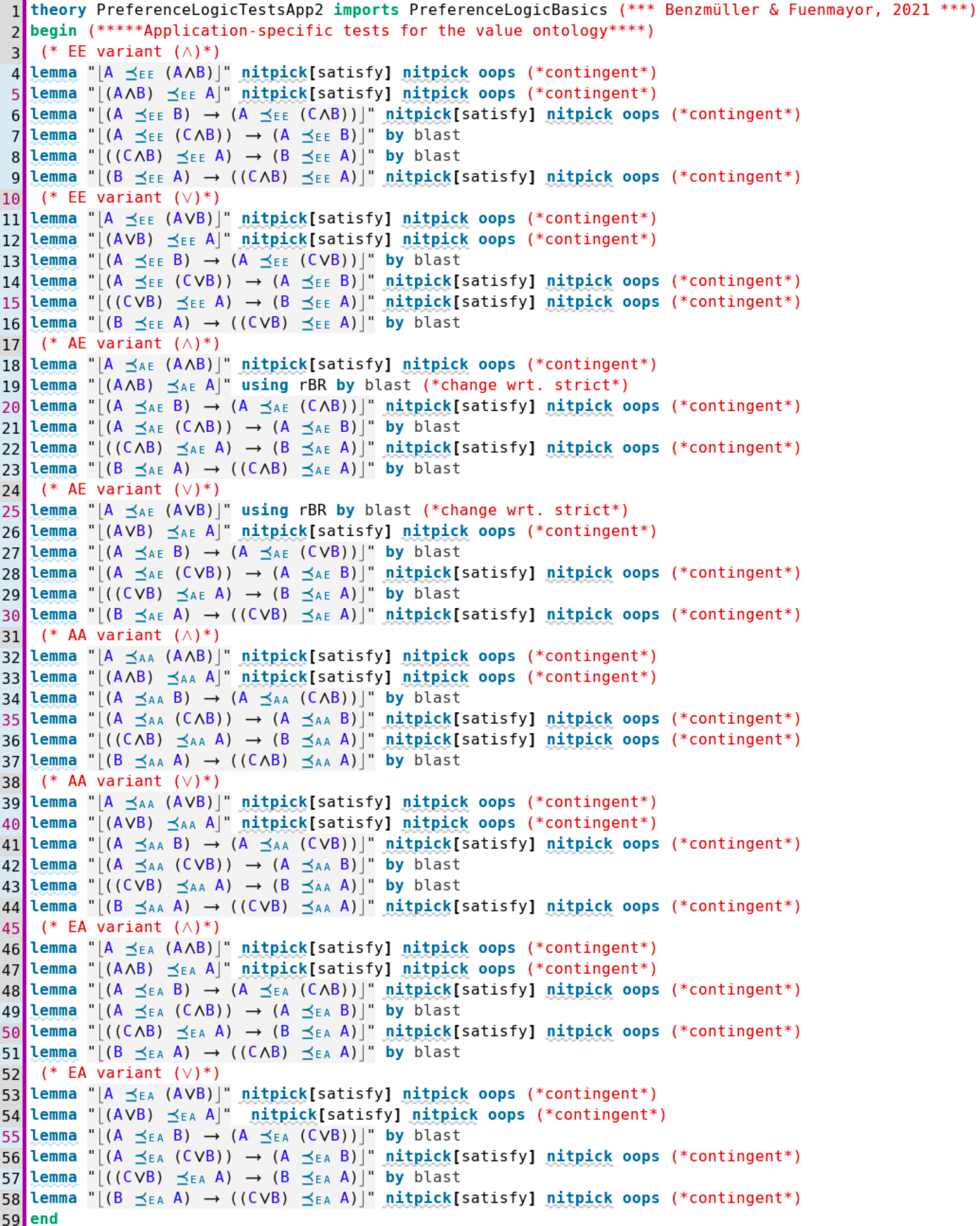}}
  \caption{Experiments (continued): Checking properties of strict
    preference relations}
  \label{fig:MetaTheoryApp2}
\end{figure}

{ Meta-theoretical results on \logic\ as presented by
  \textcite{BenthemGR09} are automatically verified by the reasoning
  tools in {\Isabelle}; see
  Figs.~\ref{fig:MetaTheory1}-\ref{fig:MetaTheoryApp2}; we in fact
  prove all relevant results from \parencite{BenthemGR09}.  The
  experiments shown in Fig.~\ref{fig:MetaTheory1} are briefly
  commented: }

\label{subsec:A2}

\begin{description}
\item[Lines 5--13] Correspondences between the semantically and
  syntactically defined preference relations are proved.
\item[Lines 15--22] It is proved that (e.g.~inclusion and interaction)
  axioms for \logic\ follow as theorems in our SSE. This tests the
  faithfulness of the embedding in one direction.
\item[Lines 25--47] We continue the mechanical verification of
  theorems, and generate countermodels (not displayed here) for
  non-theorems of \logic, thus putting our encoding to the test. Our
  results coincide with the corresponding ones claimed (and in many
  cases proved) in \textcite{BenthemGR09}, except for the claims
  encoded in lines 40-41 and 44-45, where countermodels are reported
  by {\Nitpick}.
\item[Lines 25--47] Some application-specific tests in preparation for
  the modelling of the legal DSL (including the value theory/ontology) are conducted.
\end{description}

\newpage \clearpage

\subsection{Encoding of the Legal DSL (Value Ontology)} \label{app:SecValueOntology}

\begin{figure}[!bp]
  \centering
  \colorbox{gray!30}{\includegraphics[width=.97\textwidth]{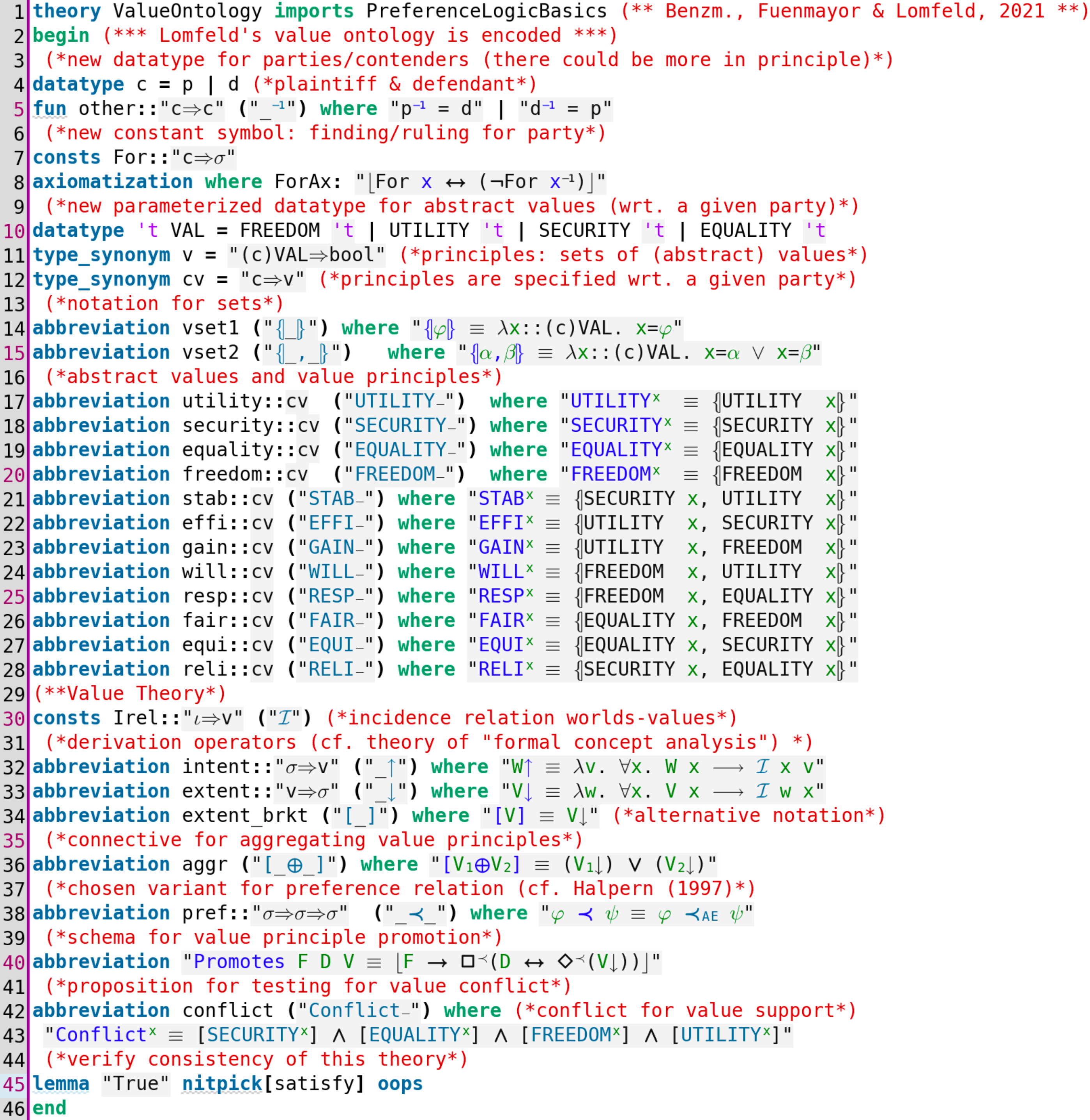}}
  \caption{Encoding of the legal DSL (value ontology)}
  \label{fig:Appl}
\end{figure}

The encoding of the legal DSL (value theory or ontology) is shown in
Fig.~\ref{fig:Appl}. The new theory is termed ``ValueOntology'', and
it imports theory ``PreferenceLogicBasics'' (and thus recursively also
\Isabelle's internal theory ``Main'').

As a preliminary, the legal parties \textit{plaintiff} and
\textit{defendant} are introduced as an (extensible) two-valued
datatype together with a function to obtain for a given party the
\textit{other} one ($x^{-1}$) (Lines 4--5); and a predicate modelling
the ruling \textit{for} a party is also provided (Lines 7--8).

As regards the \textit{discoursive grammar} value theory, a
four-valued (parameterised) datatype is introduced (Line 10) as
described in \S\ref{sec:ValueTheory}. Moreover, type-aliases (Lines
11--12) and set-constructor operators for values (Lines 14--15) are
introduced for ease of presentation. The legal principles from
\S\ref{sec:ValueTheory} are introduced as combinations of those
basic values (Lines 17--28). As an illustration, the principle
STABility is encoded as a set composed of the basic
values SECURITY and UTILITY.

Next, the incidence relation $I$ and operators $\uparrow$ and
$\downarrow$, borrowed and adapted from formal concept analysis (FCA), are
introduced (Lines 30-34).

We then define the aggregation operator $\oplus$ as
$A\oplus B :=~(A{\downarrow} \vee B{\downarrow})$; i.e., we select the
second candidate as discussed in \S\ref{sec:ValueTheory}. And as our
preference relation of choice we select the relation $\prec_{AE}$
(Line 38).

Finally we introduce ``$\text{Promotes}$'' schema for encoding the promotion of
value principles via legal decisions (Line 40) and we introduce a notion
"$\text{Conflict}^x$" expressing a legal value conflict for a party
$x$ (Lines 42-43).

The consistency of the theory is confirmed by {\Nitpick} (Line 45).

\begin{figure}[!bp]
  \centering
  \colorbox{gray!30}{\includegraphics[width=.97\textwidth]{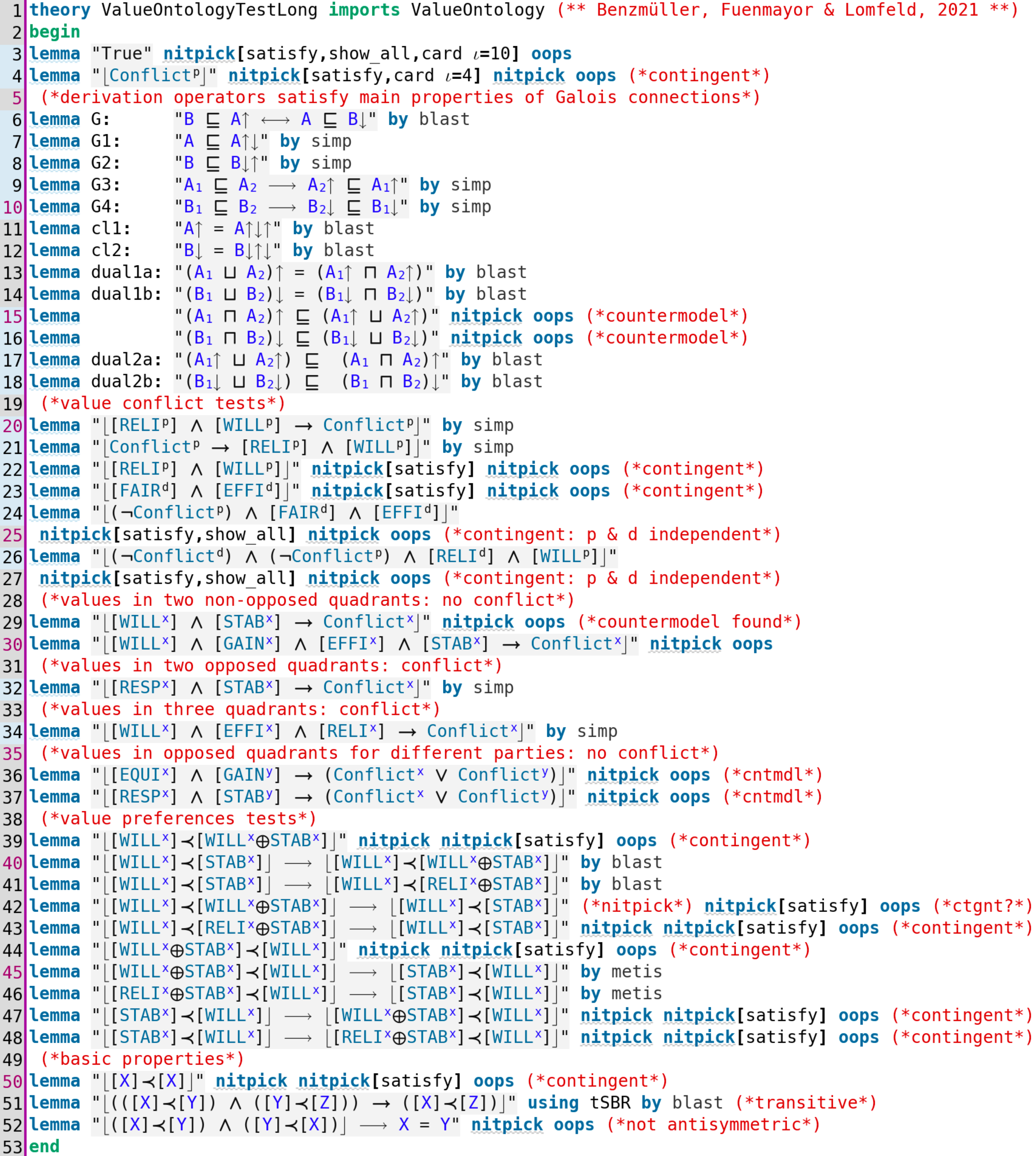}}
  \caption{Formally verifying/testing the legal DSL or value ontology}
  \label{fig:ValueOntologyTestLong}
\end{figure}

Tests on the modelling and encoding of the legal DSL are
displayed in Fig.~\ref{fig:ValueOntologyTestLong}.

Among others, we verify that the pair of operators for
\textit{extension} ($\downarrow$) and \textit{intension} ($\uparrow$),
cf.~\textit{Formal Concept Analysis} \parencite{ganter2012formal},
constitute indeed a Galois connection (Lines 6--18), and we carry out
some further tests on the value theory (extending the ones presented
in \S\ref{fig:TestingValueOntology}) concerning value aggregation and
consistency (Lines 20ff.).

\clearpage

\subsection{Legal and World Knowledge}
\label{subsec:LWK}
\begin{figure}[!bp]
  \centering
  \colorbox{gray!30}{\includegraphics[width=.97\textwidth]{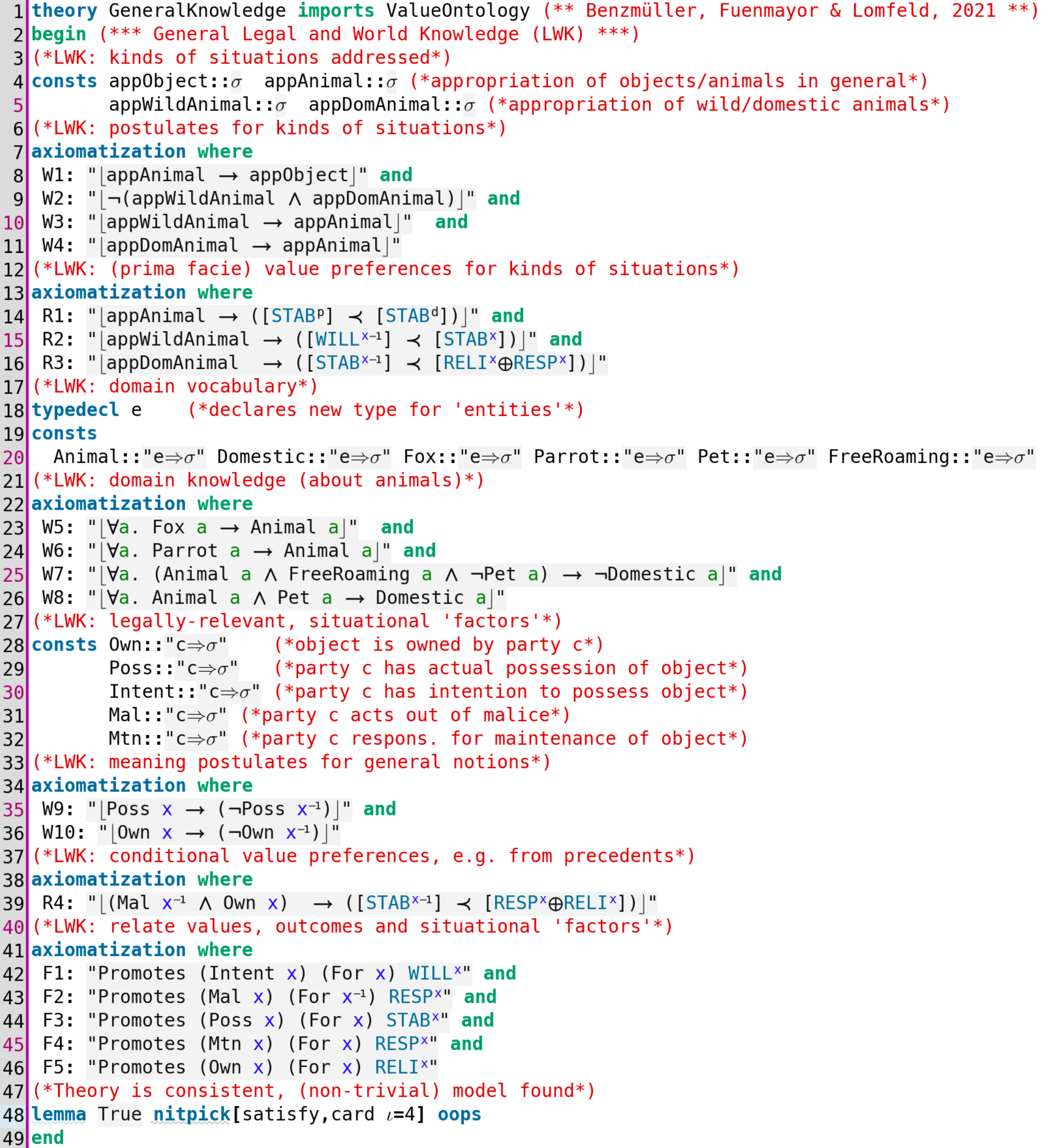}}
  \caption{Encoding of relevant legal \& world knowledge}
  \label{fig:GeneralKnowledge}
\end{figure}

The encoding of the relevant legal \& world knowledge (LWK) is shown
in Fig.~\ref{fig:GeneralKnowledge}.  The defined \Isabelle\ theory is
termed ``GeneralKowledge'' and imports the ``ValueOntology'' (and thus
recursively also ``PreferenceLogicBasics'') theory.

\begin{description}
\item[Lines 4--5] Declaration of logical constant symbols that stand
  for kinds of legally relevant situations.
\item[Lines 8--11] Meaning postulates for these kinds of legally
  relevant situations are introduced.
\item[Lines 14--16] Preference relations for these kinds of legally
  relevant situations are introduced.
\item[Lines 18--26] Some simple vocabulary is introduced and some
  taxonomic relations for wild and domestic animals are specified.
\item[Lines 28--36] Some relevant situational \textit{factors} are
  declared and subsequently constrained by meaning postulates.
\item[Line 39] An example for a value preference conditioned on
  \textit{factors} is specified.
\item[Lines 41--46] The situational \textit{factors} are related with
  values and with ruling outcomes according to the notion of value
  \textit{promotion}.
\item[Line 48] The model finder {\Nitpick} is used to confirm the
  consistency of the introduced theory.
\end{description}

\clearpage

\subsection{Modelling Pierson v.~Post}
\label{subsec:Pierson}
\begin{figure}[!hp] \centering
  \colorbox{gray!30}{\includegraphics[width=.97\textwidth]{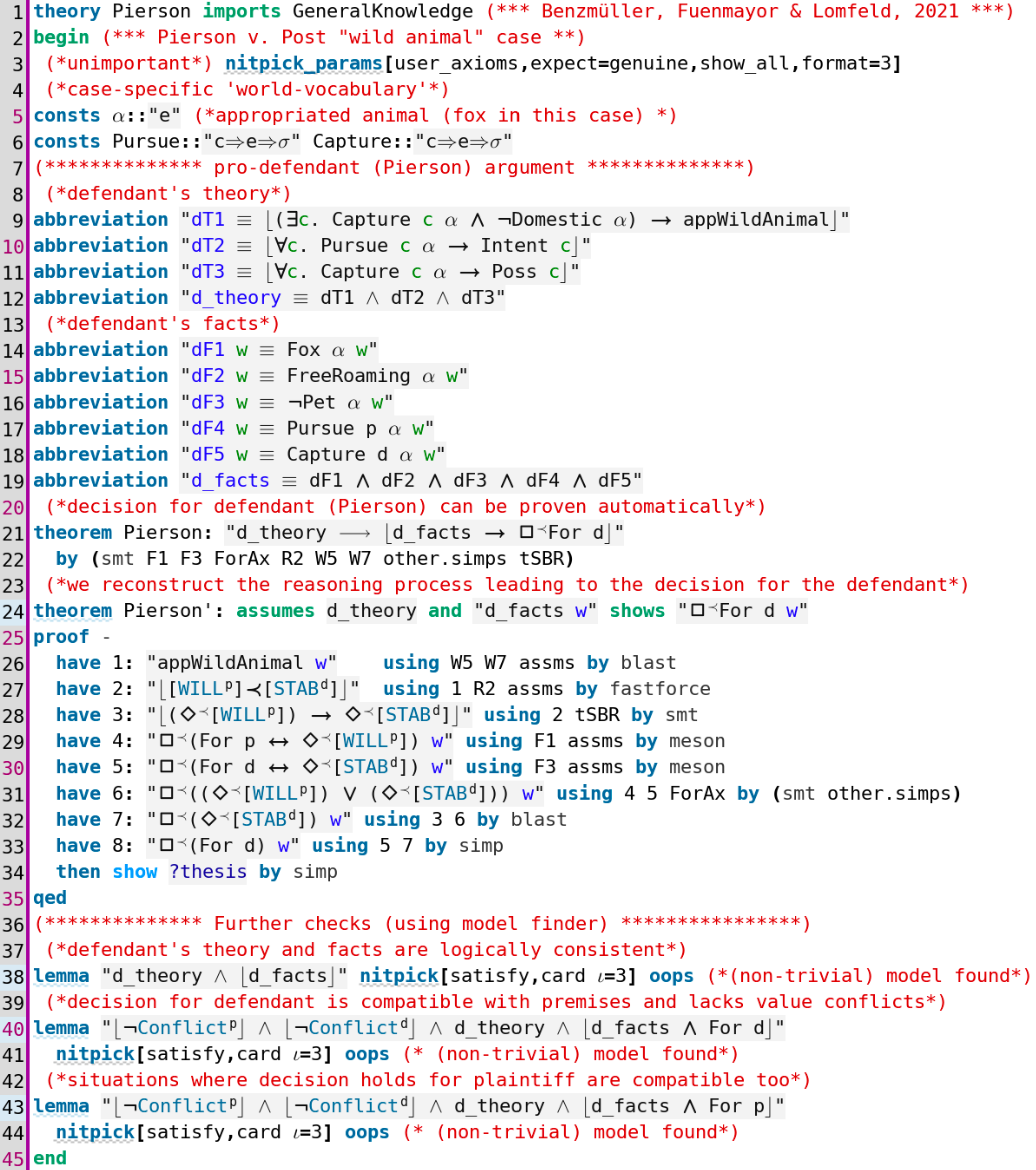}}
  \caption{Modelling the Pierson v. Post case; ruling for Pierson}
  \label{fig:Pierson}
\end{figure}

\begin{figure}[!bp] \centering
  \colorbox{gray!30}{\includegraphics[width=.97\textwidth]{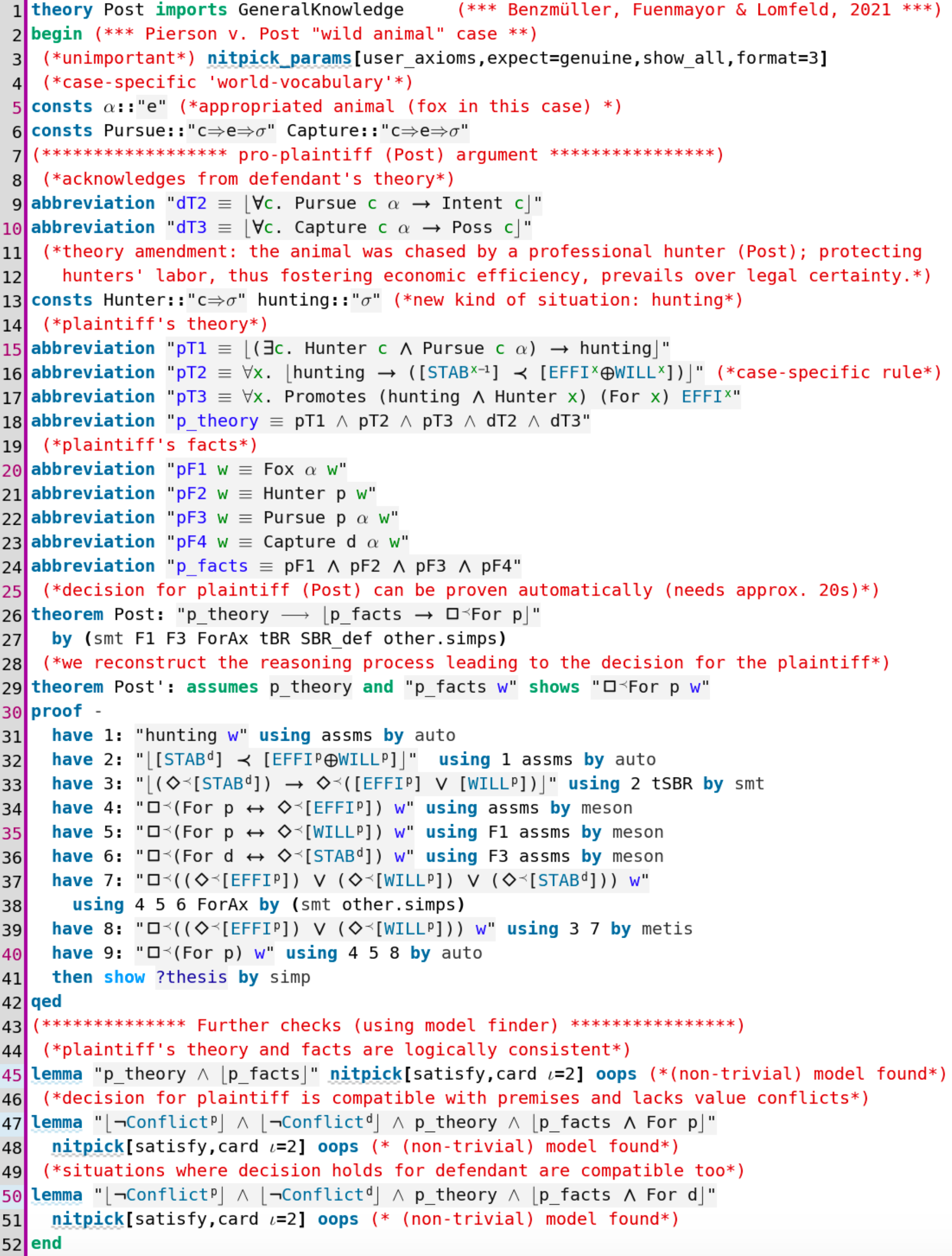}}
  \caption{Modelling the Pierson v. Post case; ruling for Post}
  \label{fig:Post}
\end{figure}

The \Isabelle\ encoding of two scenarios in the Pierson v.~Post case
is presented in Figs.~\ref{fig:Pierson} and ~\ref{fig:Post}.

In Fig.~\ref{fig:Pierson}, which presents the initial ruling in favour
of Pierson, the \Isabelle\ theory is termed ``Pierson'' and imports
the theory ``GeneralKnowledge'' (which recursively imports theories
``ValueOntology'' and ``PreferenceLogicBasics'').

\begin{description}
\item[Lines 5--19] (generic) theory and (contingent) facts suitable to
  the defendant (Pierson) are postulated.
\item[Lines 21--22] automated proof justifying the ruling for Pierson;
  the dependencies of the proof are shown.
\item[Lines 24--35] corresponding interactive proof (with the same
  dependencies as for the automated one) modelling the argument
  justifying the finding for Pierson.
\item[Lines 36--44] various checks for consistency of the assumptions
  and the absence of value conflicts.
\end{description}

As a further illustration, we present in Fig.~\ref{fig:Post} a
plausible counterargument by Post.  The \Isabelle\ theory is now
termed ``Post'' and imports the theory ``GeneralKnowledge'' (which
recursively imports theories ``ValueOntology'' and
``PreferenceLogicBasics'').

\begin{description}
\item[Lines 5--24] theory and facts suitable to the plaintiff (Post)
  are postulated.
\item[Lines 26--27] automated proof justifying the ruling for Post;
  the dependencies of the proof are shown.
\item[Lines 29--42] corresponding interactive proof (with the same
  dependencies as for the automated one) modelling the argument
  justifying the finding for Post.
\item[Lines 43--51] various checks for consistency of the assumptions
  and the absence of value conflicts.
\end{description}

\clearpage

\subsection{Modelling Conti v.~ASPCA} \label{sec:Conti} 
\begin{figure}[!bp]
  \centering
  \colorbox{gray!30}{\includegraphics[width=.97\textwidth]{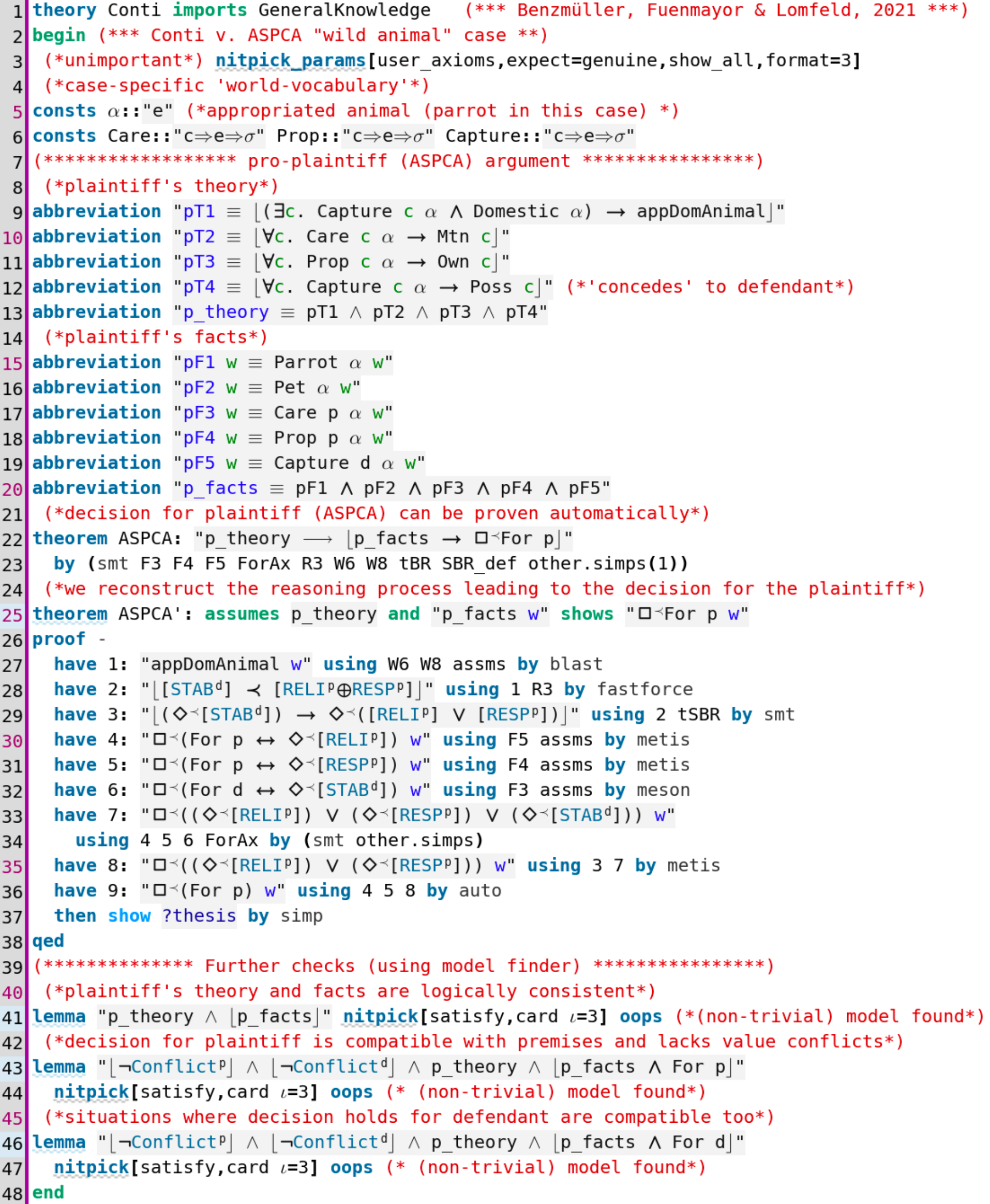}}
  \caption{Modelling of the Conti v. ASPCA case}
  \label{fig:Conti}
\end{figure}

The
reconstructed theory for the Conti v. ASPCA case is displayed in
Fig.~\ref{fig:Conti}. The \Isabelle\ theory is termed ``Conti'' and
imports the theory ``GeneralKnowledge'' (which recursively imports
theories ``ValueOntology'' and ``PreferenceLogicBasics'').

\begin{description}
\item[Lines 5--20] the theory and the facts of the pro-plaintiff
  (ASPCA) argument are formulated.
\item[Lines 22--23] automated proof justifying the ruling for ASPCA;
  the dependencies of the proof are shown.
\item[Lines 25--38] corresponding interactive proof (with the same
  dependencies as for the automated one) modelling the argument
  justifying the finding for ASPCA.
\item[Lines 39--47] various checks for consistency of the assumptions
  and the absence of value conflicts.
\end{description}

\clearpage

\subsection{Complex (Counter-)Models}
\begin{figure*}[!htp]
  \centering
  \includegraphics[width=\textwidth]{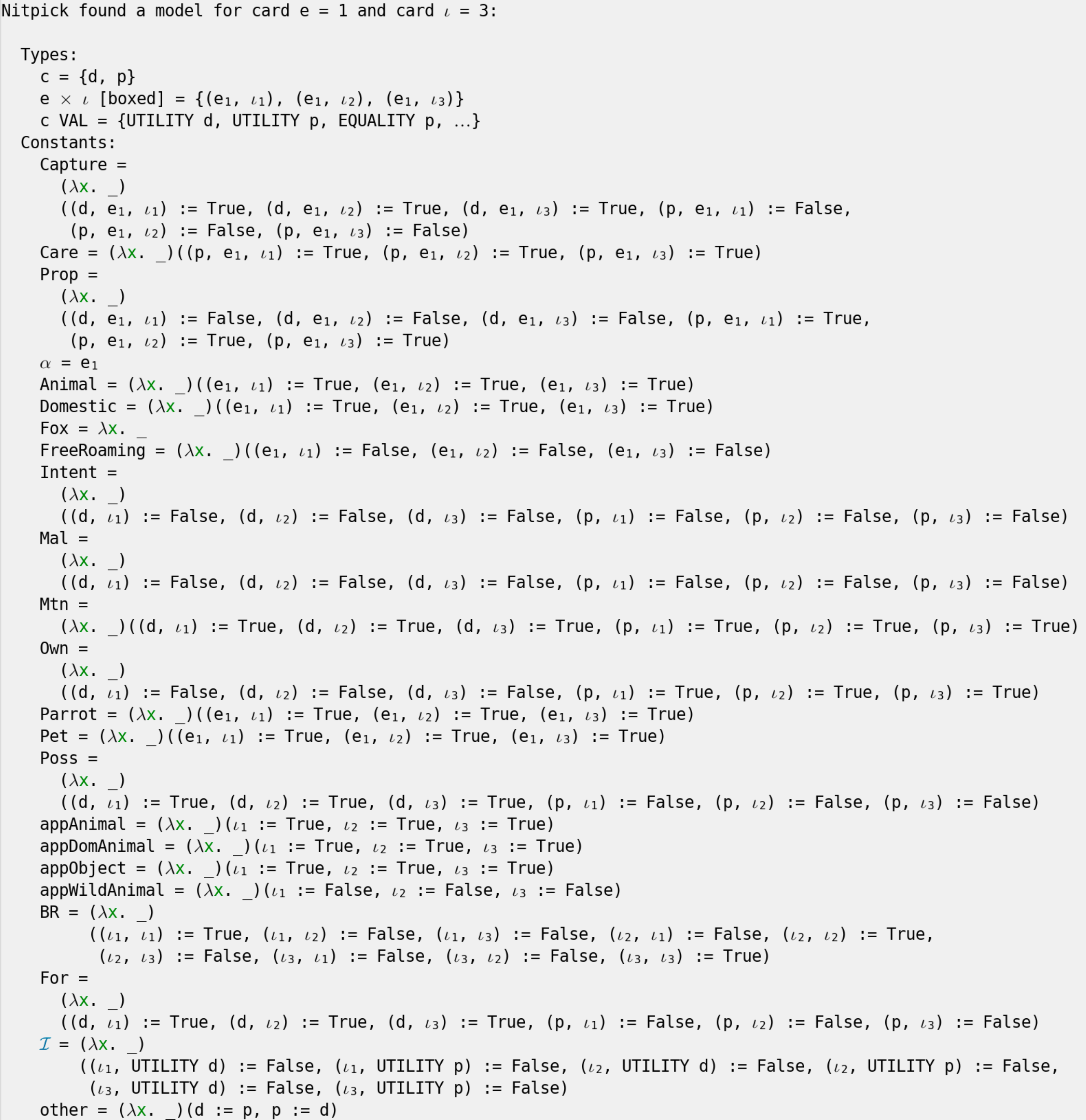}
  \caption{Example of a (satisfying) model to the statement in Line 26
    in Fig.~\ref{fig:Conti}}
  \label{fig:Model2}
\end{figure*}

In Fig.~\ref{fig:Model2} we present an example of a model computed by
model finder {\Nitpick} for the statement in Line 41 in
Fig.~\ref{fig:Conti}.  This non-trivial model features three possible
worlds/states.  It illustrates the richness of the information and the
level of detail that is supported in model (and countermodel) finding
technology for HOL.  This information is very helpful to support the
knowledge engineer and user of the \logikey\ framework to gain insight
about the modelled structures.  We observe that the proof assistant
{\Isabelle} allows for the parallel execution of its integrated tools.
We can thus execute, for a given candidate theorem, all three tasks in
parallel (and in different modes): theorem proving, model finding, and
countermodel finding. This is one reason for the very good response
rates we have experienced in our work with the system -- despite the
general undecidability of HOL.

\end{appendix}
\end{document}